%% file: main.tex
\documentclass[11pt,letterpaper]{mystyle}
\usepackage[utf8]{inputenc}
\usepackage[T1]{fontenc}
\usepackage[numbers]{natbib}
\usepackage{adjustbox}
\usepackage{float}
\usepackage{breakurl}    
\usepackage{hyperref}
\usepackage{arydshln}  
\usepackage[edges]{forest}
\usepackage{subcaption}
\usepackage{soul}
\usepackage{multirow}
\usepackage[utf8]{inputenc} 
\usepackage{booktabs}       
\usepackage{amsfonts}       
\usepackage{nicefrac}      
\usepackage{microtype}     
\usepackage{xcolor}        
\usepackage{colortbl}
\usepackage{enumitem}
\usepackage{amssymb}
\usepackage{wrapfig}
\usepackage{courier}
\usepackage{bxcoloremoji}
\usepackage{CJKutf8}
\usepackage{xspace}
\usepackage{textcomp}
\usepackage{listings}
\usepackage{tabularray}
\usepackage{fontawesome5}
\usepackage{pifont}
\usepackage{makecell}
\usepackage{fancyhdr}
\usepackage{setspace}
\usepackage{threeparttable}
\usepackage{supertabular}
\usepackage{bm}
\usepackage{amsmath}
\usepackage{amsthm}
\usepackage{mathrsfs}
\usepackage{siunitx}
\usepackage{footmisc}
\usepackage{array}
\usepackage{CJK}
\usepackage{footnote}
\usepackage{tabularx}
\usepackage{tikz}

\usetikzlibrary{trees,positioning,shapes,shadows,arrows.meta}

\definecolor{hidden-draw}{RGB}{0,0,0}
\definecolor{hidden-blue}{RGB}{194,232,247}
\definecolor{hidden-orange}{RGB}{243,202,120}
\definecolor{hidden-yellow}{RGB}{242,244,193}
\definecolor{tree-level-1}{RGB}{245,20,85}
\definecolor{tree-level-2}{RGB}{246,86,118}
\definecolor{tree-level-3}{RGB}{248,177,193}
\definecolor{tree-leaf}{RGB}{176,230,198}

\tcbuselibrary{skins}

\definecolor{darkblue}{rgb}{0, 0, 0.5}
\definecolor{darkgreen}{RGB}{50,100,0}
\definecolor{darkred}{RGB}{200, 0, 0}
\definecolor{lightblue}{RGB}{220,235,250}
\hypersetup{colorlinks=true, citecolor=darkblue, linkcolor=darkblue, urlcolor=darkblue}

\definecolor{gray}{gray}{0.5}

\newcommand{\blackbox}{\rule{0.8em}{0.8em}}

\newcommand{\github}{\raisebox{-1.5pt}{\includegraphics[height=1.05em]{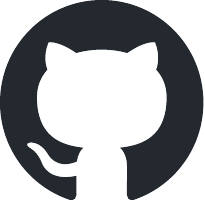}}}

\fancypagestyle{headstyle}{
    \fancyhead[C]{}
    \fancyhead[R]{}

}

\definecolor{hidden-red}{RGB}{205, 44, 36}
\definecolor{hidden-blue}{RGB}{194,232,247}
\definecolor{hidden-orange}{RGB}{243,202,120}
\definecolor{hidden-green}{RGB}{34,139,34}
\definecolor{hidden-pink}{RGB}{255,245,247}
\definecolor{hidden-black}{RGB}{20,68,106}
\definecolor{purple}{RGB}{144,153,196}
\definecolor{yellow}{RGB}{255,228,123}
\definecolor{hidden-yellow}{RGB}{255,248,203}
\definecolor{tkcolor}{RGB}{224,223,255}
\definecolor{darkblue}{rgb}{0, 0.40, 0.75}
\newcommand{\eg}{\textit{e.g.,}\xspace}
\newcommand{\ie}{\textit{i.e.,}\xspace}
\definecolor{lightblue}{RGB}{220,235,250}
\hypersetup{colorlinks=true, citecolor=darkblue, linkcolor=darkblue, urlcolor=darkblue}

\newcommand{\drs}{deep research system\xspace}
\newcommand{\sps}{\,\,}

\newcommand{\Drs}{Deep research system\xspace}

\newcommand{\header}[1]{
  \vspace{0.5em}\noindent \textbf{#1}%
}

\newtcolorbox{TakeawayBox}[2][]{takeawaybox,title=#2,#1}

\tcbset{
  takeawaysbox/.style={
    colback=lightblue!80,
    colframe=black,
    fonttitle=\bfseries\small,
    coltitle=white,
    colbacktitle=black,
    enhanced,
    attach boxed title to top left={xshift=2.5mm,yshift=-2.5mm},
    boxed title style={rounded corners, size=small, colframe=black, colback=black},
    width=\linewidth,
    arc=3.5mm
  }
}

\title{
Deep Research: A Systematic Survey}
\author{
\textbf{Zhengliang Shi$^{1}$ \sps Yiqun Chen$^{2}$ \sps Haitao Li$^{3}$ \sps Weiwei Sun$^{4}$ \sps Shiyu Ni$^{5}$ \sps  Yougang Lyu$^{6}$}\\
\textbf{Run-Ze Fan$^{7}$ \sps Bowen Jin$^{8}$ \sps Yixuan Weng$^{9}$ \sps Minjun Zhu$^{9}$ \sps Qiujie Xie$^{9}$ \sps Xinyu Guo$^{10}$ \sps Qu Yang$^{11}$ }\\
\textbf{ Jiayi Wu$^{11}$ \sps Jujia Zhao$^{12}$ \sps Xiaqiang Tang$^{11}$ \sps Xinbei Ma$^{11}$ \sps Cunxiang Wang$^{3}$ \sps  Jiaxin Mao$^{2}$}\\ 
\textbf{Qingyao Ai$^3$ \sps Jen-Tse Huang$^{13}$ \sps Wenxuan Wang$^2$ \sps Yue Zhang$^{9}$ \sps Yiming Yang$^{4}$}\\
\textbf{Zhaopeng Tu$^{11, { } \coloremojicode{2709}}$ \sps Zhaochun Ren$^{12, { } \coloremojicode{2709}}$}\\
$^1$Shandong University \sps  $^2$Renmin University of China \sps $^3$Tsinghua University   \\
$^4$Carnegie Mellon University \sps $^5$UCAS \sps  $^6$University of Amsterdam \\
$^7$University of Massachusetts Amherst \sps $^8$University of Illinois Urbana-Champaign \\
$^{9}$Westlake University \sps  $^{10}$University of Arizona \sps $^{11}$Tencent \\
$^{12}$Leiden University \sps  $^{13}$Johns Hopkins University
 }

\begin{document}

\begin{abstract}
  \vspace{5mm}
  \textbf{\large Abstract:}
Large language models (LLMs) have rapidly evolved from text generators into powerful problem solvers.
Yet, many open tasks demand critical thinking, multi-source, and verifiable outputs, which are beyond single-shot prompting or standard retrieval-augmented generation.
Recently, numerous studies have explored \textit{Deep Research} (DR), which aims to combine the reasoning capabilities of LLMs with external tools, such as search engines, thereby empowering LLMs to act as research agents capable of completing complex, open-ended tasks.
This survey presents a comprehensive and systematic overview of deep research systems, including a clear roadmap, foundational components, practical implementation techniques, important challenges, and future directions.
Specifically, our main contributions are as follows:
(i) we formalize a three-stage roadmap and distinguish deep research from related paradigms;
(ii) we introduce four key components: query planning, information acquisition, memory management, and answer generation, each paired with fine-grained sub-taxonomies;
(iii) we summarize optimization techniques, including prompting, supervised fine-tuning, and agentic reinforcement learning; and
(iv) we consolidate evaluation criteria and open challenges, aiming to guide and facilitate future development.
\textit{\textcolor{blue}{As the field of deep research continues to evolve rapidly, we are committed to continuously updating this survey to reflect the latest progress in this area.}}
  \vspace{5mm}

  $^{\coloremojicode{2709}}$ \textit{Corresponding Author}

  \vspace{5mm}
  \vspace{5mm}

  \coloremojicode{1F4C5} \textbf{Date}: November 13, 2025

  \github{} \textbf{Code Repository}: \url{https://github.com/mangopy/Deep-Research-Survey}

  \coloremojicode{1F4E7} \textbf{Contact}: 
  \href{mailto:zhengliang.shii@gmail.com}{zhengliang.shii@gmail.com}
  \href{mailto:chenyiqun990321@ruc.edu.cn}{chenyiqun990321@ruc.edu.cn}
  \href{mailto:z.ren@liacs.leidenuniv.nl}{z.ren@liacs.leidenuniv.nl}

\end{abstract}
\maketitle

\vspace{3mm}
\pagestyle{headstyle}
\thispagestyle{empty}
\newpage
\tableofcontents

\clearpage

\input{sections/01-introduction}

\input{sections/02-roadmap}

\input{sections/03-component}

\input{sections/04-technique}

\input{sections/05-evaluation}

\input{sections/06-challenges}

\input{sections/07-discussion}

\input{sections/08-related-work}
\input{sections/08-conclusion}

\newpage

\bibliographystyle{refstyle}
\bibliography{ref}

\end{document}

%% file: sections/01-introduction.tex
\section{Introduction}\label{sec:intro}

Large language models (LLMs), trained on web-scale corpora, have rapidly evolved from fluent text generators into autonomous agents capable of long-horizon reasoning in practical complex applications~\citep {naveed2025comprehensive, gao2024large,zhang2025rlvmr,shi2025social}.
They have exhibited strong generalization across diverse domains, including mathematical reasoning~\citep{He2025DeepMath103KAL,zhang2025deeptheorem}, creative writing~\citep{GomezRodriguez2023ACO}, and practical software engineering~\citep{Hou2023LargeLM, Jimenez2023SWEbenchCL, lee2025unidebugger}.
Many real-world tasks are inherently open-ended, involving \textbf{critical thinking}, \textbf{factually grounded information}, and the production of \textbf{self-contained} responses.
This is far beyond what single-shot prompting or static parametric knowledge can provide~\citep{Huang2023ASO,li2025webthinker,shi2025iterative}.
To address this gap, the \textbf{Deep Research (DR)} paradigm~\cite{OpenAI2025a, Google2025b, du2025deepresearch, zheng2025deepresearcher, huang2025deep, lyu2025deepshop} has emerged.
DR frames LLMs within an end-to-end research workflow that iteratively decomposes complex problems, acquire evidence via tool use, and synthesizes validated insights into coherent long-form answers.


Despite rapid progress, there remains no comprehensive survey that systematically analyzes the key components, technical details, and open challenges of DR.
Most existing work~\citep{Zhang2025DeepRA, chen2025ai4research} mainly summarizes developments in related areas such as Retrieval-Augmented Generation (RAG) and web-based agents~\citep{Xi2025ASO,endgraph,shi-etal-2024-learning,Zhang2025FromWS,sun2023contrastive}.
However, in contrast to RAG~\citep{Gao2023RetrievalAugmentedGF, Fan2024ASO}, DR adopts a more flexible, autonomous workflow that eschews handcrafted pipelines and aims to produce coherent, evidence-grounded reports.
Therefore, a clear overview of its technical landscape is urgent but remains a challenge.
This survey fills this gap by providing a comprehensive synthesis of DR: mapping its core components to representative system implementations, consolidating key techniques and evaluation methodologies, and establishing a foundation for consistent benchmarking and sustained progress in AI-driven research.

\begin{figure}[htbp]
        \centering
\includegraphics[width=1\textwidth]{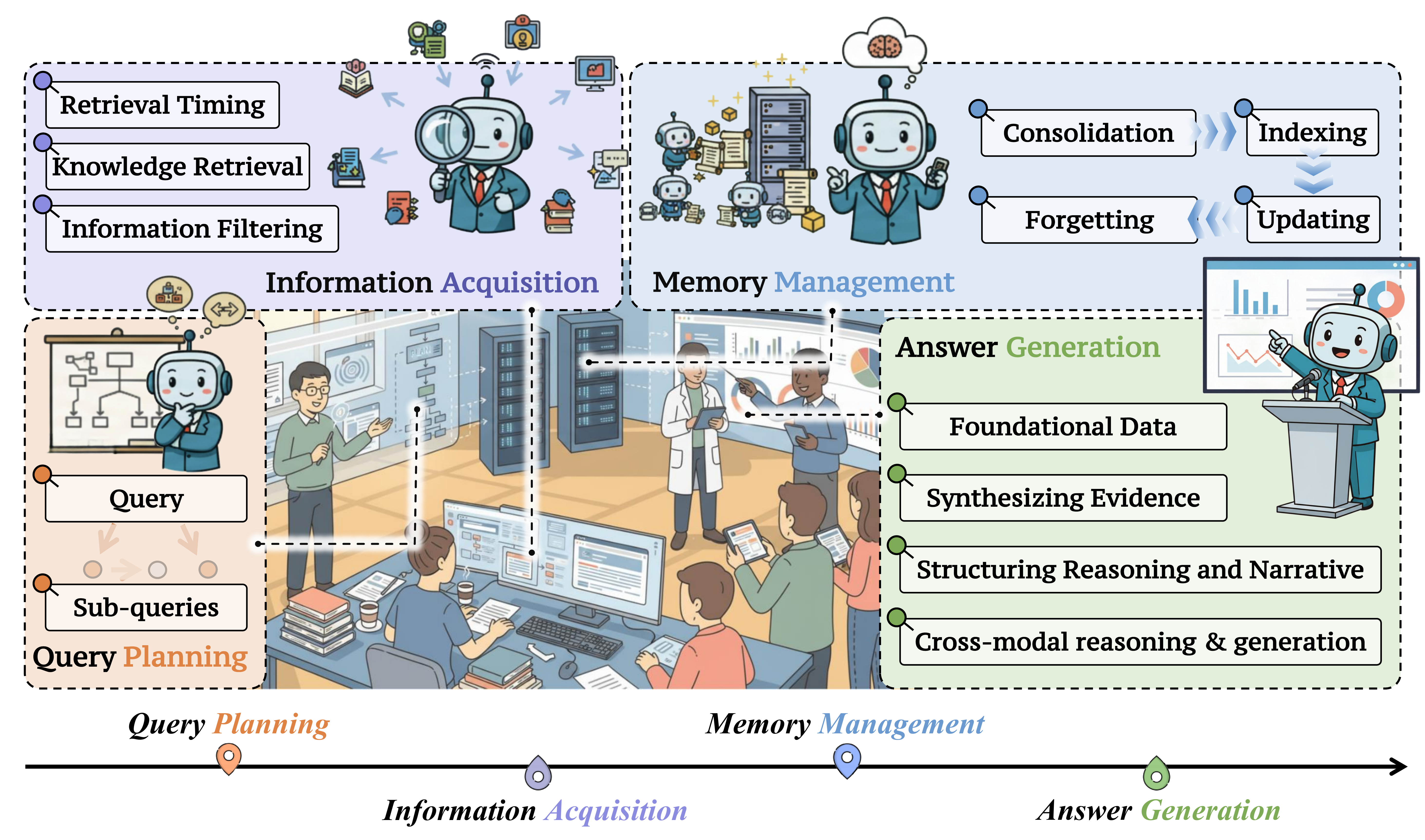}
        \caption{An overview of four key components in a general deep research system, including: Task Planning (Section~\ref{query-planning}). Information Acquisition (Section~\ref{information-acquision}).
        Memory Management  (Section~\ref{memory-management}) and Answer Generation  (Section~\ref{answer-generation}).}
\label{fig:intro}
\end{figure}

In this survey, we propose a three-stage roadmap for DR systems, illustrating their broad applications ranging from agentic information seeking to autonomous scientific discovery.
Based on the roadmap, we summarize the key components of the task-solving workflow for the most commonly used DR systems.
Specifically, we present four foundational components in DR: 
(i) \textit{query planning}, which decomposes the initially input query into a series of simpler, sub-queries~\citep{press2022measuring, yao2023react}; 
(ii) \textit{information acquisition}, which invokes external retrieval, web browsing, or various tools on demand~\citep{lewis2020retrieval, nakano2021webgpt}; 
(iii) \textit{memory management}, which ensures relevant task-solving context through controlled updating or folding~\citep{packer2023memgpt};
(iv) \textit{answer generation}, which produces comprehensive outputs with explicit source attribution, \eg a scientific report.
This scope is distinct from standard RAG~\cite{Gao2023RetrievalAugmentedGF,Fan2024ASO} techniques, which typically treat retrieval as a heuristic augmentation step, without a flexible research workflow or a broader action space.
We also introduce how to optimize DR systems in effectively coordinating these components, categorizing existing approaches into three types: (i) \textit{workflow prompting}; (ii) \textit{supervised fine-tuning} (SFT), and (iii) \textit{end-to-end reinforcement learning} (RL).


The remainder of this paper is organized as follows: 
Section~\ref{sec:roadmap} clearly defines DR and its boundaries; 
Section~\ref{sec:components} introduces four key components in DR;
Section~\ref{sec:practice} introduces technique details about optimizing a DR system;
Section~\ref{sec:evaluation} summarizes well-known evaluation datasets and resources, and Section~\ref{sec:challenge} discusses challenges for future directions.

To sum up, our survey makes the following contributions:
(i) We formalize a three-stage roadmap of DR and clearly distinguish it from related techniques such as standard retrieval-augmented generation;
(ii) We introduce four key components of DR systems, together with fine-grained sub-taxonomies for each, to provide a comprehensive view of the research loop;
(iii) We summarize detailed optimization approaches for building DR systems, offering practical insights into workflow prompting, supervised fine-tuning, and reinforcement learning; and
(iv) We consolidate evaluation criteria and open challenges to enable comparable reporting and to guide future research.

%% file: sections/02-roadmap.tex
\input{sections/tree_taxonomy}

\section{Preliminary Concept of Deep Research}\label{sec:roadmap}

\subsection{What is Deep Research}\label{sec:definition}

DR aims to endow LLMs with an \textbf{end-to-end research workflow}, enabling them to function as agents that generate coherent, source-grounded reports with minimal human supervision.
Such systems automate the entire research loop, spanning planning, evidence acquisition, analysis, and reporting.
In a DR setting, the LLM agent plans queries, acquires and filters evidence from heterogeneous sources (\eg the web, tools, and local files), maintains and revises a working memory, and synthesizes verifiable answers with explicit attribution.
Below, we formally introduce a three-phase roadmap that structures the rapidly evolving, capability-oriented landscape of DR, and we compare it systematically with conventional RAG paradigms.

\subsection{Understanding Deep Research from Three Phases}\label{sec:evolution}

We view DR as a capability trajectory rather than a value hierarchy. The three phases below capture a progressive expansion of what systems can reliably do, from acquiring precise evidence, to synthesizing it into readable analyses, and finally to forming defensible insights. 

\header{Phase I: \textbf{Agentic Search}.}
Phase~I systems specialize in finding the correct sources and extracting answers with minimal synthesis. They typically reformulate the user query (via rewriting or decomposition) to improve recall, retrieve and re-rank candidate documents, apply lightweight filtering or compression, and produce concise answers supported by explicit citations. The emphasis is on faithfulness to retrieved content and predictable runtime. Representative applications include open-domain question answering~\cite{nguyen2016ms,kwiatkowski2019natural}, multi-hop question answering~\cite{yang2018hotpotqa, trivedi2022musique,rein2024gpqa}, and other information-seeking tasks~\cite{roy-etal-2024-learning,zaib2022conversational,tanjim2025disambiguation,fan2019eli5,mialon2023gaia} where truth is localized to a small set of sources.
Evaluation prioritizes retrieval recall@k and answer exact matching, complemented by citation correctness and end-to-end latency, reflecting the phase’s focus on accuracy-per-token and operational efficiency.

\header{Phase II: \textbf{Integrated Research}.}
Phase~II systems move beyond isolated facts to produce coherent, structured reports that integrate heterogeneous evidence while managing conflicts and uncertainty. 
The research loop becomes explicitly iterative: systems plan sub-questions, retrieve and extract key evidence from various raw content (\eg  HTML~\cite{tan2025htmlrag}, tables~\cite{chernyshevich2025core,nguyen2024interpretable}, and charts~\cite{masry2022chartqa,masry2022chartqa}), and ultimately synthesize comprehensive, narrative reports.
The most commonly-used applications include market and competitive analysis~\cite{zhao2023competeai,vinogradova2025llm}, policy briefs~\cite{wang2025policypulse}, itinerary design under constraints~\cite{tang2024itinera}, and other long-horizon question answering~\cite{du2025deepresearch,yoran2024assistantbench,wei2025browsecomp,coelho2025deepresearchgym}. 
Accordingly, evaluation shifts from superficial short-form lexical matching to long-form quality, including:
fine-grained factuality~\cite{chern2023factool,min-etal-2023-factscore},
verified citations~\cite{sun-etal-2024-towards-verifiable,gao-etal-2023-enabling},
structural coherence~\cite{cao2024structeval},
key points coverage~\cite{wei2024long}.
Phase~II thus trades a modest increase in compute and complexity for substantial gains in clarity, coverage, and decision support.

\header{Phase III: \textbf{Full-stack AI Scientist}.}
Phase~III aims at advancing scientific understanding and creation beyond mere information aggregation, representing a broader and more ambitious stage of DR 
In this phase, DR agents are expected not only to aggregate evidence but also to generate hypotheses~\cite{zhou-etal-2024-hypothesis}, conduct experimental validation or ablation studies~\cite{nathani2025mlgym}, critique existing claims~\cite{zhu-etal-2025-deepreview}, and propose novel perspectives~\cite{weng2025deepscientist}. 
Common applications include paper reviewing~\cite{zhuang2025large,peng2024review,zhu-etal-2025-deepreview}, scientific discovery~\cite{zhang2024comprehensive,shojaee2025llm,shojaee2024llm}, and experiment automation~\cite{wang2024openhands,zhao2024commit0}. 
Evaluation at this stage emphasizes the novelty and insightfulness of the findings, the argumentative coherence, the reproducibility of claims (including the ability to re-derive results from cited sources or code), and calibrated uncertainty disclosure.


\input{table/comparison}

\subsection{Comparing Deep Research with RAG}\label{sec:why}

Many real-world tasks are inherently open-ended, involving \textbf{critical thinking}, \textbf{factually grounded information}, and \textbf{self-contained} responses.
These present several fundamental limitations of existing approaches. Below, we summarize three key challenges that cannot be solved by conventional RAG or scaling LLM parameters alone:
\begin{itemize}
    \item  \colorbox{gray!25}{\textit{Flexible Interaction with the Digital World.}} 
    Conventional RAG systems operate in a static retrieval loop, relying solely on pre-indexed corpora~\cite{oche2025systematic,neha2025traditional}.
    However, real-world tasks often require active interaction with dynamic environments such as search engines, web APIs, or even Code executors~\cite{zhou2023webarena,nathani2025mlgym,wang2024openhands}.
    DR systems extend this paradigm by enabling LLMs to perform multi-step, tool-augmented interactions, allowing agents to access up-to-date information, execute operations, and verify hypotheses within a digital ecosystem.

    \item \colorbox{blue!25}{\textit{Long-horizon Planning with Autonomous Workflows.}}
    Complex research-like problems often require agents to coordinate multiple subtasks~\cite{wei2025browsecomp}, manage task-solving context~\cite{xu2025memory}, and iteratively refine intermediate outcomes~\cite{shinn2023reflexion}. DR addresses this limitation through closed-loop control and multi-turn reasoning, allowing agents to autonomously plan, revise, and optimize their workflows toward long-horizon objectives.

    \item \colorbox{green!25}{\textit{Reliable Language Interfaces for Open-ended Tasks.}}
    LLMs are prone to hallucination and inconsistency~\cite{hadi2023survey,zhao2023survey,huang2025survey,bang-etal-2025-hallulens,cossio2025comprehensive}, particularly in open-ended settings.
    DR systems introduce verifiable mechanisms that align natural language outputs with grounded evidence, establishing a more reliable interface between human users and autonomous research agents.
\end{itemize}



%% file: sections/tree_taxonomy.tex
\tikzstyle{my-box}=[
rectangle,
draw=hidden-black,
rounded corners,
text opacity=1,
minimum height=1.5em,
minimum width=5em,
inner sep=2pt,
align=left,
fill opacity=.5,
]
\tikzstyle{leaf3}=[
my-box,
minimum height=1.5em,
fill=yellow!32,
text=black,
align=left,
font=\normalsize,
inner xsep=5pt,
inner ysep=4pt,
align=left,
text width=45em,
]
\tikzstyle{leaf6}=[
my-box,
minimum height=1.5em,
fill=purple!30,
text=black,
align=left,
font=\normalsize,
inner xsep=5pt,
inner ysep=4pt,
]
\tikzstyle{leaf4}=[
my-box,
minimum height=1.5em,
fill=hidden-blue!57,
text=black,
align=left,
font=\normalsize,
inner xsep=5pt,
inner ysep=4pt,
]
\tikzstyle{leaf2}=[
my-box,
minimum height=1.5em,
fill=hidden-green!20,
text=black,
align=left,
font=\normalsize,
inner xsep=5pt,
inner ysep=4pt,
]
\tikzstyle{leaf}=[
my-box,
minimum height=1.5em,
fill=hidden-red!20,
text=black,
align=left,
font=\normalsize,
inner xsep=5pt,
inner ysep=4pt,
]
\tikzstyle{leaf5}=[
my-box,
minimum height=1.5em,
fill=darkblue!15,
text=black,
align=left,
font=\normalsize,
inner xsep=5pt,
inner ysep=4pt,
]
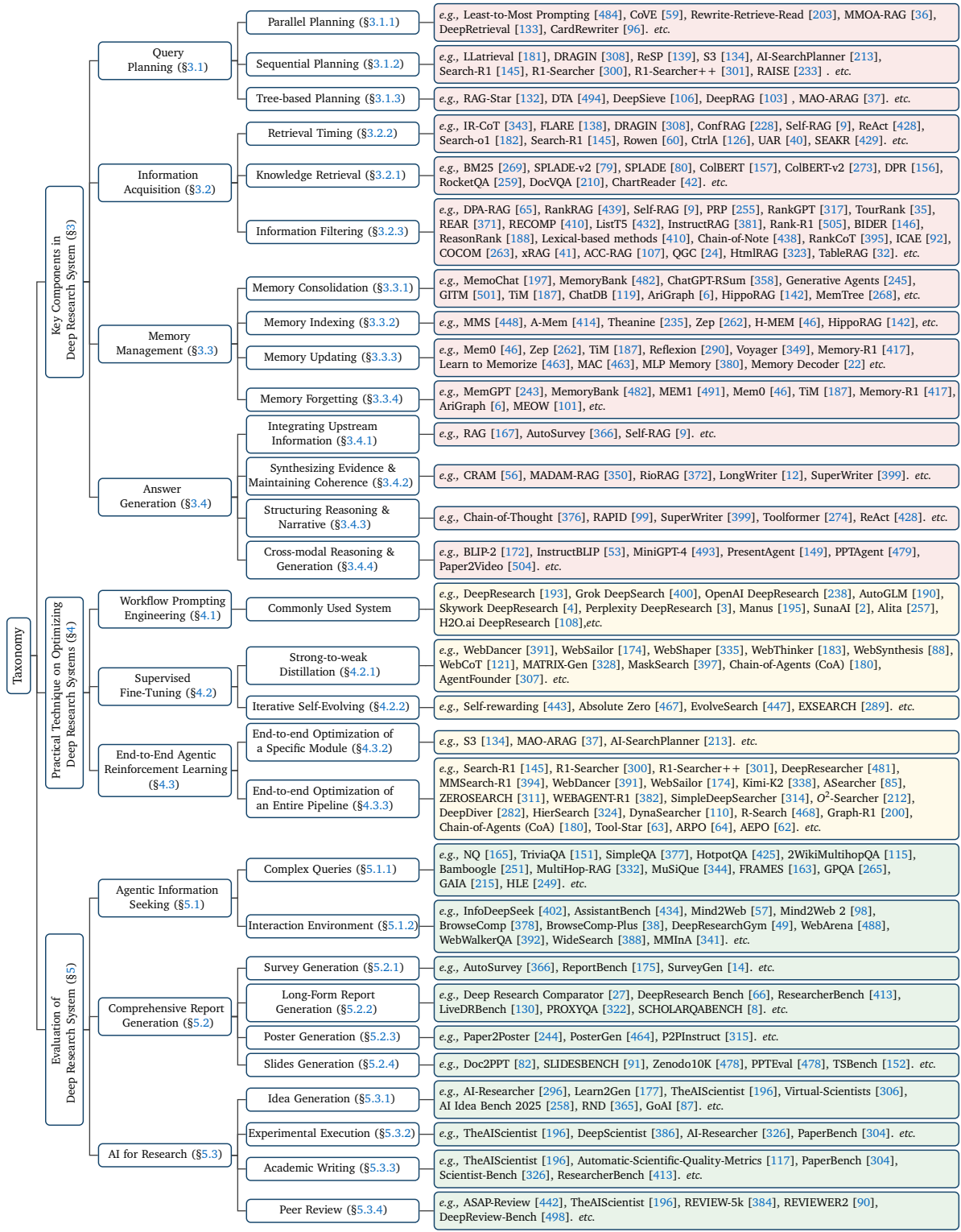
\begin{figure*}[!t]
        \vspace{-2mm}
        \centering
        \resizebox{0.96\textwidth}{!}{
                \begin{forest}
                        forked edges,
                        for tree={
                        grow=east,
                        reversed=true,
                        anchor=base west,
                        parent anchor=east,
                        child anchor=west,
                        base=left,
                        font=\large,
                        rectangle,
                        draw=hidden-black,
                        rounded corners,
                        align=left,
                        minimum width=4em,
                        edge+={darkgray, line width=1pt},
                        s sep=3pt,
                        inner xsep=2pt,
                        inner ysep=4pt,
                        line width=1.1pt,
                        ver/.style={rotate=90, child anchor=north, parent anchor=south, anchor=center},
                        },
                        where level=1{text width=16em,font=\normalsize,align=center,}{},
                        where level=2{text width=11em,font=\normalsize,align=center,}{},
                        where level=3{text width=14.4em,font=\normalsize,align=center,}{},
                        where level=4{text width=12em,font=\normalsize,align=left,}{},
                        where level=5{text width=50em,font=\normalsize,align=left}{},
                        [\ \ Taxonomy\ \ \ , ver
                        [\ \ \ \ \ \ Key Components in \\ \ \ \ \ Deep Research System~(\S\ref{sec:components}),ver
                        [\ \ \ \ \ Query \\ \ \ \ \ \ \ \ Planning~(\S\ref{query-planning})\ \ \
                            [\ \ \ \ \ Parallel Planning~(\S\ref{sec:parallel-planning})
                                [\eg ~Least-to-Most Prompting~\citep{zhou2022least}{,} CoVE~\citep{dhuliawala2023chain}{,} Rewrite-Retrieve-Read~\citep{ma2023query}{,} MMOA-RAG~\citep{chen2025improving}{,} \\ DeepRetrieval~\citep{jiang2025deepretrieval}{,} CardRewriter~\citep{gong2025cardrewriter}{.}
                                \textit{etc.}, leaf, text width=44em]SurveyForgeSurveyForge
                            ]
                            [\ \ Sequential Planning~(\S\ref{sequential-planning})
                                [\eg ~LLatrieval~\citep{li2023llatrieval}{,} DRAGIN~\citep{su2024dragin}{,} ReSP~\citep{jiang2025retrieve}{,} S3~\citep{jiang2025s3}{,} AI-SearchPlanner~\citep{mei2025ai}{,} \\ Search-R1~\citep{jin2025search}{,} R1-Searcher~\citep{song2025r1}{,} R1-Searcher++~\citep{song2025r1++}{,} RAISE~\cite{oh2025raise} {.}
                                \textit{etc.}, leaf, text width=44em]SurveyForgeSurveyForge
                            ]
                            [\ \ Tree-based Planning~(\S\ref{tree-based-planning})
                                [\eg ~RAG-Star~\citep{jiang2024rag}{,} DTA~\citep{zhu2025divide}{,} DeepSieve~\citep{guo2025deepsieve}{,} DeepRAG~\cite{guan2025deeprag} {,} MAO-ARAG~\citep{chen2025mao}{.}
                                \textit{etc.}, leaf, text width=44em]SurveyForgeSurveyForge
                            ]
                        ]
                        [\ \ \ \ Information \\ \ \ \ \ \ \   Acquisition~(\S\ref{information-acquision})\ \
                            [\ \ \ \ \ Retrieval Timing~(\S\ref{retrieval-timing})
                                [\eg ~IR-CoT~\citep{trivedi2022interleaving}{,} FLARE~\citep{jiang2023active}{,} DRAGIN~\citep{su2024dragin}{,} ConfRAG~\citep{ni2024llms}{,} Self-RAG~\citep{asai2024self}{,} ReAct~\citep{yao2022react}{,} \\ Search-o1~\citep{li2025search}{,} 
                                Search-R1~\citep{jin2025search}{,} Rowen~\citep{ding2024retrieve}{,} CtrlA~\citep{liu2024ctrla}{,} UAR~\citep{cheng2024unified}{,} SEAKR~\citep{yao2024seakr}{.}                                 \textit{etc.}, leaf, text width=44em]SurveyForgeSurveyForge
                            ]
                            [\ \ Knowledge Retrieval~(\S\ref{knowledge-retrieval})
                                [\eg ~BM25~\citep{robertson2009bm25}{,} SPLADE-v2~\citep{formal2021splade2}{,} SPLADE~\citep{formal2021splade}{,} ColBERT~\citep{khattab2020colbert}{,} ColBERT-v2~\citep{santhanam2021colbertv2}{,} DPR~\citep{karpukhin2020dense}{,} \\
                                RocketQA~\citep{qu2020rocketqa}{,} DocVQA~\citep{mathew2021docvqa}{,} ChartReader~\citep{cheng2023chartreader}{.}                                 \textit{etc.}, leaf, text width=44em]SurveyForgeSurveyForge
                            ]
                            [\ \ Information Filtering~(\S\ref{information-filtering})
                                [\eg ~DPA-RAG~\citep{dong2025understand}{,}  RankRAG~\citep{yu2024rankrag}{,} Self-RAG~\citep{asai2024self}{,} PRP~\citep{qin2023large}{,} RankGPT~\citep{sun2023chatgpt}{,} 
                                TourRank~\citep{chen2025tourrank}{,} \\
                                REAR~\citep{wang2024rear}{,}
                                RECOMP~\citep{xu2023recomp}{,} 
                                ListT5~\citep{yoon2024listt5}{,} InstructRAG~\citep{wei2024instructrag}{,} Rank-R1~\citep{zhuang2025rank}{,} 
                                BIDER~\citep{jin2024bider}{,} 
                                \\
                                ReasonRank~\citep{liu2025reasonrank}{,}
                                Lexical-based methods~\citep{xu2023recomp}{,} 
                                Chain-of-Note~\citep{yu2023chain}{,} RankCoT~\citep{wu2025rankcot}{,} 
                                ICAE~\citep{ge2023context}{,} \\
                                COCOM~\citep{rau2024context}{,} xRAG~\citep{cheng2024xrag}{,} ACC-RAG~\citep{Guo2025dynamic}{,} 
                                QGC~\citep{cao2024retaining}{,} HtmlRAG~\citep{tan2025htmlrag}{,} TableRAG~\citep{chen2024tablerag}{.}
                                \textit{etc.}, leaf, text width=44em]SurveyForgeSurveyForge
                            ]
                        ]
                        [\ \ \  Memory \\ \ \ \ \ Management~(\S\ref{memory-management})\ \ \
                            [\ Memory Consolidation~(\S\ref{memory-consolidation})
                                [\eg ~MemoChat~\cite{lu2023memochat}{,} MemoryBank~\cite{zhong2024memorybank}{,} ChatGPT-RSum~\cite{wang2025recursively}{,} Generative Agents~\cite{park2023generative}{,} \\ GITM~\cite{zhu2023ghost}{,} TiM~\cite{liu2023think}{,} ChatDB~\cite{hu2023chatdb}{,}
                                AriGraph~\cite{anokhin2024arigraph}{,}
                                HippoRAG~\cite{jimenez2024hipporag}{,}
                                MemTree~\cite{rezazadeh2024isolated}{,}
                                \textit{etc.}, leaf, text width=44em]SurveyForgeSurveyForge
                            ]
                            [\ \ \ \ Memory Indexing~(\S\ref{memory-indexing})
                                [\eg ~MMS~\cite{zhang2025multiple}{,} A-Mem~\cite{xu2025mem}{,} Theanine~\cite{ong2024towards}{,} Zep~\cite{rasmussen2025zep}{,} H-MEM~\cite{chhikara2025mem0}{,} HippoRAG~\cite{jimenez2024hipporag}{,}                               \textit{etc.}, leaf, text width=44em]SurveyForgeSurveyForge
                            ]
                            [\ \ \ \ Memory Updating~(\S\ref{memory-updating})
                                [\eg ~Mem0~\citep{chhikara2025mem0}{,} Zep~\cite{rasmussen2025zep}{,} TiM~\cite{liu2023think}{,} Reflexion~\cite{shinn2023reflexion}{,}
                                Voyager~\cite{wang2023voyager}{,}
                                Memory-R1~\cite{yan2025memory}{,} \\
                                Learn to Memorize~\cite{zhang2025learn}{,}
                                MAC~\cite{zhang2025learn}{,}
                                MLP Memory~\cite{wei2025mlp}{,}
                                Memory Decoder~\cite{cao2025memory}
                                \textit{etc.}, leaf, text width=44em]SurveyForgeSurveyForge
                            ]
                            [\ \ \ Memory Forgetting~(\S\ref{memory-forgetting})
                                [\eg ~MemGPT~\cite{packer2023memgpt}{,} MemoryBank~\cite{zhong2024memorybank}{,}
                                MEM1~\cite{zhou2025mem1}{,} 
                                Mem0~\cite{chhikara2025mem0}{,}
                                TiM~\cite{liu2023think}{,}
                                Memory-R1~\cite{yan2025memory}{,} \\
                                AriGraph~\cite{anokhin2024arigraph}{,}
                                MEOW~\cite{gu2024meow}{,}
                                 \textit{etc.}, leaf, text width=44em]SurveyForgeSurveyForge
                            ]
                        ]
                        [ \ Answer \\ \ \ \ \ \ Generation~(\S\ref{answer-generation})\ \ \
                            [\ \ \ \ \ Integrating Upstream \\ \ \ \ \ \ \ Information~(\S\ref{integrating-upstream-information})
                                [\eg ~RAG~\citep{lewis2020retrieval}{,} AutoSurvey~\citep{wang2024autosurvey}{,} Self-RAG~\citep{asai2024self}{.}                                 \textit{etc.}, leaf, text width=44em]SurveyForge
                            ]
                            [\ Synthesizing Evidence \& \\ Maintaining Coherence~(\S\ref{synthesizing-evidence-and-maintaining})
                                [\eg ~CRAM~\citep{deng2025cram}{,} MADAM-RAG~\citep{kim2024retrieval}{,} RioRAG~\citep{li2025reinforced}{,} LongWriter~\citep{zhang2024longwriter}{,} SuperWriter~\citep{liu2025superwriter}{.}                                 \textit{etc.}, leaf, text width=44em]SurveyForge
                            ]
                            [\ \ \ \ Structuring Reasoning \& \\ \ \ \ Narrative~(\S\ref{structuring-reasoning-and-narrative})
                                [\eg ~Chain-of-Thought~\citep{wei2022chain}{,} RAPID~\citep{yang2025rapid}{,} SuperWriter~\citep{liu2025superwriter}{,} Toolformer~\citep{schick2023toolformer}{,} ReAct~\citep{yao2022react}{.}                                 \textit{etc.}, leaf, text width=44em]SurveyForge
                            ]
                            [\ \ \ \ Cross-modal Reasoning \& \\ \ \ \ Generation~(\S\ref{cross-modal-reasoning-and-generation})
                                [\eg ~BLIP-2~\citep{li2023blip}{,} InstructBLIP~\citep{dai2023instructblip}{,} MiniGPT-4~\citep{chen2023minigpt}{,} PresentAgent~\citep{jingwei2025presentagent}{,} PPTAgent~\citep{zheng2025pptagentgeneratingevaluatingpresentations}{,} \\ Paper2Video~\citep{zhu2025paper2video}{.}                                 \textit{etc.}, leaf, text width=44em]SurveyForge
                            ]
                        ]
                        ]
                    [\ \ Practical Technique on Optimizing \\ Deep Research Systems~(\S\ref{sec:practice}), ver
                        [\ \ \ \ \ Workflow Prompting \\ \ \ \ \ \ Engineering~(\S\ref{sec:workflow})
                            [\ \ \ \ \  Commonly Used System \ \ \ \ \ \
                                [\eg ~DeepResearch~\citep{gemini_deep_research}{,} Grok DeepSearch~\citep{grok_deepsearch}{,}                                 OpenAI DeepResearch~\citep{openai_deep_research}{,} 
                                AutoGLM~\citep{liu2024autoglm}{,} \\
                                Skywork DeepResearch~\citep{skywork2025deepresearch}{,} Perplexity DeepResearch~\citep{perplexity_deep_research}{,} Manus~\citep{manus2025}{,} SunaAI~\citep{suna2025}{,} Alita~\citep{qiu2025alita}{,}            
                                \\
                                H2O.ai DeepResearch~\citep{H2Oai_DeepResearch}{,}\textit{etc.}, leaf3, text width=44em]SurveyForge
                            ]
                        ]
                        [ \ \ Supervised \\ \ \ \ \ \ Fine-Tuning~(\S\ref{sec:SFT})
                            [\ \ \ \ \ \ \ Strong-to-weak \\ \ \ \ \ \ \ \ \ Distillation~(\S\ref{sec:strong-to-weak-distillation})
                                [\eg ~WebDancer~\citep{wu2025webdancer}{,} WebSailor~\citep{li2025websailor}{,} WebShaper~\citep{tao2025webshaper}{,} WebThinker~\citep{li2025webthinker}{,} WebSynthesis~\citep{gao2025websynthesis}{,} 
                                \\
                                WebCoT~\citep{hu2025webcot}{,}
                                MATRIX-Gen~\citep{tang2024synthesizing}{,} MaskSearch~\citep{wu2025masksearch}{,} Chain-of-Agents (CoA)~\citep{li2025chain}{,} 
                                \\
                                AgentFounder~\citep{su2025scaling}{.}                                 \textit{etc.}, leaf3, text width=44em]SurveyForge
                            ]
                            [\ Iterative Self-Evolving~(\S\ref{sec:interative-self-evolving})
                                [\eg ~Self-rewarding~\citep{yuan2024self}{,} Absolute Zero~\citep{zhao2025absolute}{,} EvolveSearch~\citep{zhang2025evolvesearch}{,} EXSEARCH~\citep{shi2025iterative}{.}                                 \textit{etc.}, leaf3, text width=44em]SurveyForge
                            ]
                        ]
                        [ \ End-to-End Agentic \\  \ Reinforcement Learning \\(\S\ref{sec:agentic-RL})
                            [\ End-to-end Optimization of \\ \ \ a Specific Module~(\S\ref{subsec:specific_module_optimization})
                                [\eg ~S3~\citep{jiang2025s3}{,} MAO-ARAG~\citep{chen2025mao}{,} AI-SearchPlanner~\citep{mei2025ai}{.}
                                \textit{etc.}, leaf3, text width=44em]SurveyForge
                            ]
                            [\ End-to-end Optimization of \\ \ \ an Entire Pipeline~(\S\ref{subsec:entire_pipeline_optimization})
                                [\eg ~Search-R1~\citep{jin2025search}{,} R1-Searcher~\citep{song2025r1}{,} R1-Searcher++~\citep{song2025r1++}{,} DeepResearcher~\citep{zheng2025deepresearcher}{,} 
                                \\
                                MMSearch-R1~\citep{wu2025mmsearch}{,} WebDancer~\citep{wu2025webdancer}{,} WebSailor~\citep{li2025websailor}{,} Kimi-K2~\citep{team2025kimi}{,} ASearcher~\citep{gao2025beyond}{,} 
                                \\
                                ZEROSEARCH~\cite{sun2025zerosearch}{,} WEBAGENT-R1~\cite{wei2025webagent}{,} SimpleDeepSearcher~\cite{sun2025simpledeepsearcher}{,} $\textit{O}^2$-Searcher~\cite{mei20252}{,} 
                                \\
                                DeepDiver~\cite{shi2025pangu}{,} HierSearch~\cite{tan2025hiersearch}{,} DynaSearcher~\cite{hao2025dynasearcher}{,} R-Search~\cite{zhao2025r}{,} Graph-R1~\cite{endgraph}{,}
                                \\
                                Chain-of-Agents~(CoA)~\citep{li2025chain}{,} Tool-Star~\cite{dong2025tool}{,} ARPO~\cite{dong2025agentic}{,} AEPO~\cite{dong2025agentic2}{.}
                                \textit{etc.}, leaf3, text width=44em]SurveyForge
                            ]
                        ]
                    ]
                    [\ \ \ \ \ \ Evaluation of \\ \ \ \ \ \ Deep Research System~(\S\ref{sec:evaluation}), ver
                        [\ \ \ \ Agentic Information \\  \ \ \ \ Seeking (\S\ref{agentic-information-seeking})
                            [\ \ \ \ Complex Queries~(\S\ref{complex-queries})
                                [\eg ~NQ~\citep{kwiatkowski2019natural}{,} TriviaQA~\citep{joshi2017triviaqa}{,} SimpleQA~\citep{wei2024measuring}{,} HotpotQA~\citep{yang2018hotpotqa}{,} 2WikiMultihopQA~\citep{ho2020constructing}{,} 
                                \\
                                Bamboogle~\citep{press2023measuring}{,} MultiHop-RAG~\citep{tang2024multihop}{,} MuSiQue~\citep{trivedi2022musique}{,} FRAMES~\citep{krishna2025fact}{,} GPQA~\citep{rein2024gpqa}{,} 
                                \\
                                GAIA~\citep{mialon2023gaia}{,} HLE~\citep{phan2025humanity}{.}                                 \textit{etc.}, leaf2, text width=44em]SurveyForge
                            ]
                            [Interaction Environment~(\S\ref{interation-environment})
                                [\eg ~InfoDeepSeek~\citep{xi2025infodeepseek}{,} AssistantBench~\citep{yoran2024assistantbench}{,} Mind2Web~\citep{deng2023mind2web}{,} Mind2Web 2~\citep{gou2025mind2web}{,} 
                                \\
                                BrowseComp~\citep{wei2025browsecomp}{,} BrowseComp-Plus~\citep{chen2025browsecomp}{,} DeepResearchGym~\citep{coelho2025deepresearchgym}{,} WebArena~\citep{zhouwebarena}{,} 
                                \\
                                WebWalkerQA~\citep{wu2025webwalker}{,} WideSearch~\citep{wong2025widesearch}{,} MMInA~\citep{tian2024mmina}{.}                                 \textit{etc.}, leaf2, text width=44em]SurveyForge
                            ]
                        ]
                        [\ \ Comprehensive Report \\ \ \ \  Generation (\S\ref{comprehensive-report-generation})
                            [\ \ \ \ Survey Generation~(\S\ref{survey-generation})
                                [\eg ~AutoSurvey~\citep{wang2024autosurvey}{,} ReportBench~\citep{li2025reportbenchevaluatingdeepresearch}{,} SurveyGen~\citep{bao2025surveygenqualityawarescientificsurvey}{.}                                 \textit{etc.}, leaf2, text width=44em]SurveyForge
                            ]
                            [\ \ \ \ \ \ \ \ Long-Form Report \\ \ \ \ \ \ \ \ Generation~(\S\ref{long-form-report-generation})
                                [\eg ~Deep Research Comparator~\citep{chandrahasan2025deep}{,} DeepResearch Bench~\citep{du2025deepresearch}{,} ResearcherBench~\citep{xu2025researcherbench}{,} 
                                \\
                                LiveDRBench~\citep{java2025characterizing}{,} PROXYQA~\citep{tan2024proxyqa}{,} SCHOLARQABENCH~\citep{asai2024openscholar}{.}                                 \textit{etc.}, leaf2, text width=44em]SurveyForge
                            ]
                            [\ \ \ \ Poster Generation~(\S\ref{poster-generation})
                                [\eg ~Paper2Poster~\citep{pang2025paper2poster}{,} PosterGen~\citep{zhang2025postergen}{,} P2PInstruct~\citep{sun2025p2p}{.}                                 \textit{etc.}, leaf2, text width=44em]SurveyForge
                            ]
                            [\ \ \ \ Slides Generation~(\S\ref{slides-generation})
                                [\eg ~Doc2PPT~\citep{fu2022doc2ppt}{,} SLIDESBENCH~\citep{ge2025autopresent}{,} Zenodo10K~\citep{zheng2025pptagent}{,} PPTEval~\citep{zheng2025pptagent}{,} TSBench~\citep{jung2025talk}{.}                                 \textit{etc.}, leaf2, text width=44em]SurveyForge
                            ]
                        ]
                        [\ \ AI for Research (\S\ref{sec:ai-for-research})
                            [\ \ \ \ \ Idea Generation~(\S\ref{idea-generation})
                                [\eg ~AI-Researcher~\citep{si2024can}{,} Learn2Gen~\citep{li2024learning}{,} TheAIScientist~\citep{lu2024ai}{,} Virtual-Scientists~\citep{su2024two}{,} 
                                \\
                                AI Idea Bench 2025~\citep{qiu2025ai}{,} RND~\citep{wang2025enabling}{,} GoAI~\citep{gao2025graph}{.}                                 \textit{etc.}, leaf2, text width=44em]SurveyForge
                            ]
                            [Experimental Execution~(\S\ref{experimental-execution})
                                [\eg ~TheAIScientist~\citep{lu2024ai}{,} DeepScientist~\citep{weng2025deepscientist}{,} AI-Researcher~\citep{tang2025ai}{,} PaperBench~\citep{starace2025paperbench}{.}                                 \textit{etc.}, leaf2, text width=44em]SurveyForge
                            ]
                            [\ \ \ \ Academic Writing~(\S\ref{academic-writing})
                                [\eg ~TheAIScientist~\citep{lu2024ai}{,} Automatic-Scientific-Quality-Metrics~\citep{hopner2025automatic}{,} PaperBench~\citep{starace2025paperbench}{,} 
                                \\
                                Scientist-Bench~\citep{tang2025ai}{,} ResearcherBench~\citep{xu2025researcherbench}{.}                                 \textit{etc.}, leaf2, text width=44em]SurveyForge
                            ]
                            [\ \ \ \ \ \ \ \ Peer Review~(\S\ref{peer-reviewing})
                                [\eg ~ASAP-Review~\citep{yuan2022can}{,} TheAIScientist~\citep{lu2024ai}{,} REVIEW-5k~\citep{weng2024cycleresearcher}{,} REVIEWER2~\citep{gao2024reviewer2}{,} 
                                \\ DeepReview-Bench~\citep{zhu-etal-2025-deepreview}{.}                                 \textit{etc.}, leaf2, text width=44em]SurveyForge
                            ]
                        ]
                    ]
                ]
                \end{forest}
        }
        \caption{Taxonomy of the main content of this survey.}\label{fig:taxonomy}
\end{figure*}

%% file: table/comparison.tex
\begin{table*}[!t]
\small
\centering
\setlength{\tabcolsep}{2mm}{
\begin{tabular}{@{}l c  ccc@{}}
\toprule
\textbf{Capability} & \textbf{Standard} & \textbf{Agentic} & \textbf{Integrated} & \textbf{Full-stack} \\ 
\textbf{(Key Feature)}& \textbf{RAG} & \textbf{Search} & \textbf{Research} & \textbf{AI Scientist} \\ 

\midrule

\rowcolor{gray!15} Search Engine Access & \ding{51} & \ding{51} & \ding{51} & \ding{51} \\
\rowcolor{gray!15} Use of Various Tools (e.g., Web APIs) & \ding{55} & \ding{51} & \ding{51} & \ding{51} \\
\rowcolor{gray!15} Code Execution for Experiment & \ding{55} & \ding{55} & \ding{55} & \ding{51} \\

\rowcolor{blue!12} Reflection for Action Correction & \ding{55} & \ding{51} & \ding{51} & \ding{51} \\
\rowcolor{blue!12} Task-solving Memory Management & \ding{55} & \ding{51} & \ding{51} & \ding{51} \\
\rowcolor{green!12} Innovation and Hypothesis Proposal & \ding{55} & \ding{55} & \ding{55} & \ding{51} \\
\rowcolor{green!12} Long-form Answer Generation \& Validation  & \ding{51} & \ding{55} & \ding{51} & \ding{51} \\

\specialrule{0em}{1pt}{1pt}
\cdashline{1-5}[6pt/6pt]
\specialrule{0em}{1pt}{1pt}

\rowcolor{gray!15} Action Space & Narrow & Broad & Broad & Broad \\ 
\rowcolor{blue!12} Reasoning Horizon & Single & Long-horizon & Long-horizon & Long-horizon \\
\rowcolor{blue!12} Workflow Organization & Fixed & Flexible & Flexible & Flexible \\ 
\rowcolor{green!12} Output Form and Application & Short Span  & Short Span  & Report  & Academic Paper \\
\bottomrule
\end{tabular}
}
\caption{Comparison between conventional RAG (leftmost column) and the three envisioned stages of Deep Research (right columns). The capabilities evolve from static retrieval and generation to adaptive, autonomous, and scientifically creative workflows.
}
\label{tab:comparison}
\end{table*}


%% file: sections/03-component.tex
\section{Key Components in Deep Research System}\label{sec:components}

A DR system can be viewed as a closed-loop workflow that takes a complex research question as input and produces a structured answer, typically in the form of long-form text with citations or synthesized reports. As illustrated in Figure~\ref{fig:intro}, the DR system iteratively cycles through a set of interconnected components:
(i) \textit{query planning}, which decomposes the original question into sub-queries and tool calls that guide the workflow;
(ii) \textit{knowledge acquisition}, which retrieves and filters relevant information from external corpora, tools, or APIs;
(iii) \textit{memory management}, which stores, updates, and prunes intermediate findings to maintain context over long horizons; and
(iv) \textit{answer generation}, which synthesizes the accumulated evidence into a coherent, verifiable response with citations and checks for consistency.
In this work, we provide detailed definitions and functionality for each component, along with representative works.

\subsection{Query Planning}\label{query-planning}

\begin{figure*}[!t]
    \centering
    \includegraphics[width=\textwidth]{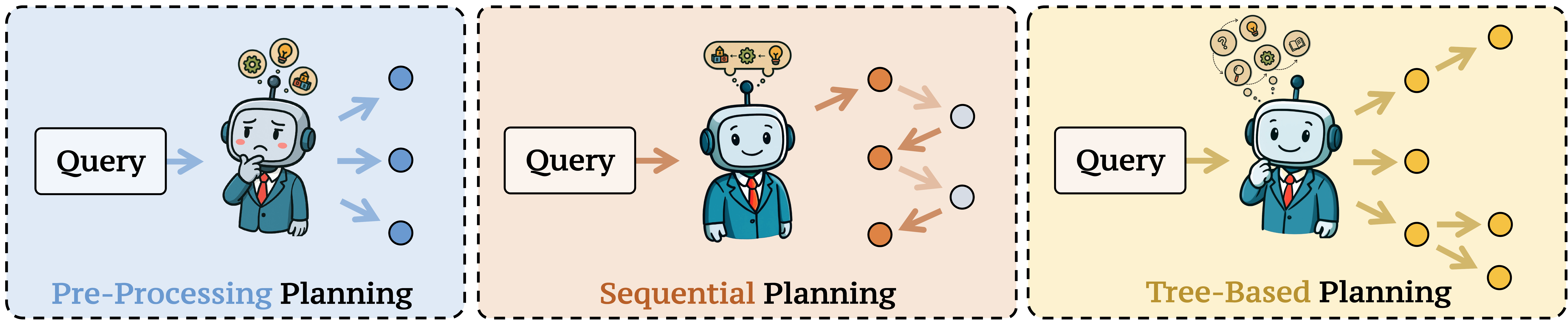}
    \caption{Three commonly-used types of query planning: (i) parallel planning; (ii) sequential planning; and (iii) tree-based planning.}
    \label{planner}
\end{figure*}

\textit{Query Planning} refers to the process of transforming a complex and logically intricate question into a structured sequence of executable sub-queries (\textit{aka.,} sub-tasks), each of which can be addressed incrementally. 
This decomposition allows stepwise reasoning and knowledge acquisition, thereby enhancing the reliability and accuracy of the final output generated by \drs.

Figure~\ref{planner} shows three widely-used strategies for query planning:
(i) \textit{parallel planning}, which decomposes the input into independent sub-queries that may be resolved in parallel~\cite{chen2025improving,dhuliawala2023chain}; 
(ii) \textit{sequential planning}, which arranges sub-queries into a linear order where each step depends on intermediate outcomes~\cite{shi-etal-2024-generate,jin2025search}; and 
(iii) \textit{tree-based planning}, which explores branching decision spaces and selects among candidate paths through pruning, backtracking, or heuristic-guided search~\cite{yao2023tree}.

\subsubsection{Parallel Planning}\label{sec:parallel-planning}

\header{Definition.} As illustrated in Figure~\ref{planner}(a), parallel planning operates by rewriting or decomposing the original query into multiple sub-questions in a single pass, typically without iterative interaction with downstream components. The primary advantage of this strategy lies in its efficiency: simultaneous generation enables parallel processing of sub-queries.

\header{Representative Work.} Early research typically instantiates parallel planning modules through heuristic approaches, most notably via prompt engineering~\cite{zhou2022least,dhuliawala2023chain} or training on manually annotated datasets. 
For example, Least-to-Most Prompting~\cite{zhou2022least} guides GPT-3~\cite{brown2020language} to decompose a complex task into an ordered sequence of simpler, self-contained sub-queries in a few-shot setting.
Similarly, CoVE~\cite{dhuliawala2023chain} prompts LLMs to first generate multiple independent sub-questions and then ground each one with well-established evidence in parallel, a strategy widely adopted in knowledge-intensive applications.


Despite these advancements, query planning based on general heuristics or task-agnostic supervision often suffers from misalignment with end-to-end objectives in downstream applications, particularly in complex QA scenarios~\cite{yang2018hotpotqa,ho2020constructing,trivedi2022musique,press2022measuring}.
To mitigate this issue, recent work has turned to end-to-end planning optimization via RL. 
For example, the Rewrite-Retrieve-Read framework~\cite{ma2023query} trains a query planner to maximize final answer accuracy using the Proximal Policy Optimization algorithm~\cite{schulman2017proximal}.
Crucially, the planner is reinforced only when documents retrieved by its sub-queries enable an LLM to generate a correct answer, which replaces reliance on heuristic decomposition rules.
Building on this approach, subsequent efforts such as DeepRetrieval~\cite{jiang2025deepretrieval} and CardRewriter~\cite{gong2025cardrewriter} have extended reward modeling for query planners to incorporate diverse downstream metrics (\eg evidence recall, retrieval NDCG@k).
More recently, studies have also explored jointly optimizing query planning with other components in modular dense retrieval pipelines through multi-agent RL methods~\cite{chen2025improving}.

\header{Advantages \& Disadvantages.}
Despite their efficiency, parallel planning has two primary limitations. 
First, they typically operate in a \textit{one-shot} fashion, interacting with other modules (\eg retriever, reasoner, aggregator) non-iteratively. As a result, they lack mechanisms to incorporate intermediate evidence, correct earlier decisions, or adaptively allocate computational resources. Second, they often \textit{ignore data and logical dependencies} across sub-queries. Parallel execution assumes conditional independence, yet many real-world queries involve sequential reasoning in which later subtasks depend on the resolution of earlier ones. This can result in ill-posed or unanswerable sub-queries due to missing contextual information.

\subsubsection{Sequential Planning}\label{sequential-planning}

\header{Definition.} As illustrated in Figure~\ref{planner}(b), the sequential planning decomposes the original query through multiple iterative steps, where each round of decomposition builds upon the outputs of previous rounds. 
At each stage, the sequential planning may invoke different modules or external tools to process intermediate results, enabling a dynamic, feedback-driven reasoning process. 
This multi-turn interaction allows the sequential planning to perform logically dependent query decompositions that are often intractable for pre-processing planning, which typically assumes conditional independence among sub-queries. 
By incorporating intermediate evidence and adapting the query trajectory accordingly, sequential planning is particularly well-suited for complex tasks that require stepwise inference, disambiguation, or progressive information gathering.

\header{Representative Work.} The sequential planning is often used to provide a series of sub-queries for the external knowledge needed in a step-by-step manner, which has been widely used in iterative QA systems~\citep{li2023llatrieval,su2024dragin,jiang2025retrieve}.
For example, LLatrieval~\cite{li2023llatrieval} introduces an iterative query planner that, whenever the current documents fail verification, leverages the LLM to pinpoint missing knowledge and generate a new query, either a question or a pseudo-passage, to retrieve supplementary evidence, repeating the cycle until the accumulated context fully supports a verifiable answer.
DRAGIN~\citep{su2024dragin} introduces a query planner that can utilize the self-attention scores to select the most context-relevant tokens from the entire generation history and reformulate them into a concise and focused query.
This dynamic, attention-driven approach produces more accurate queries compared to the static \textit{last sentence} or \textit{last n tokens} strategies in previous methods, resulting in higher-quality retrieved knowledge and improved downstream generation. 
In ReSP~\citep{jiang2025retrieve}, the query planner dynamically guides each retrieval iteration by formulating novel sub-questions explicitly targeted at identified information gaps whenever the currently accumulated evidence is deemed insufficient. 
By conditioning this reformulation process on both global and local memory states and by disallowing previously issued sub-questions, the approach mitigates the risks of over‑planning and redundant retrieval. This design ensures that each newly generated query substantially contributes to advancing the multi‑hop reasoning trajectory toward the final answer.
RAISE~\cite{oh2025raise} sequentially decomposes a scientific question into sub-problems, generates logic-aware queries for each, and retrieves step-specific knowledge to drive planning and reasoning.
Additionally, S3~\citep{jiang2025s3} and AI‑SearchPlanner~\citep{mei2025ai} both adopt sequential decision‑making to control when and how to propose retrieval queries during multi‑turn search. At each turn, the sequential planner evaluates the evolving evidence state and decides whether to retrieve additional context or to stop.
Besides, more recent studies, including Search-R1~\citep{jin2025search}, R1-Searcher~\citep{song2025r1,song2025r1++} integrate a sequential planning strategy into an end-to-end, multi-turn search framework, thereby leveraging LLMs' internal reasoning for query planning.

\header{Advantages \& Disadvantages.}
Sequential planning enables dynamic, context-aware reasoning and fine-grained query reformulation, thereby facilitating more accurate acquisition of external knowledge.
However, excessive reasoning turns or overly long reasoning chains can incur substantial computational costs and latency. In addition, an increased number of turns may introduce cumulative noise and error propagation, potentially causing instability during reinforcement learning training.

\subsubsection{Tree-based Planning}\label{tree-based-planning}

\header{Definition.} As illustrated in Figure~\ref{planner}(c), the tree-based planning integrates features of both parallel and sequential planning by recursively treating each sub-query as a node within a structured search space, typically represented as a tree or a directed acyclic graph (DAG)~\cite{cormen2022introduction}. 
This structure enables the use of advanced search algorithms, such as Monte Carlo Tree Search (MCTS)~\cite{browne2012survey}, to explore and refine potential reasoning paths. Compared to linear or flat decompositions, this approach supports more flexible and fine-grained decomposition of the original query, facilitating comprehensive knowledge acquisition.

\header{Representative Work.} 
A representative example is RAG-Star~\cite{jiang2024rag}, which leverages MCTS in conjunction with the Upper Confidence Bound for Trees (UCT)~\cite{kocsis2006bandit} to guide a query planner in the iterative decomposition of complex questions. 
At each iteration, the planning model selects the most promising node using the UCT criterion, expands it by generating a sub-query and corresponding answer using a language model, evaluates the quality of the expansion via a retrieval-based reward model, and back-propagates the resulting score. 
This iterative process grows a reasoning tree of sub-queries until a satisfactory final answer is obtained.
Other examples include DTA~\cite{zhu2025divide} and DeepSieve~\cite{guo2025deepsieve}, which use a tree-based planner to restructure sequential reasoning traces into a DAG. 
This design enables the planning to aggregate intermediate answers along multiple branches and improves the model's ability to capture both hierarchical and non-linear dependencies across sub-tasks.
DeepRAG~\cite{guan2025deeprag} introduces tree-based planning via binary-tree exploration to iteratively decompose queries and decide parametric vs. retrieved reasoning, yielding large accuracy gains with fewer retrievals.
More recently, MAO-ARAG~\cite{chen2025mao} trains a planning agent that can dynamically orchestrate multiple, diverse query reformulation modules through a DAG structure. 
This adaptive workflow enables comprehensive query decomposition to enhance performance.

\header{Advantages \& Disadvantages.}
Tree-based planning integrates the strengths of parallel and sequential planning. It facilitates the decomposition of interdependent sub-queries and supports local parallel execution, striking an effective balance between efficiency and effectiveness. Nevertheless, training a robust Tree-based Planning module is challenging, requiring precise dependency modeling, careful trade-offs between speed and quality, addressing data scarcity, and tackling credit assignment issues in reinforcement learning.

\begin{tcolorbox}[takeawaysbox, title={Takeaway}]
    This section on query planning provides a detailed overview of strategies for enhancing DR systems by decomposing complex queries into simpler, manageable subtasks. Each type of planning strategy offers unique benefits and faces specific challenges.
    \begin{itemize}
        \item  \textit{Pre-processing planning} is efficient in executing sub-queries simultaneously, though they may overlook dependencies between them. 
    
        \item \textit{Sequential planning} excels in managing dependencies through iterative processes but can incur higher computational costs.
    
        \item \textit{Tree-based planning} strikes a balance by combining the strengths of both sequential and pre-processing approaches, allowing for adaptive and flexible query decomposition. 
    \end{itemize}
\end{tcolorbox}


\subsection{Information Acquisition}\label{information-acquision}

DR systems often acquire external information to augment LLMs' internal knowledge. 
However, due to the cost of retrieval and the uncertainty of document quality, it is necessary to determine when retrieval is needed~\cite{zhang2025multi,wu-etal-2025-search,zhang2024agentic}. 
Moreover, how to perform retrieval and manage retrieved information is key to the DR system’s interaction with external knowledge.
In the following, we discuss retrieval tools, retrieval timing, and information filtering in turn.

\subsubsection{Retrieval Tools}\label{knowledge-retrieval}

\header{Definition.} In the context of DR, \textit{retrieval tools}~\cite{information_retreival1,information_retreival2,information_retreival3} are used to identify relevant information from large-scale corpora in response to a query, typically containing indexing and search techniques.
Within typical DR workflows, retrieval serves as a core mechanism for bridging knowledge gaps by surfacing candidate evidence that can then be checked for accuracy, filtered for relevance, or combined into a coherent answer.
Below, we systematically review widely adopted retrieval techniques, organized by modality: (i) \textit{text-only retrieval}, and (ii) \textit{multimodal retrieval}.

\header{Text Retrieval.} Conceptually, modern text retrieval can be organized into three families: (i) lexical retrieval, (ii) semantic retrieval, and (iii) commercial web search.
Lexical and semantic retrieval are typically implemented on local resources, while commercial web search is typically accessed only via paid APIs. Specifically, \textit{lexical retrieval} refers to methods that match documents based on exact term overlaps and statistical term weighting, including traditional approaches like TF-IDF and BM25~\cite{robertson2009bm25}, as well as neural sparse models that learn to expand queries and documents with relevant terms while maintaining interpretable inverted-index structures~\cite{formal2021splade,formal2021splade2, khattab2020colbert, karpukhin2020dense, qu2020rocketqa, wang2022text, formal2021splade, santhanam2021colbertv2}.

Different from the lexical retrieval, \textit{semantic retrieval} refers to dense neural methods that encode queries and documents into continuous vector spaces to capture semantic similarity beyond exact term matching~\cite{shi2025search, he2025self, shi2023towards, jin2025empirical}, which has been widely adopted in recent works~\cite{jin2025search, shi2025iterative}.

More recently, \textit{commercial web search} (like Google or Bing) has also been widely used in DR systems and web agents~\cite{tao2025webleaper,fang2025towards,ou2025browseconf,chen2025agentfrontier}.
It diverges from lexical and semantic retrieval models by providing access to real-time information, leveraging massive-scale web crawling and indexing, incorporating sophisticated ranking algorithms that consider authority and freshness signals, and offering built-in fact verification through cross-source validation.
Previous work, such as WebGPT~\cite{nakano2021webgpt} and SearchGPT~\cite{xu2023chatgpt}, demonstrates that commercial search APIs enable research agents to access current events and dynamic content that would be missing from static corpora. 

Recent studies~\cite{li2025search, li2025webthinker, tang2025agent, qiao2025webresearcher} exemplify a shift towards more autonomous and capable research agents. These models feature deep web exploration capabilities, allowing them to interactively navigate beyond static search results to gather information.
Overall, the evolution from lexical and semantic retrieval to commercial web search marks a shift from static, closed-corpus search toward dynamic, real-world information access, enabling DR systems to retrieve not only relevant but also timely and verifiable knowledge.

\header{Multimodal Retrieval.} Multimodal retrieval aims to mine multimodal information, including text, layout, and visuals (figures, tables, charts), and to preserve grounded pointers (spans, cells, coordinates) for verifiable citation, while maximizing recall under tight latency to support iterative DR. Multimodal information retrieval can be organized into three classes based on the primary type of information modality being indexed and retrieved: (i)\textit{text-aware retrieval with layout}, which indexes titles, captions, callouts, and surrounding prose and leverages document understanding models (LayoutLM~\cite{xu2020layoutlm}, Donut~\cite{kim2022donut}, DocVQA~\cite{mathew2021docvqa}) plus layout/metadata filters; (ii) \textit{visual retrieval via text–image similarity}, which encodes figures and chart thumbnails with CLIP~\cite{radford2021clip}, SigLIP~\cite{zhai2023siglip}, or BLIP~\cite{li2022blip} and performs ANN search for text-to-image matching or composed image retrieval~\cite{yang2023composed}; and (iii) \textit{structure-aware retrieval over parsed tables and charts}, which indexes axes, legends, data marks, and table schemas to support grounded lookup of numeric facts and relations (\eg  ChartReader~\cite{cheng2023chartreader} or Chartformer~\cite{zheng2024chartformer}). These three approaches are typically combined: queries are searched across all indices simultaneously, with results fused using reciprocal-rank fusion~\cite{cormack2009rrf} or cross-modal reranking to preserve grounded pointers for citations. Recent chart-focused VLMs~\cite{meng2024chartassistant,carbune2024chartpali,zhang2024tinychart,masry2023unichart} further enhance the quality of visual-textual features.

\header{Comparing Text Retrieval and Multimodal Retrieval.}
Compared to text-only retrieval, multimodal retrieval provides several key advantages. First, it captures visually encoded information and numeric trends that text-based methods often overlook, and facilitates cross-modal verification through hybrid fusion~\cite{cormack2009rrf}. Second, it enables grounded citations using techniques such as layout parsing (\eg  LayoutLM~\cite{xu2020layoutlm}, Donut~\cite{kim2022donut}) and chart understanding (\eg  ChartReader~\cite{cheng2023chartreader} or Chartformer~\cite{zheng2024chartformer}). 
However, multimodal retrieval also presents several challenges, including increased computational costs for visual processing~\cite{radford2021clip,zhai2023siglip}, sensitivity to OCR errors and variations in chart formats~\cite{tang2025chartmuseum,xia2024chartx}, and the complexity of aligning information across different modalities.

\begin{tcolorbox}[takeawaysbox, title={Takeaway}]
Knowledge retrieval for DR has evolved from traditional lexical and dense-text search to the use of real-time commercial web search engines for up-to-date information. However, text-only methods fail to capture information embedded in visual elements like charts, tables, and layouts. Multimodal retrieval addresses this gap by modeling visual and structural data. Its primary contribution is enabling grounded, verifiable citations by linking retrieved evidence back to specific data points (\eg  table cells or chart coordinates), though this introduces higher computational costs and challenges in cross-modal alignment and format processing.
\end{tcolorbox}

\begin{figure}[t]
    \centering
    \includegraphics[width=0.97\textwidth]{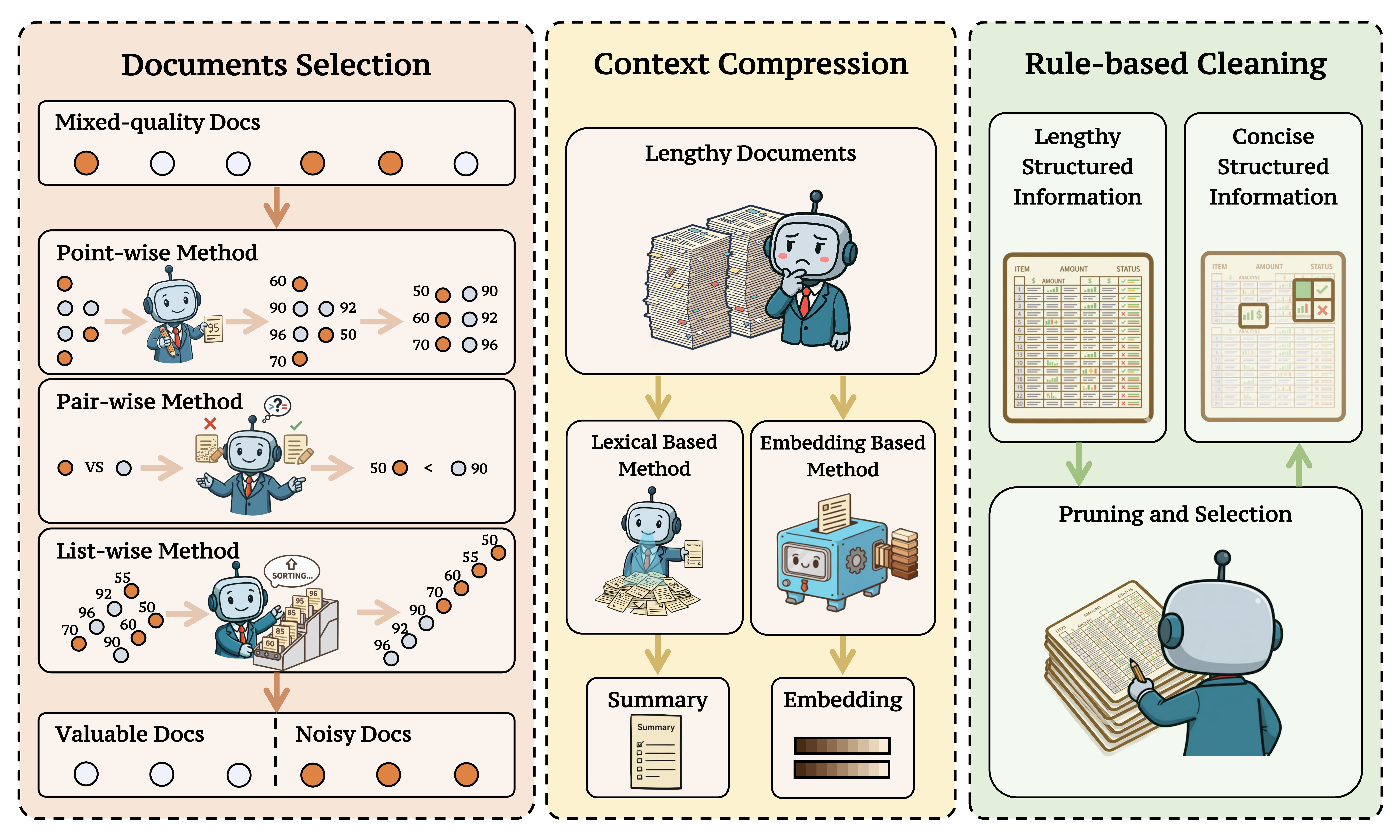}
    \caption{Existing information filtering approaches can be broadly categorized into the following types: (i) \textit{Document Selection}; (ii) \textit{Context Compression};  and (iii) \textit{Rule-based Cleaning}.}
    \label{fig:knowledge filtering}
\end{figure}

\subsubsection{Retrieval Timing}\label{retrieval-timing}

\header{Definition.}
Retrieval timing refers to determining when a model should trigger retrieval tools during information seeking, which is also known as adaptive retrieval~\cite{jeong2024adaptive,yao2024seakr,ding2024retrieve}.
Because the quality of retrieved documents is not guaranteed, blindly performing retrieval at every step is often suboptimal~\cite{mallen-etal-2023-trust,zhao2024towards,shi2025iterative}. 
Retrieval introduces additional computational overhead, and low-quality or irrelevant documents may even mislead the model or degrade its reasoning performance~\cite{shi-etal-2024-generate}.
Consequently, adaptive retrieval aims to invoke retrieval only when the model lacks sufficient knowledge, which requires the model to recognize its own knowledge boundaries~\cite{li2025knowledge,zhang2025self,xiao2025analyzing,ren2025investigating}, \ie knowing what it knows and what it does not.

Prior work on adaptive retrieval follows two main directions: (i) estimating and enhancing a model’s ability to \textit{recognize its own knowledge boundaries} for a given query, and (ii) optimizing the \textit{retrieval-trigger mode}l in multi-step settings to maximize downstream task performance.

\header{Confidence Estimation as a Proxy for Boundary Perception.}
There are extensive works that investigate LLMs' perception of their knowledge boundaries. The degree to which a model perceives its boundaries is typically measured by the alignment between its confidence and factual correctness. Since factual correctness is typically evaluated by comparing the model’s generated answer with the ground-truth answer, existing studies focus on how to measure the model’s confidence, which can be broadly divided into four categories.

\begin{itemize}
    \item \textit{Probabilistic Confidence}. This line of work treats a model’s token-level generation probabilities as its confidence in the answer~\citep{guo2017calibration, desai2020calibration, jiang2021can, kadavath2022language, si2022prompting, kuhn2023semantic, duan2023shifting}. Prior to the emergence of LLMs, a line of work had already shown that neural networks tend to be poorly calibrated, often producing overconfident predictions even when incorrect~\citep{guo2017calibration,desai2020calibration,jiang2021can}.
    More recently, some research\citep{kadavath2022language, si2022prompting} reported that LLMs can be well calibrated on structured tasks such as multi-choice question answering or appropriate prompts, but for open-ended generation tasks, predicted probabilities still diverge from actual correctness. To address this gap, \citet{duan2023shifting} proposed SAR, which computes confidence by focusing on important tokens, while \citet{kuhn2023semantic} introduced semantic uncertainty, which estimates confidence from the consistency of outputs across multiple generations. 

    \item \textit{Consistency-based Confidence}. Since probabilistic confidence often fails to capture a model’s semantic certainty and is inapplicable to black-box models without accessible generation probabilities, recent works represent confidence via semantic consistency across multiple responses~\citep{fomicheva2020unsupervised,manakul2023selfcheckgpt,kuhn2023semantic,zhang2023sac3,ding2024retrieve}. The key idea is that a confident model should generate highly consistent answers across runs. ~\citet{fomicheva2020unsupervised} first measured consistency through lexical similarity, while later studies used NLI (\ie natural language inference) models or LLMs to assess semantic consistency~\citep{manakul2023selfcheckgpt,kuhn2023semantic}. To address the issue of consistent but incorrect answers, ~\citet{zhang2023sac3} measure consistency across different models, as incorrect answers tend to vary between models, whereas correct ones align. ~\citet{ding2024retrieve} further extended this idea to multilingual settings.

    \item \textit{Confidence Estimation Based on Internal States}. LLMs' internal states have been shown to capture the factuality of their generated content~\citep{azaria2023internal,su2024unsupervised,chen2024inside,wang2024hidden,ni2025towards,ni2025annotation}. \citet{azaria2023internal} first discovered that internal states can signal models’ judgment of textual factuality. Subsequent studies~\citep{su2024unsupervised,chen2024inside} found that internal states after response generation reflect the factuality of self-produced answers. More recently, \citet{wang2024hidden} and \citet{ni2025towards} demonstrated that factuality-related signals already exist in the pre-generation states, enabling the prediction of whether the output will be correct.

    \item \textit{Verbalized Confidence}. Several studies explore enabling LLMs to express confidence in natural language, akin to humans, viewing such verbalization as a sign of intelligence~\citep{lin2022teaching,yin2023large,tian2023just,xiong2023can,zhang2024r,yang2023alignment,ni2024llms}. \citet{yin2023large} and \citet{ni2024llms} examined whether LLMs can identify unanswerable questions, finding partial ability but persistent overconfidence. Other works~\citep{tian2023just,xiong2023can} investigated fine-grained confidence expression. \citet{xiong2023can} offered the first comprehensive study for black-box models, while \citet{tian2023just} proposed generating multiple answers per pass for more accurate estimation. Beyond prompting, some methods explicitly train models to verbalize confidence~\citep{lin2022teaching,yang2023alignment,zhang2024r}, with \citet{lin2022teaching} introducing this idea and using correctness-based supervision.
\end{itemize}


\header{Representative Adaptive Retrieval Approaches.}
Deep research systems typically involve iterative interactions between model inference and external document retrieval, differing mainly in how they determine when to retrieve. Early works such as IR-CoT~\citep{trivedi2022interleaving} enforce retrieval after every reasoning step, ensuring continual grounding in external knowledge but at the cost of efficiency.
Building on insights from studies of models’ perceptions of their own knowledge boundaries, recent approaches treat retrieval as a model-issued action, enabling the model to perform it dynamically only when needed. Similar to techniques in confidence estimation, these methods assess whether the model can answer a question correctly given the current context and perform retrieval when knowledge is deemed insufficient. They can be broadly categorized into four paradigms.
\begin{itemize}
    \item \textit{Probabilistic Strategy}. It triggers retrieval based on token-generation probabilities: when the model produces a token with low confidence, retrieval is initiated~\citep{jiang2023active,su2024dragin}.

    \item \textit{Consistency-based Strategy}. Recognizing that both token-level probabilities and single-model self-consistency may fail to capture true semantic uncertainty, Rowen~\citep{ding2024retrieve} evaluates consistency across responses generated by multiple models and languages, triggering retrieval when cross-model or cross-lingual agreement is low.

    \item \textit{Internal States Probing}. CtrlA~\citep{liu2024ctrla}, UAR~\citep{cheng2024unified}, and SEAKR~\citep{yao2024seakr} further propose that compared to generated responses, a model’s internal states provide a more faithful reflection of its confidence, using them to guide adaptive retrieval decisions.

    \item \textit{Verbalized Strategy}. It enables the model to directly express its confidence via natural language.
    These methods typically generate special tokens directly in the response to indicate the need for retrieval. ReAct~\citep{yao2022react} directly prompts the model to generate corresponding action text when retrieval is needed. Self-RAG~\citep{asai2024self} trains the model to explicitly express uncertainty through the special token (\ie \texttt{<retrieve>}), signaling the need for retrieval.
    With LLMs' growing reasoning capacity, recent research has shifted toward determining retrieval timing through reasoning and reflection. Search-o1~\citep{li2025search} introduces a Reason-in-Documents module, which prompts the model to selectively invoke search during reasoning. 
    Search-R1~\citep{jin2025search} further frames retrieval as part of the environment and employs reinforcement learning to jointly optimize both when and what to retrieve.
\end{itemize}
Collectively, these methods trace an evolution from fixed or per-step retrieval (\eg IR-CoT~\cite{trivedi2022interleaving}) to dynamically triggered retrieval (\eg ReAct~\cite{yao2022react}, Self-RAG~\cite{asai2024self}, Search-o1~\cite{li2025search}), and finally to RL–based systems that explicitly train retrieval policies (\eg Search-R1~\cite{jin2025search}).

\subsubsection{Information Filtering}\label{information-filtering}

\header{Definition.}
Information filtering refers to the process of selecting, refining, or transforming retrieved documents so that only the most relevant and reliable evidence is passed to subsequent steps.
Since retrieval tools are not perfect, the retrieved information often contains considerable noise~\cite{yoran2023making, wu2024pa, fang2024enhancing}.
This includes the content that is entirely irrelevant to the query or plausible-looking statements that nevertheless provide incorrect or misleading context.
As shown in prior work~\cite{yoran2023making, jinlong}, LLMs are highly sensitive to such noise; without additional filtering or optimization, they can be easily misled into generating incorrect or hallucinated responses.
Figure~\ref{fig:knowledge filtering} summarizes three information filtering approaches:
(i) \textit{Document Selection},
(ii) \textit{Context Compression}, and
(iii) \textit{Rule-based Cleaning}.

\header{Document Selection.}
Document selection aims to rank a set of candidate documents based on their relevance and usefulness to the query,  selecting the top-k helpful documents for question answering~\cite{xu2023recomp,yu2024rankrag,wei2024instructrag}. 
This selection operation reduces the impact of noisy documents on LLMs, improving the question-answering accuracy in downstream tasks.
Below, we review three document selection strategies: \textit{point-wise} selection, \textit{pair-wise} selection, and \textit{list-wise} selection.
\begin{itemize}[leftmargin=*,nosep]
    \item \textit{Point-wise Selection}. Given an initially retrieved document list, \textbf{point-wise} methods independently score each candidate document. The most common approach involves fine-tuning an embedding model (\eg BGE~\cite{xiao2024c}) that encodes the query and each document separately, after which their relevance is estimated via inner-product similarity~\cite{xu2023recomp,izacard2021unsupervised}. Another widely adopted strategy employs a cross encoder, which takes the concatenation of the query and a document as input and directly predicts a binary relevance score~\cite{dong2025understand,wang2024rear}. More recently, several studies have leveraged LLMs' natural language understanding capabilities for relevance assessment. These methods train LLMs to output special tokens, such as \texttt{<ISREL>}~\cite{asai2024self} or the identifier \texttt{True}~\cite{yu2024rankrag}, to indicate whether an input document is relevant to the query.
    

   \item \textit{Pair-wise Selection}. Unlike the point-wise approach, which assigns an absolute relevance score, the pair-wise method compares the relevance of two input candidate information snippets (typically two documents) and predicts which one is more relevant to the query.
    Pair-wise selection is less common than point-wise selection. 
    A representative work is PRP~\cite{qin2023large}, which adopts a pairwise-ranking-prompting approach.
    In PRP, the LLM receives a query and two candidate documents to decide which is more relevant, and the final ranking list is then obtained using a heapsort algorithm. To mitigate positional bias, PRP performs the comparison twice, swapping the document order each time, and aggregates the results to yield a more stable judgment.

    \item \textit{List-wise methods}.Given a document list, a list-wise selection strategy directly selects the final set of relevant documents from the candidate list. 
    A representative work is RankGPT~\citep{sun2023chatgpt}, which feeds the entire candidate sequence into an LLM and leverages prompt engineering to produce a global ranking. 
    In addition to RankGPT, other work, such as TourRank~\citep{chen2025tourrank}, uses a tournament-inspired strategy to generate a robust ranking list~\citep{chen2025tourrank, yoon2024listt5}.
    ListT5~\citep{yoon2024listt5} proposes a list re-ranking method based on the Fusion-in-Decoder (FiD)~\citep{izacard2020leveraging} architecture, which independently encodes multiple documents in parallel and orders them by relevance, mitigating positional sensitivity while preserving efficiency. For large document sets, it builds m-ary tournament trees to group, rank, and merge results in parallel.
    Recently, more and more work has employed the reasoning model for list-wise document selection, advancing document selection by explicitly modeling a chain of thought.
    For example, InstructRAG~\citep{wei2024instructrag} trains an LLM to generate detailed rationales via instruction tuning~\citep{shengyu2023instruction}, directly judging the usefulness of each document in the raw retrieved document list.
    Rank-R1~\citep{zhuang2025rank} employs the reinforcement learning algorithm GRPO~\citep{shao2024deepseekmath} to train the LLM, enabling it to learn how to select the documents most relevant to a query from a list of candidates. 
    ReasonRank~\citep{liu2025reasonrank} empowers a list-wise selection model through a proposed multi-view ranking-based GRPO~\citep{shao2024deepseekmath}, training an LLM on automatically synthesized multi-domain training data. 
\end{itemize}

\header{Content Compression.}
Content Compression aims to remove redundant or irrelevant information from retrieved knowledge, thereby increasing the density of useful content within the model's context. 
Existing approaches primarily fall into two categories: \textit{lexical-based} and \textit{embedding-based} methods.

\begin{itemize}[leftmargin=*, nosep]
    \item \textbf{Lexical-based methods} condense retrieved text into concise natural language, aiming to only include the key point related to the given query~\cite{xu2023recomp,wang2023learning}.
    Representative works such as RECOMP~\cite{xu2023recomp} fine-tune a smaller, open-source LLM to summarize the input retrieved documents, where the ground truth is synthesized by prompting powerful commercial LLMs like GPT-4~\cite{achiam2023gpt}.
    Chain-of-Note~\cite{yu2023chain} introduces a reading-notes mechanism that compels the model to assess the relevance of retrieved documents to the query and extract the most critical information before generating an answer, with training data annotated by GPT-4~ and further validated through human evaluation.
    Other work, like BIDER~\cite{jin2024bider}, eliminates reliance on external model distillation by synthesizing Key Supporting Evidence (KSE) for each document, using it for compressor SFT, and further optimizing with PPO based on gains in answer correctness.
    \citet{zhu2024information} argue that previous compressors optimized with log-likelihood objectives failed to precisely define the scope of useful information, resulting in residual noise. They proposed a noise-filtering approach grounded in the information bottleneck principle, aiming to maximize the mutual information between the compressed content and the target output while minimizing it between the compressed content and retrieved passages. 
    RankCoT~\cite{wu2025rankcot} implicitly learns document reranking during information refinement. It first employs self-reflection to generate summary candidates for each document. In subsequent DPO~\cite{schulman2017proximal} training, the compression model is encouraged to assign higher probabilities to correct summaries when all documents are fed in, thereby inducing implicit reranking in the final summarization.

    \item \textbf{Embedding-based methods} compress context into dense embedding sequences~\cite{mu2023learning, Guo2025dynamic,chevalier2023adapting}.
    Because embedding sequences can store information flexibly, embedding-based methods can be more efficient and effective than lexical-based methods.
    ICAE~\cite{ge2023context} uses an encoder to compress context into fixed-length embedding sequences and designs training tasks to align the embedding space with the answer generation model.
    COCOM~\cite{rau2024context} jointly fine-tunes the encoder and answer generation model, enhancing the latter’s ability to capture the semantics of embeddings.
    xRAG~\cite{cheng2024xrag} focuses on achieving extreme compression rates.
    It introduces a lightweight bridging module, initialized with a two-layer MLP and trained through paraphrase pretraining and context-aware instruction tuning. This module projects the document embedding vectors originally used for initial retrieval into a single token in the answer generation model's representation space, achieving contextual compression with only a single additional token.
    ACC-RAG~\cite{Guo2025dynamic} adapts compression rates for different documents by employing a hierarchical compressor to produce multi-granularity embedding sequences and dynamically selecting compression rates based on query complexity.
    Similarly, QGC~\cite{cao2024retaining} adjusts compression rates based on query characteristics, dynamically selecting different rates for different documents based on their relevance to the query.
\end{itemize}

\header{Rule-based Cleaning.} 
Rule-based methods are effective for cleaning externally sourced information with specific structures. 
For example, HtmlRAG~\cite{tan2025htmlrag} applies rule-based compression to remove structurally present but semantically empty elements, such as CSS styling and JavaScript code, from retrieved web pages. 
This is combined with a two-stage block-tree pruning strategy that first uses embeddings for coarse pruning, followed by a generative model for fine-grained pruning.
Separately, TableRAG~\cite{chen2024tablerag} accurately extracts core table information through schema retrieval, which identifies key column names and data types, and cell retrieval, which locates high-frequency cell value pairs. This method addresses the challenges of context length limitations and information loss in large table understanding.


\header{Advantages \& Disadvantages.}
Filtering the retrieved knowledge is a simple yet effective strategy to enhance the performance of DR systems, as widely demonstrated in previous work~\cite{xu2023recomp,tan2025htmlrag,dong2025understand}.
However, incorporating an additional filtering module typically incurs additional computational costs and increased latency~\cite{wu2024pa}.
Moreover, overly filtering may remove useful or even correct information, thereby degrading model performance.
Therefore, balancing filtering precision and information retention is crucial for building efficient and reliable DR systems.


\begin{tcolorbox}[takeawaysbox, title={Takeaway}]
Knowledge filtering can further process the metadata retrieved by DR systems, providing them with more concise and useful external knowledge while reducing noise interference and attention dilution caused by long context lengths. Filtering methods can be categorized into post-ranking selection, context compression, and rule-based cleaning. However, these knowledge filtering techniques often introduce additional time and computational costs. Therefore, different DR systems should choose the most suitable filtering method based on task characteristics to balance performance and resource consumption.
\end{tcolorbox}

\begin{figure*}[!t]
    \centering
    \includegraphics[width=1.0\textwidth]{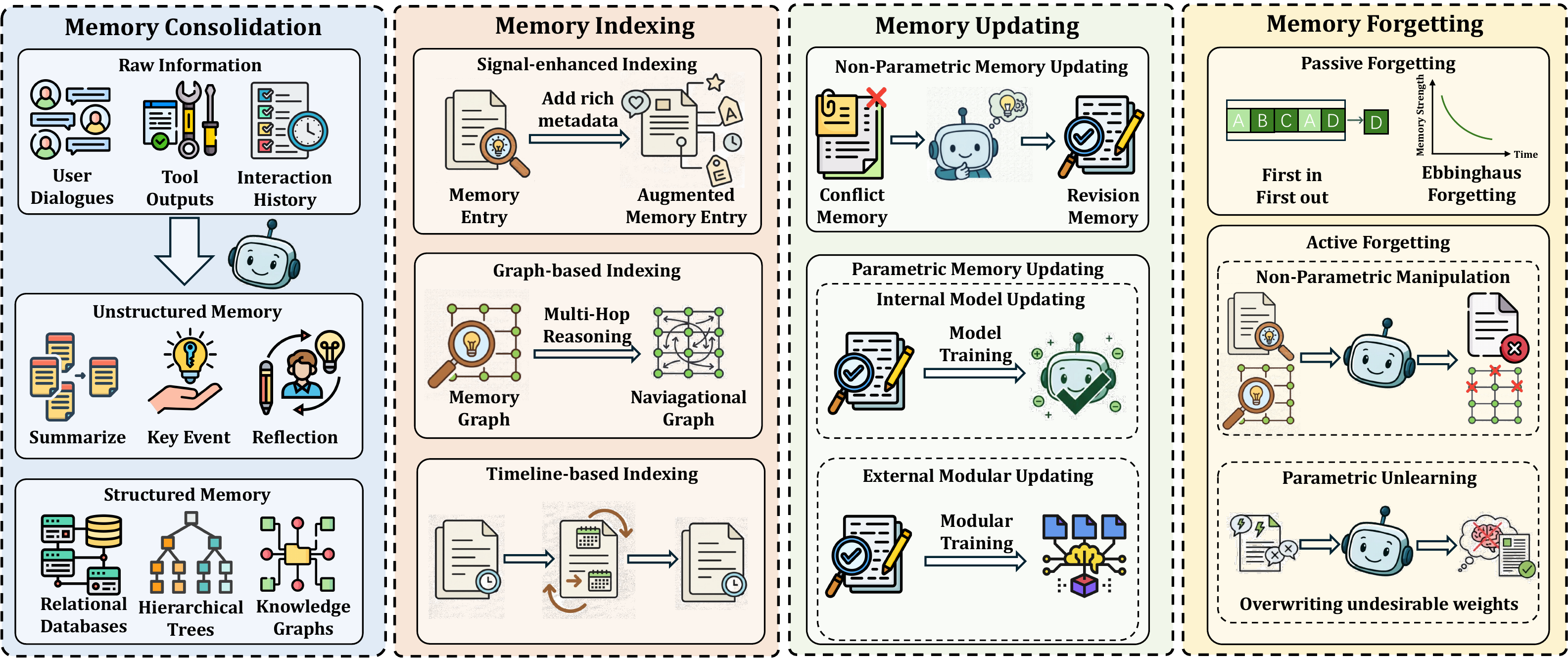}
    \caption{Memory management contains four key stages: (1) Memory Consolidation, (2) Memory Indexing, (3) Memory Updating, and (4) Memory Forgetting.}
    \label{fig:memory}
\end{figure*}

\subsection{Memory Management}\label{memory-management}

\header{Definition.} Memory management is a foundational component of advanced DR architectures, which governs the dynamic lifecycle of context used by DR agents in complex, long-horizon tasks~\cite{wu2025human,du2025rethinking,jiang2024long}, 
aiming to maintain coherent and relevant task-solving context~\cite{he2024human,zhang2024survey,Sun2025ScalingLL}.

\header{Core Operation.}
As illustrated in Figure~\ref{fig:memory}, memory management typically involves four core operations: consolidation, indexing, updating, and forgetting.
Consolidation converts short-term experiences into durable representations that form the basis for later indexing.
Indexing organizes these representations into retrieval structures that support efficient recall during problem solving.
Updating refines or corrects stored knowledge, whereas forgetting selectively removes outdated or irrelevant content to reduce interference.
In the following sections, we discuss consolidation, indexing, updating, and forgetting in detail.

\subsubsection{Memory Consolidation}\label{memory-consolidation}

\header{Definition.}
Memory consolidation is the process of transforming transient, short-term information, such as user dialogues or tool execution outputs, into stable, long-term representations~\cite{squire2015memory,du2025rethinking,wu2025human}. Drawing an analogy to cognitive neuroscience, this process encodes and abstracts raw inputs to create durable memory engrams, laying the groundwork for efficient long-term storage and retrieval~\cite{wu2025human}. 

Memory consolidation involves transforming interaction histories into durable formats, including but not limited to model parameters~\cite{wang2024towards}, structured graphs~\cite{zhao2025eventweave}, or knowledge bases~\cite{lu2023memochat,du2025rethinking}. 
Distinct from memory indexing, which creates navigable access pathways over existing memories, consolidation is fundamentally concerned with the initial transformation and structural organization of raw experience. Two primary paradigms for this process have emerged: (i) \textit{unstructured memory consolidation} and (ii) \textit{structured memory consolidation}.

\header{Unstructured Memory Consolidation.} This paradigm distills lengthy interaction histories or raw texts into high-level, concise summaries or key event logs. For example, MemoryBank~\cite{zhong2024memorybank} processes and distills conversations into a high-level summary of daily events, which helps in constructing a long-term user profile. Similarly, MemoChat~\cite{lu2023memochat} summarizes conversation segments by abstracting the main topics discussed, while ChatGPT-RSum~\cite{wang2025recursively} adopts a recursive summarization strategy to manage extended conversations. Other approaches focus on abstracting experiences; Generative Agents~\cite{park2023generative} utilize a reflection mechanism triggered by sufficient event accumulation to generate more abstract thoughts as new, consolidated memories. To create generalizable plans, GITM~\cite{zhu2023ghost} summarizes key actions from multiple successful plans into a common reference memory.

\header{Structured Memory Consolidation.} This paradigm transforms unstructured information into highly organized formats such as databases, graphs, or trees. This structural encoding is the primary act of consolidation, designed to capture complex inter-entity relationships and create an organized memory corpus. For instance, TiM~\cite{liu2023think} extracts entity relationships from raw information and stores them as tuples in a structured database. ChatDB~\cite{hu2023chatdb} leverages a database as a form of symbolic memory, transforming raw inputs into a queryable, relational format. AriGraph~\cite{anokhin2024arigraph} implements a memory graph where knowledge is represented as vertices and their interconnections as edges. Similarly, HippoRAG~\cite{jimenez2024hipporag} constructs knowledge graphs over entities, phrases, and summaries to form an interconnected web of fragmented knowledge units. MemTree~\cite{rezazadeh2024isolated} builds and updates a tree structure by traversing from the root and deciding whether to deepen the tree with new information or create new leaf nodes based on semantic similarity. This hierarchical organization is the core of its consolidation strategy, enabling structured storage of memories.

\subsubsection{Memory Indexing}\label{memory-indexing}

\header{Definition.}
Memory indexing involves constructing a navigational map over a DR agent's consolidated memories, analogous to a library's catalog or a book's index for efficient information retrieval~\cite{maekawa2023generative}. Unlike memory consolidation, which focuses on the initial transformation of raw data into a durable format, indexing operates on already consolidated memories to create efficient, semantically rich retrieval pathways. This process builds auxiliary access structures that enhance retrieval not only in efficiency but also in relevance. 

Effective indexing goes beyond simple keyword matching by encoding temporal~\cite{mehta2022dsi++} and relational~\cite{jimenez2024hipporag} dependencies among memories. This is typically achieved by generating auxiliary codes, such as vector embeddings, summaries, or entity tags, which serve as retrieval entry points into the memory store. Given the vast, high-dimensional spaces these codes inhabit, specialized search techniques are required, such as Locality-Sensitive Hashing (LSH)~\cite{datar2004locality}, Hierarchical Navigable Small World (HNSW) graphs~\cite{malkov2018efficient}, or libraries like FAISS~\cite{johnson2019billion}  for high-speed similarity search. These access mechanisms are commonly organized through three established paradigms:
\begin{itemize}[leftmargin=*,nosep]
    \item \textbf{Signal-enhanced Indexing.} This paradigm augments consolidated memory entries with auxiliary metadata, including emotional context, topics, and keywords, which function as granular pivots for context-aware retrieval~\cite{sun2025hierarchical,zhang2025multiple}. For instance, LongMemEval~\cite{wu2024longmemeval} enhances memory keys by integrating temporal and semantic signals to improve retrieval precision. Similarly, the Multiple Memory System (MMS)~\cite{zhang2025multiple} decomposes experiences into discrete components, such as cognitive perspectives and semantic facts, thereby facilitating multifaceted retrieval strategies.

    \item \textbf{Graph-based Indexing.} This paradigm leverages a graph structure, where memories are nodes and their relationships are edges, as a sophisticated index. By representing memory networks in this way, agents can perform complex multi-hop reasoning by traversing chains of connections to locate information that is not explicitly linked to the initial query~\cite{chhikara2025mem0,long2025seeing}. For instance, HippoRAG~\cite{jimenez2024hipporag} uses lightweight knowledge graphs to explicitly model inter-memory relations, enabling structured, interpretable access. A-Mem~\cite{xu2025mem} adopts a dynamic strategy where the agent autonomously links related memory notes, progressively growing a flexible access network.

    \item \textbf{Timeline-based Indexing.} This paradigm creates a temporal index by organizing memory entries along chronological or causal sequences. Such structuring provides a historical access pathway, which is essential for understanding progression, maintaining conversational coherence, and supporting lifelong learning~\cite{wang2025comorag}. For example, the Theanine system~\cite{ong2024towards} arranges memories along evolving timelines to facilitate retrieval based on both relevance and temporal dynamics. Zep~\cite{rasmussen2025zep} introduces a bi-temporal model for its knowledge graph, indexing each fact with $t_{valid}$ and $t_{invalid}$ timestamps, which allows the agent to navigate the memory based on temporal validity.
\end{itemize}

\subsubsection{Memory Updating}\label{memory-updating}

\header{Definition.}
Memory updating is a core capability of DR agents, involving the reactivation and modification of existing knowledge in response to new information or environmental feedback~\cite{wang2024knowledge, tack2024online, wang2024memoryllm}. This process is essential for maintaining the consistency, accuracy, and relevance of the agent’s internal world model, thereby enabling continual learning and adaptive behavior in dynamic environments~\cite{wang2023voyager, wang2024wise}.

Memory updating governs how an agent corrects factual inaccuracies, incorporates new information, and gradually improves its knowledge base~\cite{shinn2023reflexion, de2021editing, mitchell2021fast}.
Although related to memory forgetting, which focuses on removing outdated or incorrect content, memory updating centers on modifying and refining existing knowledge to increase its fidelity.
In the following, we introduce two updating strategies, depending on whether the memory is external (non-parametric) or internal (parametric) to the model~\cite{wang2024wise}.

\header{Non-Parametric Memory Updating.}
Non-parametric memory, stored in external formats such as vector databases or structured files, is updated via explicit, discrete operations on the data itself. This approach offers flexibility and transparency. Key operations include:

\begin{itemize}[leftmargin=*,nosep]
    \item \textit{Integration and Conflict Updating.} This operation focuses on incorporating new information and refining existing entries to maintain logical consistency. For example, the Mem0 framework employs an LLM to manage its knowledge base through explicit operations, such as adding new facts (\texttt{ADD}) or modifying existing entries with new details (\texttt{UPDATE}) to resolve inconsistencies~\cite{chhikara2025mem0}. To handle temporal conflicts, Zep updates its knowledge graph by modifying an existing fact's effective time range, setting an invalidation timestamp ($t_{invalid}$) to reflect that a newer fact has superseded it~\cite{rasmussen2025zep}. Similarly, the TiM framework curates its memory by using \texttt{MERGE} operations to combine related facts into a more coherent representation~\cite{liu2023think}

    \item \textit{Self-Reflection Updating.} Inspired by human memory reconsolidation, this paradigm enables agents to iteratively refine their knowledge by reflecting on past experiences~\cite{shinn2023reflexion, zhou2025memento}. Early systems like Reflexion~\cite{shinn2023reflexion} and Voyager~\cite{wang2023voyager} implement this through verbal self-correction and updates to a skill library. More dynamically, A-Mem~\cite{xu2025mem} triggers a Memory Evolution process that re-evaluates and autonomously refines previously linked memories based on new contextual information.
\end{itemize}

\header{Parametric Memory Updating.}
Parametric memory, encoded directly in a model’s weights, is updated by modifying internal representations. This is typically more complex and computationally intensive. Three main approaches have emerged:

\begin{itemize}[leftmargin=*,nosep]
    \item \textit{Global Updating.} This approach integrates new knowledge by continuing model training on additional datasets~\cite{shao2023character}. While effective for large-scale adaptation, it is computationally expensive and prone to catastrophic forgetting~\cite{wang2024knowledge}. To address this, instead of simply injecting factual knowledge, Memory-R1 trains a dedicated Memory Manager agent to learn an optimal policy for modification operations such as \texttt{ADD} and \texttt{UPDATE}, moving beyond heuristic rules~\cite{yan2025memory}. Additionally, a recent framework refines this process by employing methods such as Direct Preference Optimization to fine-tune the model’s memory utilization strategy~\cite{zhang2025learn}.

    \item \textit{Localized Updating.} This technique modifies specific facts in the model's parameters without requiring full retraining~\cite{de2021editing, mitchell2021fast}. It is especially suited for online settings where rapid adaptation is needed, such as updating a user's preference~\cite{tack2024online}. Methods typically follow a \textit{locate-and-edit} strategy or use meta-learning to predict weight adjustments while preserving unrelated knowledge~\cite{mitchell2021fast, tack2024online}.

    \item \textit{Modular Updating.} This emerging paradigm avoids the risks of continual weight modification by distilling knowledge into a dedicated, plug-and-play parametric module. Frameworks such as MLP Memory~\cite{wei2025mlp} and Memory Decoder~\cite{cao2025memory} train a lightweight external module to imitate the output distribution of a non-parametric $k$NN retriever. This process effectively compiles a large corpus of external knowledge into the compact weights of the module. The resulting module can then be attached to any compatible LLM to provide specialized knowledge without modifying the base model’s parameters, thereby avoiding catastrophic forgetting and reducing the latency of real-time retrieval~\cite{wei2025mlp, cao2025memory}.
\end{itemize}

\subsubsection{Memory Forgetting}\label{memory-forgetting}

\header{Definition.}
Forgetting constitutes a fundamental mechanism in advanced agent architectures, enabling the selective removal or suppression of outdated, irrelevant, or potentially erroneous memory content. Rather than a system defect, forgetting is a functional process critical for filtering noise, reclaiming finite storage resources, and mitigating interference between conflicting information. In contrast to memory updating, which modifies existing knowledge to improve its accuracy, forgetting is a subtractive process that streamlines the memory store by eliminating specific content. This process can be broadly categorized into passive and active mechanisms.

\header{Passive Forgetting.}
This simulates the natural decay of human memory, in which infrequently accessed or temporally irrelevant memories gradually lose prominence. This mechanism is particularly critical for managing the agent's immediate working memory or context window. Implementations are typically governed by automated, time-based rules rather than explicit content analysis. For instance, MemGPT~\cite{packer2023memgpt} employs a First-In-First-Out (FIFO) queue for recent interactions, automatically moving the oldest messages from the main context into long-term storage. MemoryBank~\cite{zhong2024memorybank} draws inspiration from the Ebbinghaus forgetting curve, in which memory traces decay over time unless reinforced, allowing the agent to naturally prioritize recent content. A more aggressive approach, MEM1~\cite{zhou2025mem1}, employs a \textit{use-and-discard} policy: after each interaction, the agent synthesizes essential information into a compact state and immediately discards all prior contextual data to maintain constant memory consumption.

\header{Active Forgetting.}
Active forgetting involves the intentional and targeted removal or invalidation of specific memory content. This process is a deliberate action, often triggered by the detection of contradictions or the need to correct inaccurate information, and its implementation varies depending on the memory type.

\begin{itemize}[leftmargin=*,nosep]
    \item \textit{Non-Parametric Memory.} Active forgetting in external memory stores involves direct data manipulation. For example, Mem0~\cite{chhikara2025mem0} implements an explicit \texttt{DELETE} command to remove outdated or contradictory facts. Similarly, TiM~\cite{liu2023think} introduces a dedicated \texttt{FORGET} operation to actively purge irrelevant or incorrect thoughts from its memory cache. Reinforcement learning can also be used to train a specialized Memory Manager agent to autonomously decide when to execute a \texttt{DELETE} command, as seen in the Memory-R1 framework~\cite{yan2025memory}. AriGraph~\cite{anokhin2024arigraph} maintains a structured memory graph by removing outdated vertices and edges. Some systems employ non-destructive forgetting; the Zep architecture~\cite{rasmussen2025zep}, for example, uses \textit{edge invalidation} to assign an invalid timestamp to an outdated entry, effectively retiring it without permanent deletion.

    \item \textit{Parametric Memory.} In this context, active forgetting is typically achieved through machine unlearning techniques that modify a model’s internal parameters to erase specific knowledge without full retraining. Approaches include locating and deactivating specific neurons or adjusting training objectives to promote the removal of targeted information. For example, MEOW~\cite{gu2024meow} facilitates efficient forgetting by fine-tuning an LLM on generated contradictory facts, effectively overwriting undesirable memories stored in its weights.
\end{itemize}

\begin{tcolorbox}[takeawaysbox, title={Takeaway}]
Memory management is a cornerstone of the DR paradigm, enabling agents to transcend single-turn interactions and conduct complex, long-horizon investigations by governing the information lifecycle. Through the interdependent operations of consolidation, indexing, updating, and forgetting, the DR system maintains the context and coherence essential for an iterative research loop. Consequently, a sophisticated memory framework is what fundamentally distinguishes a DR agent from a simple RAG system, equipping it with the consistency, adaptability, and self-evolution necessary to autonomously synthesize comprehensive, trustworthy, and verifiable reports from a vast and dynamic information landscape.
\end{tcolorbox}

\subsection{Answer Generation}\label{answer-generation}

\header{Definition.}
Answer generation typically represents the culminating stage of a DR system. It synthesizes information from upstream components, such as query planning (Section~\ref{query-planning}), information acquisition (Section~\ref{memory-consolidation}), and memory systems (Section~\ref{memory-consolidation}), and generates a coherent, comprehensive, and well-supported response that accurately reflects the user’s original intent.

\begin{figure}[!t]
    \centering
    \includegraphics[width=\linewidth]{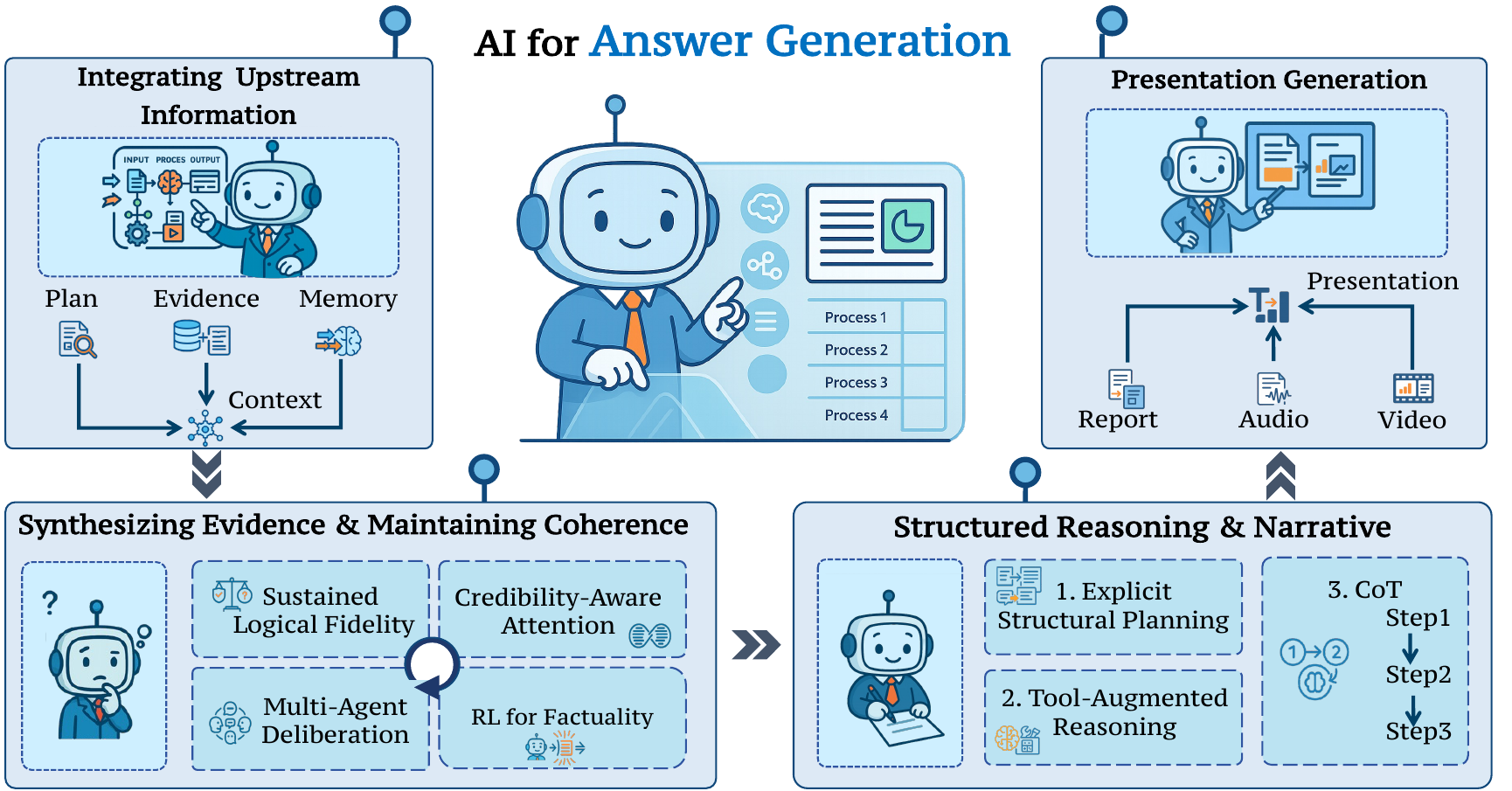}
    \caption{Illustrating the schematic of the answer generation process in DR. The workflow begins by integrating upstream information, moves to synthesizing evidence and ensuring coherence, then constructs a structured narrative via reasoning, and culminates in a multimodal presentation output.}
    \label{fig:answergeneration}
\end{figure}

Unlike traditional text generation, the answer generation within an advanced DR workflow addresses complex challenges such as reconciling conflicting evidence, maintaining long-range coherence, and structuring outputs with transparent reasoning and proper citations. 
It has evolved from template-based generation~\citep{kim2019improving} to sophisticated synthesis shown in Figure~\ref{fig:answergeneration}, which reflects the growing demand for trustworthy, explainable, and multimodal research outputs~\cite{lewis2020retrieval, borgeaud2022improving}.
To deconstruct this process, we will explore it across four progressive stages: beginning with the integration of diverse information sources, moving to the synthesis of evidence and maintenance of coherence, then structuring the reasoning and narrative, and finally, advancing to the frontier of cross-modal generation.

\subsubsection{Integrating Upstream Information}\label{integrating-upstream-information}

\header{Definition.}
The main principle of trustworthy answer generation is to ensure that every statement is grounded in verifiable external evidence.
Thus, the first stage of answer generation is integrating information from its upstream components, including:
the sub-queries from the query planning, the ranked and potentially conflicting evidence, and the evolving contextual state stored in memory.

Recent developments in this area demonstrate sophisticated strategies for integrating upstream information, moving from simple evidence-feeding to dynamic, state-aware synthesis. The most common approach involves integrating ranked evidence from the retrieval module. Frameworks like Self-RAG~\cite{asai2024self}, for example, employ a more dynamic integration by adaptively retrieving passages on demand. It then generates reflection tokens to assess the relevance of the retrieved information and its own generation, effectively integrating an internal self-correction mechanism to steer the synthesis. Moving beyond static evidence, more advanced systems integrate the query plan with an evolving memory state to ensure long-range coherence, a paradigm known as Stateful Query Planning. For instance, graph-centric frameworks like Plan-on-Graph (PoG)~\cite{chen2024plan} explicitly integrate the plan with a dynamic memory (storing sub-goal status, explored paths, and retrieved entities). This memory is then actively used during a \textit{reflection} step to guide and self-correct subsequent planning, tightly coupling the reasoning state with the generation process. Similarly, search-based frameworks like MCTS-OPS~\cite{yu2025optimizing} formalize this by treating the MCTS tree itself as the state of the evolving query plan. Here, the system integrates its experiential memory (node values from past rollouts) to guide the \texttt{SELECTION} and \texttt{EXPANSION} of the next planning step, ensuring the final answer synthesizes the full context of the problem-solving process. 
 
While these architectures provide a robust foundation, the core challenges of synthesizing contradictory evidence and maintaining long-form coherence remain the next frontier.

\subsubsection{Synthesizing Evidence and Maintaining Coherence}\label{synthesizing-evidence-and-maintaining}

Producing answers to research-level questions requires resolving informational conflicts and sustaining coherent, information-dense narration across extended outputs.

\header{Resolving Conflicting Evidence.} 
Research queries frequently surface contradictory sources, requiring the model to discriminate among varying levels of reliability. Building on fact-verification paradigms~\cite{thorne2018fever}, recent systems adopt three major strategies.
\begin{itemize}
    \item \textit{Credibility-Aware Attention}: Instead of treating all retrieved information equally, this approach intelligently weighs evidence based on its source. The system assigns a higher score to information coming from more credible sources (\eg  a top-tier scientific journal) compared to less reliable ones (\eg an unverified blog)~\citep{deng2025cram}. This allows the model to prioritize trustworthy information while still considering relevant insights from a wider range of sources~\cite{chen2025resolving}.

    \item \textit{Multi-Agent Deliberation}: This strategy simulates an expert committee meeting to debate the evidence. Frameworks like MADAM-RAG~\cite{kim2024retrieval} employ multiple independent AI agents, each tasked with analyzing the retrieved documents from a different perspective. Each agent forms its own assessment and conclusion. Afterwards, a final \textit{meta-reasoning} step synthesizes these diverse viewpoints to forge a more robust and nuanced final answer, much like a panel of experts reaching a consensus~\citep{wang2025retrieval}.

    \item \textit{Reinforcement Learning for Factuality}: This method trains the generator through a trial-and-error process that rewards factual accuracy~\citep{sun2025dynamicragleveragingoutputslarge}. A representative approach is RioRAG~\cite{li2025reinforced}, in which an LLM receives a positive reward when it generates statements that are strongly and consistently supported by the provided evidence. Conversely, it is penalized for making unsubstantiated claims or statements that contradict the source material, shaping the model to inherently prefer generating factually grounded and reliable answers.
\end{itemize}

\header{Long-form Coherence and Information Density.} 
Another key challenge is ensuring \textbf{Sustained Informational Accuracy}. Research answers are often lengthy, and maintaining a logical thread while avoiding repetition or verbosity is non-trivial. 
Let $L_{\text{model}}$ denote the maximum coherent length of a model’s output, and $L_{\text{SFT}}$ represent the average length of examples in its supervised fine-tuning dataset. SFT offers an intuitive approach to enhancing the long-form generation capabilities of large language models. However, LongWriter~\cite{zhang2024longwriter} empirically demonstrates that the maximum coherent length of a model’s output often scales with the average length of its fine-tuning samples, which can be formally expressed as $L_{\text{model}} \propto L_{\text{SFT}}$~\cite{zhang2024longwriter}. 
To address this, LongWriter focuses on systematic training for extended generation, while others use reflection-driven processes to iteratively improve consistency~\cite{liu2025superwriter}. Additionally, RioRAG~\cite{li2025reinforced} introduces a length-adaptive reward function to promote information density, which penalizes verbosity that fails to add informational value, preventing reward hacking through verbosity. Together, these techniques shift the focus of generation from mere content aggregation toward credible, concise, and coherent synthesis, laying the groundwork for structured reasoning.

\subsubsection{Structuring Reasoning and Narrative}\label{structuring-reasoning-and-narrative}

The research community's focus is shifting from the mere factual accuracy of DR systems to the crucial need for explainability and logical rigor in their answers. An opaque answer, which prevents users from tracing the underlying reasoning process, has significantly diminished utility in critical domains like scientific research~\citep{hong2023metagpt,luo2025intention,shi2025search}. Consequently, a significant line of work has emerged to enable models to generate structured reasoning processes rather than just monolithic final answers~\citep{wei2022chain, zhou2022least, yang2025rapid}. This trend is reflected in the design of most modern DeepResearch systems, which increasingly favor the explicit presentation of this structural information \citep{yang2025researstudio,zhou2025memento}.

\header{Prompt-based Chain-of-Thought.} This foundational approach focuses on eliciting intermediate reasoning steps before producing a final answer. The most prominent technique is Chain-of-Thought (CoT) prompting~\cite{wei2022chain}, which can be formally expressed as $\mathcal{R} = \text{LLM}(\text{CoT-Prompt} + \mathcal{Q} + \text{Evidence})$. This method enhances both interpretability and multi-step reasoning performance. Its applicability has been broadened by extensions such as zero-shot CoT~\cite{kojima2022large} and Least-to-Most prompting~\cite{zhou2022least}.

\header{Explicit Structural Planning.} More advanced systems move beyond simple linear chains to formalize the structure of the entire answer. For instance, RAPID~\cite{yang2025rapid} formalizes this process into three stages: (i) \textit{outline generation}; (ii) \textit{outline refinement through evidence discovery}; and (iii) \textit{plan-guided writing}, where the outline forms a directed acyclic graph to support complex, non-linear argumentation. Similarly, SuperWriter~\cite{liu2025superwriter} extends this idea by decoupling the reasoning and text-production phases and optimizing the entire process via hierarchical Direct Preference Optimization.

\header{Tool-Augmented Reasoning.} 
This line of work enhances reasoning by dynamically interfacing with external resources. Representative work allows models to invoke external computational or retrieval tools dynamically, ensuring both analytic rigor and factual grounding~\cite{schick2023toolformer,qin2023toolllm,hu2025owl,weng2025cycleresearcher,shi2025tool}.

\subsubsection{Presentation Generation}\label{cross-modal-reasoning-and-generation}
The frontier of answer generation extends beyond text, encompassing the integration of multimodal and structured information. Research questions increasingly demand answers that combine textual reasoning with visual, tabular, or auditory data, maintaining semantic and presentational coherence. Early breakthroughs such as BLIP-2~\cite{li2023blip} and InstructBLIP~\cite{dai2023instructblip} enable multimodal instruction-following by aligning vision-language embeddings. MiniGPT-4~\cite{chen2023minigpt} advances this by leveraging cross-modal attention to seamlessly integrate visual and textual evidence.

Recently, a series of works have demonstrated higher presentation capabilities, signaling an evolution from content generation to presentation generation~\citep{yi2023urania, xie2024controlcap, zhu2025paper2video}.
Existing work like MedConQA~\cite{xia-etal-2022-medconqa}, LIDA~\cite{victor2023lida}, ChartGPT~\cite{yuan2023chartgpt}, and Urania~\cite{yi2023urania} can synthesize data analyses into dynamic, interactive visualizations. 
Others work, including PresentAgent~\cite{jingwei2025presentagent}, Qwen2.5-Omni~\cite{jin2025qwen25omni}, and AnyToAny~\cite{zineng2023anytoany}, generates synchronized audio narrations alongside text.
More recently, PPTAgent~\cite{zheng2025pptagentgeneratingevaluatingpresentations} and Paper2Video~\cite{zhu2025paper2video} even extend to editable presentation generation, where full analytical reports are automatically transformed into slide decks with coordinated text, figures, and layout elements.
At the leading edge, video-grounded agents~\cite{10.1109/TPAMI.2024.3411045,10095026} retrieve or generate relevant visual footage, delivering answers through multimodal storytelling. As summarized in Table~\ref{tab:dr_output_capabilities}, while most DR systems still focus on textual synthesis with citations, only a handful, such as OpenAI DeepResearch~\cite{openai_deep_research} and H2O.ai DeepResearch~\cite{H2Oai_DeepResearch}, currently support comprehensive multimodal output. 
The emerging consensus suggests that rich, multi-format answer generation will soon become a standard expectation~\citep{xie2025far}, bridging the gap between knowledge synthesis and human-centered presentation.

\begin{tcolorbox}[takeawaysbox, title={Takeaway}]
Answer generation represents the synthesis core of DR systems, integrating upstream information, reconciling conflicting evidence, and structuring coherent, evidence-grounded narratives. Recent advances, from credibility-aware attention and multi-agent deliberation to reinforcement learning for factuality, have enhanced both factual reliability and interpretability. Systems now move beyond content aggregation toward concise, logically structured synthesis supported by transparent reasoning frameworks such as Chain-of-Thought and plan-guided writing. Moreover, the frontier of answer generation extends into multimodal generation, where text, visuals, tables, and audio coalesce into rich, human-centered outputs. These developments mark a paradigm shift from generating text to generating explainable, trustworthy, and presentation-ready knowledge.
\end{tcolorbox}

\input{table/project}

%% file: table/project.tex
\begin{table}[!t] 
  {\scriptsize
  \setlength{\tabcolsep}{4pt}
  \rowcolors{4}{gray!10}{white}
  \centering
  \caption{Comparing output capabilities of contemporary representative  DR systems, where the \blackbox \textbf{ }indicates  \textbf{supported capability}}
  \label{tab:dr_output_capabilities}
  \begin{tabular}{%
      >{\centering\arraybackslash}m{3.5cm}|
      >{\centering\arraybackslash}m{0.85cm}
      >{\centering\arraybackslash}m{0.85cm}
      >{\centering\arraybackslash}m{0.85cm}
      >{\centering\arraybackslash}m{0.85cm}
      >{\centering\arraybackslash}m{0.85cm}|
      >{\centering\arraybackslash}m{0.85cm}
      >{\centering\arraybackslash}m{0.85cm}
      >{\centering\arraybackslash}m{0.85cm}|
      >{\centering\arraybackslash}m{0.85cm}
      >{\centering\arraybackslash}m{0.85cm}
      >{\centering\arraybackslash}m{0.85cm}
    }
    \toprule
    \multirow{2}{*}{\small \textbf{System}} & \multicolumn{5}{c|}{\small \textbf{Content Generation}} & \multicolumn{3}{c|}{\small \textbf{Structured Output}} & \multicolumn{3}{c}{\small \textbf{Advanced}} \\
      & \tiny \textbf{Text} & \tiny \textbf{Image} & \tiny \textbf{Audio} & \tiny \textbf{Video} & \tiny \textbf{Pres.} & \tiny \textbf{Table} & \tiny \textbf{JSON} & \tiny \textbf{Code} & \tiny \textbf{Chart} & \tiny \textbf{GUI} & \tiny \textbf{Cite} \\
    \midrule
    Gemini DeepResearch~\citep{gemini_deep_research} & \blackbox & \blackbox &  &  &  & \blackbox &  &  & \blackbox &  & \blackbox \\
    Grok DeepSearch~\citep{grok_deepsearch} & \blackbox &  &  &  &  &  &  &  &  &  & \blackbox \\
    OpenAI DeepResearch~\citep{openai_deep_research} & \blackbox & \blackbox &  &  &  & \blackbox &  & \blackbox & \blackbox & \blackbox & \blackbox \\
    AutoGLM~\citep{liu2024autoglm} & \blackbox &  &  &  &  &  &  &  &  &  & \blackbox \\
    H2O.ai DeepResearch~\citep{H2Oai_DeepResearch} & \blackbox & \blackbox & \blackbox &  & \blackbox & \blackbox & \blackbox & \blackbox & \blackbox &  & \blackbox \\
    Skywork DeepResearch~\citep{skywork2025deepresearch} & \blackbox &  &  & \blackbox & \blackbox &  &  &  &  &  & \blackbox \\
    Perplexity DeepResearch~\citep{perplexity_deep_research} & \blackbox &  &  &  &  &  &  &  &  &  & \blackbox \\
    Manus~\citep{manus2025} & \blackbox &  &  &  & \blackbox & \blackbox &  &  & \blackbox &  & \blackbox \\
    OpenManus~\citep{openmanus2025} & \blackbox &  &  &  &  &  &  &  & \blackbox &  & \blackbox \\
    OWL (CAMEL-AI)~\citep{hu2025owl} & \blackbox &  &  &  & \blackbox & \blackbox &  &  & \blackbox &  & \blackbox \\
    SunaAI~\citep{suna2025} & \blackbox &  &  &  & \blackbox & \blackbox &  &  & \blackbox &  &  \\
    Alita~\citep{qiu2025alita} & \blackbox &  &  &  &  &  &  &  & \blackbox &  &  \\
    \bottomrule
  \end{tabular}}
\end{table}

%% file: sections/04-technique.tex
\section{Practical Techniques for Optimizing Deep Research Systems}\label{sec:practice}

So far, we have introduced the core components that constitute a typical DR system. 
Building on these foundation, we now delve into practical techniques for improving such DR systems in real-world settings. These techniques focus on how to flexibly coordinate and enhance the key components, with the goal of achieving more reliable and effective task completion.
Below, we discuss three commonly used paradigms: workflow prompting, supervised fine-tuning, and agentic reinforcement learning.
Workflow prompting typically relies on a carefully designed pipeline (\textit{aka.,} prompting engineering) that guides the agents.
The latter two paradigms aim to train a specific DR agent capable of reasoning, retrieving information, and generating high-quality answers.

\subsection{Workflow Prompt Engineering}\label{sec:workflow}

\header{Definition.}
A simple yet effective way to build a DR system is to construct a complex workflow that enables collaboration among multiple agents.
In the most common setting, an orchestration agent coordinates a team of specialized \textit{worker agents}, allowing them to operate in parallel on different aspects of a complex research task.
To illustrate the key principles and design considerations behind such a DR workflow, we introduce Anthropic Deep Research~\citep{Anthropic2025Research} as a representative example.

\subsubsection{Deep Research System of Anthropic}\label{sec:anthropic}

Anthropic proposes a multi-agent Deep Research (DR) framework where a \textbf{lead orchestrator} coordinates multiple \textbf{worker agents} through structured, auditable interactions. The system transforms an open-ended research query into a complete workflow, from planning and delegation to synthesis and citation, under an explicit research budget controlling agent count, tool usage, and reasoning depth.
We highlight several \textbf{core points} that enable the system’s efficiency and reliability:
\begin{itemize}[leftmargin=*,nosep]
    \item \textit{Query Stratification and Planning.} The orchestrator first analyzes the semantic type and difficulty of the input query (\eg  depth-first vs.~breadth-first) to determine research strategy and allocate a corresponding budget of agents, tool calls, and synthesis passes.
    
    \item \textit{Delegation and Scaling.} Effort scales with complexity: from 1–2 agents for factual lookups to up to 10 or more for multi-perspective analyses, each assigned with clear quotas and stopping criteria to enable dynamic budget reallocation.
    
    \item \textit{Task Decomposition and Prompt Specification.} The main query is decomposed into modular subtasks, each encoded as a structured prompt specifying objectives, output schema, citation policy, and fallback actions to ensure autonomy with accountability.
    
    \item \textit{Tool Selection and Evidence Logging.} A central tool registry (\eg  web fetch, PDF parsing, calculators) is used following freshness, verifiability, and latency rules. Agents record all tool provenance in an evidence ledger for traceable attribution.
    
    \item \textit{Parallel Gathering and Interim Synthesis.} Worker agents operate concurrently while the orchestrator monitors coverage, resolves conflicts, and launches micro-delegations to close residual gaps or trigger deeper reasoning where needed.
    
    \item \textit{Final Report and Attribution.} The orchestrator integrates verified findings into a coherent report, programmatically linking claims to sources and ensuring schema compliance, factual grounding, and transparent citation.
\end{itemize}
Overall, Anthropic’s system exemplifies a scalable, interpretable multi-agent research paradigm that achieves high-quality synthesis through modular delegation, explicit budgeting, and verifiable reasoning.

\subsection{Supervised Fine-Tuning}\label{sec:SFT}

\header{Definition.}
Supervised fine-tuning (SFT) is a widely adopted approach that trains models to imitate desired behaviors using input–output pairs under a supervised learning objective. 
Within DR, SFT is commonly employed as the \textit{cold start}, \eg a  warm-up process, before online reinforcement learning~\cite{jin2025search,li2025search,wu2025masksearch,song2025r1,li2025websailorv2}.
It aims to endow agents with basic task-solving skills~\cite{qi2024webrl,wei2025webagent}.

Since manual collection of expert trajectories is labor-intensive, costly, and difficult to scale, a key challenge lies in automatically constructing high-quality SFT datasets.
This has been widely explored by prior work~\cite{wang2022self,zhou2023lima,taori2023stanford,chung2024scaling}. 
Below, we categorize representative work into two main paradigms:
(i) \textit{strong-to-weak distillation}, distilling correct task-solving trajectories from powerful LLMs (\eg GPT-5 and DeepSeek-V3.1) into smaller, weak models; and
(ii) \textit{iterative self-evolution}, iteratively fine-tuning the model on the dataset produced by itself, leading to a progressive improvement.

\begin{figure}[!t]
    \centering
    \includegraphics[width=1\linewidth]{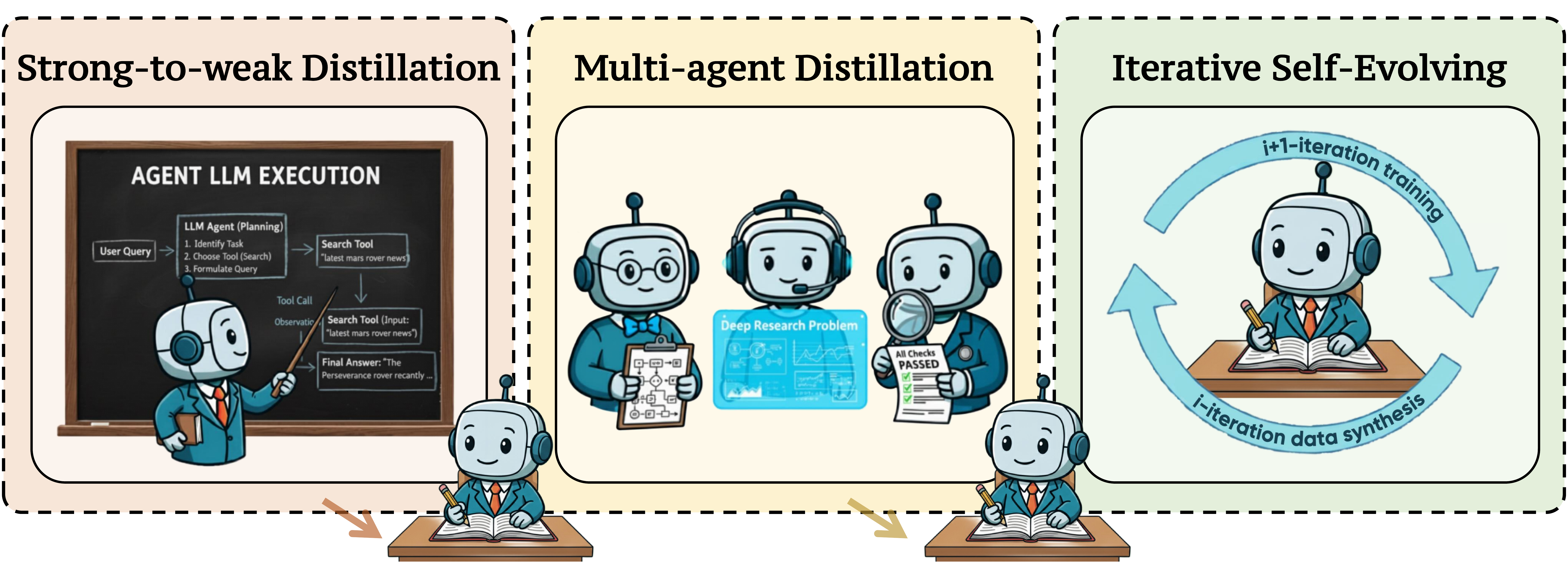}
        \caption{Comparisons among three types of data synthesis approaches, including: (i) Strong-to-Weak Distillation, (ii) Multi-Agent Distillation, and (iii) Iterative Self-Evolving.
        Each type is illustrated through the process of how agents perform tasks, learn, and refine their abilities.}
    \label{fig:method}
\end{figure}

\subsubsection{Strong-to-weak Distillation}\label{sec:strong-to-weak-distillation}

\header{Definition.}
Strong-to-weak distillation transfers high-quality decision trajectories from a powerful \textit{teacher} system to smaller, weaker \textit{student} models. 
Early work predominantly uses a single LLM-based agent to synthesize trajectories; more recent research employs multi-agent teacher systems to elicit more diverse, higher-complexity trajectories. 
We detail these two lines of work below.

\header{Single-agent distillation.}
Representative systems instantiate this pipeline in various ways. 
WebDancer~\cite{wu2025webdancer} provides the agent with search and click tools. A strong non-reasoning model generates short CoT, while a large reasoning model (LRM) generates long CoT. The agent learns from both, using rejection sampling for quality control. WebSailor~\cite{li2025websailor} uses an expert LRM to generate action-observation trajectories, then reconstructs short CoT with a non-reasoning model, ensuring the final reasoning chain is compact enough for long-horizon tasks. WebShaper~\cite{tao2025webshaper} uses search and visit tools in a ReAct-style trajectory. It performs 5 rollouts per task and filters out repeated or speculative answers using a reviewing LLM.
WebThinker~\cite{li2025webthinker} augments SFT with policy-gradient refinement and 
WebSynthesis~\cite{gao2025websynthesis} leverages a learned world model to simulate virtual web environments and employs MCTS to synthesize diverse, controllable web interaction trajectories entirely offline.

\header{Multi-agent distillation.}
Multi-agent distillation synthesizes training data using an agentic teacher system composed of specialized, collaborating agents (\eg a planner, a tool caller, and a verifier), with the goal of transferring emergent problem-solving behaviors into a single end-to-end student model~\cite{ge2024scaling, tang2024synthesizing}. This paradigm tends to produce diverse trajectories, richer tool-use patterns, and explicit self-correction signals.

A representative work is MaskSearch~\cite{wu2025masksearch}, which constructs a multi-agent pipeline that includes a planner, a query rewriter, and an observer, generating \mbox{58k} verified chain-of-thought trajectories.
Similarly, Chain-of-Agents~\cite{li2025chain} builds on the expert multi-agent system OAgents~\cite{zhu2025oagents} to synthesize task-solving trajectories, and after a four-stage filtering pipeline that removes trivial or incorrect cases, it yields 16,433 high-quality trajectories for agent training.
More recently, AgentFounder~\cite{su2025scaling} propose the agentic continual pre-training, which scales up the data generation process by constructing large-scale planning traces, tool-invocation sequences, and step-by-step reasoning data.

\header{Comparing Two Types of Distillation.}
Single-agent distillation provides a simple and easy-to-deploy pipeline, but it is limited by the bias of a single teacher model and the relatively shallow nature of its synthesized trajectories~\cite{ojha2023knowledge, lukasik2021teacher, zhu2022teach, nagarajan2023student}.
Such trajectories often emphasize token-level action sequences rather than higher-level reasoning, which can restrict the student model’s generalization ability in complex tasks.
In contrast, multi-agent distillation generates longer and more diverse trajectories that expand the action space to include strategic planning, task decomposition, iterative error correction, and self-reflection~\cite{zhou2025debate, chen2024magdi}.
This broader behavioral coverage equips student models with stronger capabilities for multi-step and knowledge-intensive reasoning~\cite{li2025chain}.

Despite these advantages, multi-agent distillation introduces notable trade-offs. The pipelines require careful system design, substantial inference cost, and dedicated infrastructure for logging and verification. Data quality can also be brittle as the system’s sensitivity to prompting~\cite{qiu2025agentdistill, reid2025risk, shen2025understanding}.

\subsubsection{Iterative Self-Evolving}\label{sec:interative-self-evolving}

\header{Definition.}
Iterative self-evolving data generation is an autonomous, cyclic process in which a model continuously generates new training data to fine-tune itself, progressively enhancing its capabilities~\cite{wang2022self,wang2023voyager,zhang2025evolvesearch,zhao2025absolute}.

\header{Representative Work.}
Early evidence that large language models can improve themselves comes from self-training methods~\cite{wang2022self, yuan2024self, chen2024self}, where a model bootstraps from a small set of seed tasks to synthesize instruction–input–output triples, filters the synthetic data, and then fine-tunes itself on the resulting corpus. These approaches deliver substantial gains in instruction following with minimal human supervision.
\citet{yuan2024self} further introduces self-rewarding language models, in which the model generates its own rewards through LLM-as-a-Judge prompting.
More recently, \citet{zhao2025absolute} extends this idea to the zero-data regime by framing self-play as an autonomous curriculum. The model creates code-style reasoning tasks, solves them, and relies on an external code executor as a verifiable environment to validate both tasks and solutions.
In the context of DR, EvolveSearch~\cite{zhang2025evolvesearch} iteratively selects high-performing rollouts (\ie task-solving trajectories) and re-optimizes the model on these data via supervised fine-tuning. 

\header{Advantages \& Disadvantages.}
A key advantage of iterative self-evolving frameworks is their closed-loop design, where the model progressively improves its capabilities by tightly interleaving data generation with training.
This autonomy enables scalable training without heavy reliance on external models or human annotations, and it allows exploration of data distributions that extend beyond handcrafted knowledge~\cite{wang2022self,zeng2024automatic,yuan2024self,silver2017mastering,fawzi2022discovering}.

However, self-improvement also introduces significant risks.
Previous studies have shown that, as iterations progress, distributional drift, reward hacking, and self-reinforcing errors may accumulate and degrade data quality, potentially leading to training collapse~\cite{shumailov2023curse,shumailov2024ai,herel2024collapse}.
In addition, without robust validation mechanisms, the process may converge prematurely to narrow modes with limited performance ceilings~\cite{alemohammad2024self,bertrand2023stability,dohmatob2024tale,briesch2023large}.

\subsection{End-to-End Agentic Reinforcement Learning}\label{sec:agentic-RL}

\header{Definition.}
In this section, we dive into the application of end-to-end agentic reinforcement learning (RL) in DR, \ie using RL algorithms to incentivize DR agents that can flexibly plan, act, and generate a final answer.
We start with a brief overview, including commonly used RL algorithms and reward design for optimizing DR systems.
For a clear explanation, we provide a glossary table in Table~\ref{tab:notion} to formally introduce the key variable in this section \ref{sec:agentic-RL}.
Then we discuss two training practices: (i) \textit{specific module optimization} and (ii) \textit{entire pipeline optimization}.

\input{table/notions}

\subsubsection{Preliminary}\label{preliminary}

\header{RL algorithms in Deep Research.}
In DR, LLMs are trained to act as autonomous agents that generate comprehensive reports through complex query decomposition, multi-step reasoning, and extensive tool use.  
The primary RL algorithms used to train these agents include Proximal Policy Optimization (PPO) from OpenAI~\cite{schulman2017proximal, openai2023chatgpt}, Group Relative Policy Optimization (GRPO) from DeepSeek~\cite{shao2024deepseekmath, guo2025deepseek}, and their variants~\citep{Yu2025DAPOAO}.

\header{Proximal Policy Optimization.}
PPO~\cite{schulman2017proximal} is a clipped policy-gradient method that constrains updates within a trust region~\cite{ouyang2022training}. 
Given a current policy $\pi_{\theta}$ and a old policy $\pi_{\theta_{\text{old}}}$, the objective is to maximize the clipped surrogate:
\begin{align}
L^{\text{PPO}}(\theta) &=
\mathbb{E}_t \!\left[
    \min\!\left(
        r_t(\theta) \hat{A}_t,\;
        \text{clip}\left(r_t(\theta), 1-\epsilon, 1+\epsilon\right)\! \hat{A}_t
    \right)
\right],\\
r_t(\theta) &= \frac{\pi_{\theta}(o^t \mid s_t)}{\pi_{\theta_{\text{old}}}(o^t \mid s_t)}.
\end{align}
where $\epsilon$ bounds the policy update and $\hat{A}_t$ is the estimated advantage. 
The advantage is computed using discounted returns or generalized advantage estimation (GAE)~\citep{Schulman2015HighDimensionalCC} as:
\begin{equation}
\hat{A}_t = \sum_{l=0}^{T-t} \gamma^l \cdot r_{t+l} + \gamma^{T-t+1} \cdot V_{\phi}(s_{T+1}) - V_{\phi}(s_t).
\end{equation}
Here $r_{t+l}$ denotes the immediate reward at time step $t+l$, 
$\gamma \in [0,1)$ is the discount factor balancing the importance of long-term and short-term returns; $T$ is the terminal time step of the current trajectory (episode);
$s_{T+1}$ is the next state used for bootstrapping after termination, 
$V_{\phi}(s_t)$ is the value function predicted by the value network parameterized by $\phi$.
We define the empirical return $\hat{R}_t$ purely from rewards as:
\begin{equation}
\hat{R}_t = \sum_{l=0}^{T-t} \gamma^l r_{t+l},
\end{equation}
which represents the cumulative discounted rewards from time step $t$ until the end of the episode.
In PPO, the value function parameters $\phi$ are updated by minimizing the squared error between the predicted value and the empirical return:
\begin{equation}
\mathcal{L}^{\text{value}}(\phi) = \frac{1}{2} \, \mathbb{E}_t \left[ \left( V_\phi(s_t) - \hat{R}_t \right)^2 \right].
\end{equation}

\header{Group Relative Policy Optimization.}
Group Relative Policy Optimization (GRPO)~\cite{shao2024deepseekmath} extends PPO by normalizing rewards \emph{within groups of responses} to the same query. 
Formally, given a group $\mathcal{G}$ of $m$ responses $\{o_1, o_2, \dots, o_m\}$ sampled for the same query $s_t$, each response is assigned a scalar reward $R_j$.  
The \emph{group-relative advantage} for the $j$-th response is:
\begin{equation}
\hat{A}^{\mathcal{G}}_j = 
\frac{\mathcal{R}_j - \text{mean}(\{\mathcal{R}_i \mid i \in [m] \})}{\text{std}(\{\mathcal{R}_i \mid i \in [m] \}) + \epsilon},
\end{equation}
where $\text{mean}_\mathcal{G}$ and $\text{std}_\mathcal{G}$ denote the mean and standard deviation of rewards within group $\mathcal{G}$, and $\epsilon$ prevents numerical instability when the variance is small.
The GRPO objective mirrors PPO's clipping mechanism but replaces $\hat{A}^{\mathcal{G}}_t$ with the group-relative advantage $\hat{A}^\mathcal{G}_j$:
\begin{equation}\small
\mathcal{L}^{\text{GRPO}}(\theta) =
\mathbb{E} \!\left[
\frac{1}{|\mathcal{G}|} \sum_{j=1}^{|\mathcal{G}|}
\min\!\Big\{
\frac{\pi_{\theta}(o_j \mid q)}{\pi_{\theta_{\text{old}}}(o_j \mid q)} \hat{A}^\mathcal{G}_j,\;
\text{clip} \Big(
\frac{\pi_{\theta}(o_j \mid q)}{\pi_{\theta_{\text{old}}}(o_j \mid q)},
1-\epsilon, 1+\epsilon
\Big)\! \hat{A}^\mathcal{G}_j
\Big\}
\right].
\end{equation}

\header{Comparison between PPO and GRPO in Deep Research.}
In PPO, each sampled output is optimized using an advantage signal derived from a value model. While this approach is effective, its performance is highly reliant on accurate value estimation and requires additional resources for training the value model. 
In contrast, GRPO optimizes by contrasting each response against others within the same group. This shifts the focus to a relative-quality comparison among competing hypotheses, simplifying implementation while maintaining strong performance.


\header{Reward Design in Deep Research Agents.}
During the RL training of DR agents, the reward model, denoted as $\mathcal{R}(\cdot)$, assesses the quality (\eg  correctness) of the agents' outputs and produces scalar signals to enable policy optimization algorithms such as PPO and GRPO. 
Reward design takes a critical role in training LLM.
There are two common reward design paradigms in DR systems, \ie \emph{rule-based rewards} and \emph{LLM-as-judge rewards}.

\begin{itemize}
    \item \textit{Rule-based Rewards $\mathcal{R}_{\text{rule}}(\cdot)$.} Rule-based rewards are derived from deterministic, task-specific metrics such as Exact Match (EM) and F1 score~\cite{schutze2008introduction}. In the context of research agents, EM is a commonly used binary score that indicates whether a generated answer perfectly matches a ground-truth string~\citep{karpukhin2020dense,jin2025search,jin2025empirical}. Alternatively, the F1 score (\ie the harmonic mean of precision and recall calculated over token overlap) is also used to reward outputs~\cite{chen2025improving, chen2025mao}. However, a key limitation of rule-based rewards is that they are primarily suited for tasks with well-defined, short-span ground truths (\eg  a specific entity name) and struggle to evaluate multi-answer or open-ended questions effectively.

    \item \textit{LLM-as-judge Rewards $\mathcal{R}_{\text{LLMs}}(\cdot)$.} 
    The LLM-as-judge approach uses an external LLM to evaluate the quality of an agent's output and assign a scalar score based on a predefined rubric~\citep{Arora2025HealthBenchEL}.
    Formally, for an output $o$ to an input query $q$, the reward assigned by an LLM judge $\phi$ can be formulated as:
    \begin{equation*}
        \mathcal{R}_{\text{LLMs}}(o \mid q) = 
        \mathbb{E}_{\text{criteria} \in \mathcal{C}}
        \left[ \phi(o, q, \text{criteria}) \right]
    \end{equation*}
    where $\mathcal{C}$ is the set of evaluation criteria (\eg  accuracy, completeness, citation quality, clarity, etc) and $\phi(\cdot)$ returns a scalar score for each criterion.  
        
\end{itemize}




\subsubsection{End-to-end Optimization of a Specific Module}\label{subsec:specific_module_optimization}

\header{Definition.}
End-to-end optimization of a specific module focuses on applying RL techniques to improve individual components within a DR system, such as the query planning, document ranking, or planning modules.

\header{Representative work.}
Within DR, most existing work trains the query planner~\cite{jiang2025s3,zhu2025planner,zhu2025convsearch,liu2025opera} while freezing the parameters, leaving components such as retrieval.
MAO-ARAG~\cite{chen2025mao} treats DR as a multi-turn process where a planning agent orchestrates sub-agents for information seeking.
PPO propagates a holistic reward (\eg final F1 minus token and latency penalties) across all steps, enabling end-to-end learning of the trade-offs between answer quality and computational cost.
AI-SearchPlanner~\cite{mei2025ai} decouples a lightweight search planner from a frozen QA generator. PPO optimizes the planner with dual rewards: an outcome reward for improving answer quality and a process reward for reasoning rationality. A Pareto-regularized objective balances utility with real-world cost, guiding the planner on when to query or stop.

\header{Advantages \& Disadvantages.}
Single-module optimization usually focuses on training a single core component (\eg  the planning module) while keeping the others fixed.
Optimizing this critical module can improve the performance of a DR system by enabling more accurate credit assignment, more sophisticated algorithm design for the target module, and reduced training data and computational costs. 
However, this approach restricts the optimization space and may be inadequate when other frozen modules contain significant design or performance flaws.

\subsubsection{End-to-end Optimization of an Entire Pipeline}\label{subsec:entire_pipeline_optimization}

\header{Definition.}
End-to-end pipeline optimization involves jointly optimizing all components and processes from input to output (\eg  query decomposition, search, reading, and report generation) to achieve the best overall performance across the DR workflow.

\header{Representative work on Multi-Hop Search.}
Some work focuses on enhancing the capability of multi-hop search by training the entire DR systems end-to-end~\cite{peng2024graph,song2025r1,song2025r1++,zheng2025deepresearcher,wu2025mmsearch}.
For example, \citet{jin2025search, jin2025empirical} present Search-R1, the first work to formulate search-augmented reasoning as a fully observable Markov Decision Process and to optimize the entire pipeline via RL, containing query planning, retrieval, and extracting the final answer.
By masking retrieved tokens in the policy-gradient loss, the model learns to autonomously decide when and what to search while keeping the training signal on its own generated tokens.
Meanwhile, \citet{song2025r1} introduces R1-Searcher, a two-stage RL method in which the DR agent learns when to invoke external searches and how to use retrieved knowledge via outcome rewards.
However, it has been observed that pure RL training often leads to over-reliance on external retrieval, resulting in over-searching~\cite{song2025r1++}.
To mitigate this issue, R1-Searcher++~\cite{song2025r1++} first cold-starts the DR agent via an SFT, then applies a knowledge-assimilation RL process to encourage the agent to internalize previously retrieved documents and avoid redundant retrievals.

Besides the above early effort, recent work extends the naive Search-R1 by integrating multi-reward signals or improving the training environment.
\textit{For the reward design}, R-Search~\cite{zhao2025r} trains models to decide when to retrieve and how to integrate external knowledge in both single-hop and multi-hop question answering. The framework improves answer quality and evidence reliability by optimizing reasoning–search trajectories under a multi-objective reward design.
\textit{For the training environment}, ZeroSearch~\cite{sun2025zerosearch} and $\textit{O}^2$-Searcher~\cite{mei20252} simulate a search engine to develop retrieval capabilities without accessing the actual web, providing a more controllable setting for RL training.
In contrast, DeepResearcher~\cite{zheng2025deepresearcher} operates directly in real-world web environments, learning to plan, search, verify, and autonomously answer open-domain questions.

Besides a basic document retrieval tool, some work also integrates additional information-seeking tools, teaching the DR agent to flexibly combine them.
MMSearch-R1~\cite{wu2025mmsearch} stands out as the first RL-trained multimodal model that learns when and how to search the text or image from the web on demand. 
HierSearch~\cite{tan2025hiersearch} introduces a hierarchical DR framework for enterprise scenarios that involve both local and web knowledge sources.
Other work~\cite{endgraph,peng2024graph,hao2025dynasearcher} integrates knowledge graphs into DR agents to achieve efficient multi-hop reasoning.

\header{Representative work on Long-Chain Web Search.}
Besides the relatively simple multi-hop QA tasks, more recent work also applies end-to-end pipeline optimization to address longer-chain web search problems.
Prior works, such as WebDancer~\cite{wu2025webdancer}, WebSailor~\cite{li2025websailor}, and Kimi-K2~\cite{team2025kimi}, have focused on advancing more intricate multi-hop tasks, including GAIA~\cite{mialon2023gaia} and BrowseComp~\cite{wei2025browsecomp}. These approaches combine data synthesis with end-to-end reinforcement learning training, thereby enabling more extensive iterations in the DR process.

Furthermore, \citet{gao2025beyond} presents ASearcher, which scales end-to-end RL to extreme long-horizon search. 
A fully asynchronous RL engine removes the 10-turn ceiling that plagued earlier systems, allowing trajectories of 40+ turns and 150k tokens to be optimized without blocking GPU updates.  Coupled with an autonomous QA-synthesis agent that injects noise and fuzzes questions for difficulty, the whole pipeline is operated end-to-end, from synthetic data creation to multi-turn policy optimization.
SimpleDeepSearcher~\cite{sun2025simpledeepsearcher} leverages real-web simulation and distilled SFT to deliver agentic search capability without heavy RL, yet stays fully compatible with lightweight RL refinement. 
WebAgent-R1~\cite{wei2025webagent} and DeepDiver~\cite{shi2025pangu} are training web agents through end-to-end multi-turn RL algorithms.

In addition, some works~\cite{li2025chain,dong2025tool,dong2025agentic,dong2025agentic2} have studied Deep Research systems across multiple tool-calling scenarios. For example, \citet{li2025chain} introduces Chain-of-Agents (CoA), a novel paradigm that distills the capabilities of multi-agent systems into a single LLM. CoA enables native, end-to-end complex problem-solving by dynamically orchestrating multiple role-playing and tool agents within one model. Through multi-agent distillation and agentic RL, the authors train Agent Foundation Models (AFMs) in an end-to-end approach that achieves excellent performance on diverse web search benchmarks, while significantly reducing computational overhead compared to traditional multi-agent systems. 
Tool-Star~\cite{dong2025tool} and ARPO~\cite{dong2025agentic} investigate how to effectively leverage tools in long-horizon tasks such as Deep Research, and use the GRPO algorithm to optimize the entire pipeline end-to-end. Additionally, AEPO~\cite{dong2025agentic2} further improves rollout efficiency and, based on ARPO, optimizes both performance and efficiency for the end-to-end tool-use pipeline.

\header{Advantages \& Disadvantages.}
These end-to-end methods model the entire DR system as a multi-turn search process, achieving comprehensive optimization across reasoning, query rewriting, knowledge retrieval, tool invocation, and answer generation. This modeling and optimization approach is not only flexible but also allows for different objectives to be emphasized through the design of reward functions. However, these methods also have drawbacks, including sparse rewards, excessively long responses, and unstable training. Continuous optimization is needed to further enhance the effectiveness, stability, and efficiency of DR systems.

\begin{tcolorbox}[takeawaysbox, title={Takeaway}]
\begin{itemize}
    \item \textbf{RL Algorithms}: PPO provides stable updates based on absolute rewards, while GRPO leverages group-relative advantages to reduce resource requirements.
    \item \textbf{Specific Module End-to-End Optimization}: Targets a critical component (\eg  planner or searcher) for RL training, improving overall performance at lower cost, though limitations in other frozen modules cannot be addressed.
    \item \textbf{Entire Pipeline End-to-End Optimization}: Optimizes the full DR workflow, including retrieval, reasoning, tool use, and answer generation, yielding holistic gains but facing sparse rewards, long outputs, and training instability.
\end{itemize}
\end{tcolorbox}

%% file: table/notions.tex
\begin{table}[!t]
\centering
\rowcolors{2}{gray!10}{white}
\caption{Summary of key notations used in proximal policy optimization and group-relative policy optimization algorithms.}
\label{tab:notations}
\begin{tabular}{m{1.5cm} m{3.7cm} m{9.8cm}}
\toprule
\textbf{Symbol} & \textbf{Definition} & \textbf{Description} \\
\midrule
$\pi_{\theta}$ & Current policy & Parameterized LLM policy that generates actions (tokens or sequences) conditioned on a given state. \\
$\pi_{\theta_{\text{old}}}$ & Reference (old) policy & A frozen snapshot of the policy before the current update, used for computing probability ratios and ensuring stable optimization. \\
$q$ & Input query & Input question or prompt to the agent. \\
$o$ & Model output & Final answer produced by the policy model. \\
$o^t$ & Action at step $t$ & The token generated by the policy model conditioned on state $s_t$. \\
$s_t$ & State at step $t$ & Context of the policy model at time step $t$. \\
$\mathcal{R}(o | \cdot)$ & Reward function & Scalar score assigned to output $o$ for the input query $q$. \\
$r_t(\theta)$ & Probability ratio & Ratio between current and reference policy probabilities, computed as $\frac{\pi_{\theta}(o^t|s_t)}{\pi_{\theta_{\text{old}}}(o^t|s_t)}$. \\
$\epsilon$ & Clipping threshold & Stability constant that limits update magnitude in PPO or adds numerical robustness in GRPO. \\
\midrule
$\mathcal{G}$ & Response group & A collection of multiple sampled responses corresponding to the same query $s_t$ in GRPO. \\
$m$ & Group size & The number of candidate responses in a response group $\mathcal{G}$. \\
$o_j$ & $j$-th response in group & The $j$-th sampled output candidate among the $m$ responses in group $\mathcal{G}$. \\
\bottomrule
\end{tabular}\label{tab:notion}
\end{table}



%% file: sections/05-evaluation.tex
\section{Evaluation of Deep Research System}\label{sec:evaluation}

DR techniques have been applied to a wide range of downstream tasks, including healthcare~\citep{Arora2025HealthBenchEL}, financial report generation~\citep{tian2025templatebasedfinancialreportgeneration}, and survey generation~\citep{wang2024autosurvey}.
In this section, we systematically review common benchmarks and evaluation protocols for DR systems across three representative scenarios: (i) \textit{information seeking}, (ii) \textit{report generation}, and 
(iii) \textit{AI for research}.
These scenarios reflect the most prevalent applications of DR agents.
Each category poses distinct challenges, illuminating the limitations of current systems while offering practical insights to guide future advances.

\subsection{Agentic Information Seeking}\label{agentic-information-seeking}

Evaluating the effectiveness of agentic information-seeking is a critical component of assessing DR systems.
In DR scenarios, information seeking is not a single QA task but a multi-stage, iterative, and cross-domain process in which agents must continuously explore, reformulate, and synthesize information from diverse sources.
To capture this complexity, benchmark design has evolved from early static single-hop retrieval tasks such as Natural Questions (NQ)~\cite{kwiatkowski2019natural} to dynamic web environments requiring multi-hop reasoning and complex interactions, \eg BrowseComp~\cite {wei2025browsecomp} and HotpotQA~\cite{yang2018hotpotqa}. 
In this section, we review representative benchmarks and evaluation frameworks along two dimensions: query complexity and interaction environment complexity.

\input{table/resource_QA}

\subsubsection{Complex Queries}\label{complex-queries}
The evolution of benchmarks for agentic information seeking has closely followed the increasing complexity of query demands.
Early benchmarks such as NQ~\cite{kwiatkowski2019natural}, TriviaQA~\cite{joshi2017triviaqa}, and SimpleQA~\cite{wei2024measuring} established the foundation for question answering research. These datasets focused on single-hop queries, where answers could be retrieved with a single lookup or were already contained within the LLM’s parameters. While such tasks provided a controlled starting point, they could not capture the reasoning and synthesis required in DR.

\input{table/resource}

As research questions grew more complex, benchmarks evolved from simple fact retrieval to multi-step reasoning challenges. Multi-hop QA datasets assess an agent’s ability to reformulate queries and build reasoning chains across documents. HotpotQA~\cite{yang2018hotpotqa}, one of the earliest and most widely used multi-hop datasets, requires reasoning across multiple Wikipedia articles using supporting facts to derive the answer. 2WikiMultihopQA~\cite{ho2020constructing} extends this by integrating information from two separate Wikipedia pages per question, emphasizing cross-document reasoning.
Bamboogle~\cite{press2023measuring} consists of 125 two-hop questions generated from random Wikipedia articles, testing the ability to decompose and reason over complex queries. MultiHop-RAG~\cite{tang2024multihop} is the RAG dataset designed specifically for multi-hop queries, categorizing questions into four types: inference, comparison, temporal, and null queries.
MuSiQue~\cite{trivedi2022musique} adopts a bottom-up approach, systematically pairing composable single-hop questions where one reasoning step depends on another. FRAMES~\cite{krishna2025fact} simulates realistic multi-document queries to evaluate an LLM’s ability to retrieve relevant facts, reason accurately, and synthesize information into coherent responses.
However, most of these benchmarks rely on structured, linear reasoning paths, which fall short of reflecting the inherent ambiguity and branching, non-linear exploration required in real-world research scenarios.

Recent benchmarks have begun to capture this growing complexity, placing greater emphasis on the in-depth and progressive exploration of complex topics. For instance, GPQA~\cite{rein2024gpqa} is a graduate-level dataset in physics, chemistry, and biology that tests both domain experts and skilled non-experts, requiring extensive reasoning and problem-solving. Similarly, GAIA~\cite{mialon2023gaia} provides 466 carefully designed questions that require multi-step reasoning, real-world knowledge retrieval, and complex generation.
HLE~\cite{phan2025humanity} aims to be a comprehensive, fully closed academic benchmark across dozens of disciplines, including mathematics, humanities, and natural sciences, designed to advance reasoning skills. Its questions cannot be quickly answered through an online search. These recent datasets challenge agents to operate in environments that better reflect the ambiguity, branching evidence paths, and iterative synthesis characteristic of real-world DR systems.

\subsubsection{Interaction Environment}\label{interation-environment}

As agent capabilities have advanced, evaluation based solely on static environments and fixed corpora is no longer sufficient. Consequently, a series of benchmarks has been developed to reflect the scale and dynamics of real-world web environments, requiring agents to interact with, navigate, and creatively explore web pages to obtain complex or hard-to-find information.

Some studies have incorporated browsing tools such as Google and Bing into benchmarks, enabling agents to directly retrieve and extract information from the live web. For example, InfoDeepSeek~\cite{xi2025infodeepseek} and AssistantBench~\cite{yoran2024assistantbench} present challenging tasks that require agents to integrate multiple search and browsing tools in real-time web environments, testing their ability to operate dynamically.
Mind2Web~\cite{deng2023mind2web} replaces the overly simplified, simulated environments common in other datasets with authentic, dynamic, and unpredictable real-world websites, providing complete records of user interactions, webpage snapshots, and network traffic.
Its successor, Mind2Web 2~\cite{gou2025mind2web}, was subsequently introduced to more rigorously evaluate agent-based search systems on realistic, long-horizon tasks that involve live web search and browsing.
BrowseComp~\cite{wei2025browsecomp} and BrowseComp-Plus~\cite{chen2025browsecomp} demand persistent navigation to locate hard-to-find, entangled information across multiple sites.
Moreover, DeepResearchBench~\cite{bosse2025deep} offers a large-scale RetroSearch environment that reduces task degradation and network randomness while evaluating LLM agents on complex real-world web research tasks. DeepResearchGym~\cite{coelho2025deepresearchgym} complements this by providing an open-source sandbox with a reproducible search API and a rigorous evaluation protocol, promoting transparency and reproducibility in DR area.

Building on this trend, subsequent datasets have increasingly emphasized the authenticity and complexity of interactive environments.
WebArena~\cite{zhouwebarena} provides a highly realistic and reproducible environment for language-guided agents, built from fully functional websites across four domains. WebWalkerQA~\cite{wu2025webwalker} assesses LLMs’ ability to systematically traverse website subpages and extract high-quality data through interactive actions such as clicking, specifically testing complex, multi-step web interactions. WideSearch~\cite{wong2025widesearch} focuses on a critical yet under-evaluated task: requiring agents to thoroughly and accurately acquire all large-scale atomic information that meets a set of criteria and organize it into a structured output. MMInA~\cite{tian2024mmina} extends these challenges by providing a multi-hop, multi-modal benchmark for embodied agents performing integrated internet tasks on realistic, evolving websites, ensuring high realism and applicability to natural user tasks.
Together, these benchmarks illustrate a clear trend: web-oriented evaluation environments are becoming increasingly human-like, visually grounded, diverse, complex, and realistic, pushing the limits of agentic information seeking and DR in dynamic, real-world settings.

\subsection{Comprehensive Report Generation}\label{comprehensive-report-generation}

Another critical dimension in evaluating DR systems is their capacity to generate comprehensive reports. 
Unlike single-point answers or brief summaries, comprehensive reports require systems to integrate information from multiple sources and modalities into structured, logically coherent, and broadly informative outputs. 
This process involves information aggregation, content organization, factual consistency verification, and clarity of expression, and is therefore regarded as a core indicator of a DR system’s overall capability. Below, we introduce the relevant benchmarks by task type.

\subsubsection{Survey Generation}\label{survey-generation}
A closely related task is survey generation, which involves producing structured overviews or syntheses of a specific scientific topic by aggregating information from diverse sources. Thanks to the clear citation structure provided by gold-standard references, survey generation has been widely used to evaluate the capabilities of DR systems. AutoSurvey~\cite{wang2024autosurvey} gathers arXiv articles of varying lengths and uses a multi-LLM-as-judge framework to evaluate survey generation in terms of speed, citation quality, and content quality. Moreover, ReportBench~\cite{li2025reportbenchevaluatingdeepresearch} is a systematic benchmark for evaluating research reports generated by large language models. It focuses on two key aspects: the relevance of citations and the reliability and accuracy of the report’s statements. The evaluation corpus is constructed using high-quality survey papers published on arXiv as the gold standard.
SurveyGen~\cite{bao2025surveygenqualityawarescientificsurvey} is another survey-generation dataset, containing over 4,200 human-written surveys with chapter-level structure, cited references, and rich metadata. It enables comprehensive evaluation of content quality, citation accuracy, and structural consistency.

\subsubsection{Long-Form Report Generation}\label{long-form-report-generation}
Other benchmarks focus on different types of report generation tasks and introduce alternative evaluation frameworks. For example, Deep Research Comparator~\cite{chandrahasan2025deep} provides a unified evaluation platform for DR agents, enabling systematic assessment of long-form reports and their intermediate reasoning processes through side-by-side comparison, fine-grained human feedback, and ranking mechanisms.
DeepResearch Bench~\cite{du2025deepresearch} is a benchmark of 100 PhD-level research tasks, introducing two evaluation methods for generated reports: a reference-based assessment of overall quality and a citation-based evaluation of retrieval accuracy.
ResearcherBench~\cite{xu2025researcherbench} comprises 65 research questions focused on evaluating the capabilities of advanced agent systems on cutting-edge AI science problems, using an evaluation framework that combines rubric assessment and factual evaluation.
LiveDRBench~\cite{java2025characterizing} is a benchmark for DR tasks, offering challenging science and world-event queries and evaluating systems via intermediate reasoning steps and factual sub-propositions. 
PROXYQA~\cite{tan2024proxyqa} uses human-designed meta-questions and corresponding proxy questions to indirectly assess knowledge coverage and information richness, providing an objective measure of long-form text generation quality.
SCHOLARQABENCH~\cite{asai2024openscholar} is a benchmark for scientific literature synthesis tasks in multiple formats, comprising 2,967 expert-written queries and 208 long-form answers across the field of computer science.
Evaluating research reports is particularly challenging because there is no single gold-standard answer, and multiple valid perspectives exist for assessing quality. The diversity of acceptable content, reasoning approaches, and presentation styles makes it difficult to define objective metrics. As a result, most benchmarks rely on LLM-as-judge methods~\cite{li2024llms}, leveraging large language models' reasoning and knowledge to provide scalable, consistent, and nuanced evaluations of content quality, factual accuracy, citation relevance, and structural coherence.

\subsubsection{Poster Generation}\label{poster-generation}
Poster generation can be viewed as a highly condensed and visually structured variant of the comprehensive report generation task.
Unlike multi-page reports, a scientific poster is typically a single-page, high-density summary that concisely presents the research motivation, methodology, results, and conclusions in a format that is both navigable and visually engaging. For DR systems, this task imposes unique challenges: not only must the system aggregate and synthesize information from multiple heterogeneous sources, such as research papers, notes, and presentation slides, but it must also transform that content into an effective visual layout.
Evaluation of poster generation typically focuses on three main aspects: factual completeness, visual communication effectiveness, and readability.
For example, Paper2Poster~\cite{pang2025paper2poster} focuses exclusively on AI research papers. The dataset comprises 100 paper-poster pairs, covering 280 distinct topics across subfields. A comprehensive evaluation framework is introduced, comprising four key dimensions: visual quality, text coherence, VLM-based quality judgment, and PaperQuiz, which is a novel metric designed to assess how effectively a poster communicates the core knowledge of the original paper.
PosterGen~\cite{zhang2025postergen} adopts a two-dimensional evaluation protocol, dividing the assessment into poster content and poster design. It introduces a VLM-based metric to evaluate key design aspects, including layout balance, readability, and aesthetic consistency. P2PInstruct~\cite{sun2025p2p} is a large-scale instruction dataset for paper-to-poster generation, containing over 30,000 high-quality instruction–response pairs. It covers the full pipeline from image element processing and text generation to final layout formatting.

\subsubsection{Slides Generation}\label{slides-generation}
Slide generation represents another critical pathway for evaluating the comprehensive capabilities of DR systems. This task challenges a system not only to comprehend and summarize large volumes of heterogeneous information sources, but also to transform the distilled content into a structured, slide-based presentation format. The objectives of slide generation encompass information distillation, logical structuring, and presentation-oriented expression, making it a strong indicator of a system’s ability to coordinate across multiple dimensions.
Common benchmark tasks and datasets for this evaluation often involve generating slides from meeting transcripts, research papers, long-form reports, or collections of web documents. These benchmarks typically assess whether a model can maintain content integrity and factual accuracy while performing high-quality information selection and organization.
Doc2PPT~\cite{fu2022doc2ppt} collects paired documents and their corresponding slide decks from academic proceedings. It conducts detailed post-processing for evaluation, using metrics such as Slide-Level ROUGE, Longest Common Figure Subsequence, Text-Figure Relevance, and Mean Intersection over Union. For Example, SLIDESBENCH~\cite{ge2025autopresent} is a benchmark comprising 7,000 training examples and 585 test examples, derived from 310 slide decks across 10 distinct domains. It supports two types of evaluation: reference-based evaluation, which measures similarity to target (gold) slides, and reference-free evaluation, which assesses the design quality of the generated slides independently.
Zhang et al. introduced Zenodo10K~\cite{zheng2025pptagent}, a new dataset collected from Zenodo, a platform hosting diverse, openly licensed artifacts across various domains. They also proposed PPTEval~\cite{zheng2025pptagent}, an evaluation framework leveraging GPT-4.0 as the judge to assess presentation quality along three dimensions: content, design, and coherence.
TSBench~\cite{jung2025talk} is a benchmark dataset specifically designed to evaluate the slide editing capabilities of models and frameworks. It includes 379 distinct editing instructions along with the corresponding slide modifications.

\subsection{AI for Research}\label{sec:ai-for-research}
AI for Research seeks to harness artificial intelligence to advance scientific discovery, either by automating processes or by assisting researchers in accelerating their work~\citep{chen2025ai4research}. Its applications and corresponding benchmarks include (i) \textit{idea generation}, (ii) \textit{experimental execution}, (iii) \textit{academic writing}, and (iv) \textit{peer review}. 
Unlike report generation, research goes beyond producing extended outputs; it requires the creation of new perspectives, conclusions, and knowledge, thereby necessitating mechanisms for evaluating novelty.

\subsubsection{Idea Generation}\label{idea-generation}
A key challenge in research lies in generating genuinely novel ideas and, more importantly, in reliably assessing their novelty.
Such evaluation is typically conducted by human experts, but it remains difficult and resource-intensive. 
Existing approaches generally fall into two categories. 
The first is human- or LLM-based evaluation. 
\citet{si2024can} recruited over 100 NLP researchers to evaluate the novelty of ideas generated by humans, LLMs, and human–LLM collaboration. However, this process is not easily scalable and proves difficult even for domain experts. 
Moreover, they investigated LLMs’ ability to assess novelty, finding that LLM judgments show lower agreement with expert reviewers than human evaluations. 
\citet{li2024learning} and \citet{gao2025graph} leverage LLMs through direct prompting to evaluate novelty. 
To enhance the reliability of LLMs’ judgments, \citet{lu2024ai} and \citet{su2024two} integrate LLMs with the Semantic Scholar API and web access, enabling them to evaluate ideas against related literature. 
AI Idea Bench 2025~\citep{qiu2025ai} provides a benchmark for quantifying and comparing ideas generated by LLMs. 
It incorporates 3,495 representative papers published in AI-related conferences after October 10, 2023, together with their corresponding inspiration papers. Furthermore, it introduces an evaluation framework to assess whether the ideas derived from inspiration papers are consistent with the ground-truth papers. The second category is density-based evaluation, which relies on the absolute local density in the semantic embedding space to measure novelty. \citet{wang2025enabling} introduces the Relative Neighbor Density (RND) algorithm, which evaluates novelty by examining the distributional patterns of semantic neighbors rather than relying solely on absolute local density. Moreover, they construct large-scale semantic embedding databases for novelty assessment, encompassing more than 30 million publications across two distinct domains.

\subsubsection{Experimental Execution}\label{experimental-execution}
Evaluation of experimental execution typically involves both objective and subjective assessments. Objective evaluation generally emphasizes the outcomes produced in specific environments, such as benchmark performance or compiler outputs. For example, \citet{lu2024ai} and \citet{weng2025deepscientist} directly adopt the results of downstream tasks as evaluation metrics, while \citet{tang2025ai} leverages compiler outputs to assist LLMs in refining experiments and correcting code errors.
Subjective evaluation involves leveraging either humans or LLMs to assess the quality of experimental designs. For example, \citet{tang2025ai} employs LLMs to compare code implementations with atomic research ideas, thereby verifying whether the code satisfies the intended requirements of the ideas. \citet{starace2025paperbench} employs LLMs to evaluate source code, documentation, and configuration files against human-designed rubrics to derive a final grade.

\subsubsection{Academic Writing}\label{academic-writing}
The evaluation of academic writing differs substantially from general report generation.
It requires not only factual accuracy, logical coherence, and clarity, but also alignment with underlying ideas and experimental results, proper integration of citations from related work, and effective visualization of findings.
To capture these multifaceted criteria, \citet{lu2024ai} employ LLMs to assess writing along dimensions such as originality, quality, clarity, and significance.
Similarly, \citet{hopner2025automatic} train a domain-specific LLM to predict citation counts and review scores as proxies for paper quality, addressing the limitations of generic LLMs in academic evaluation.
Building on this direction, \citet{starace2025paperbench} introduce PaperBench, a benchmark designed to assess AI agents’ ability to replicate AI papers.
The benchmark includes 20 ICML 2024 Spotlight and Oral papers, and evaluates replication quality using LLMs guided by manually constructed rubrics that hierarchically decompose each task into graded subtasks.
\citet{tang2025ai} propose Scientist-Bench, a comprehensive benchmark built from top-cited papers published between 2022 and 2024 across 16 research areas and multiple expertise levels.
It evaluates dimensions such as technical novelty, methodological rigor, empirical validation, and potential impact, closely reflecting the criteria used in the ICLR review process.
To further push the boundaries of scientific evaluation, \citet{xu2025researcherbench} present ResearcherBench, a more challenging benchmark for evaluating DR systems, which consists of 65 research questions carefully curated from real-world scientific contexts across 35 AI subfields.
They also propose a dual evaluation framework that combines rubric-based assessment to evaluate the quality of insights with factual evaluation that measures citation faithfulness and evidence coverage.

\subsubsection{Peer Review}\label{peer-reviewing}

AI for peer review seeks to leverage an AI agent to generate feedback on scientific papers. However, evaluating such feedback is challenging, as reviews are typically lengthy and inherently subjective. \citet{yuan2022can} introduced ASAP-Review, a large-scale benchmark that collects 8,877 AI papers from ICLR (2017–2022) via OpenReview and NeurIPS papers (2016–2019) via the official proceedings, along with their corresponding reviews. The dataset is annotated across multiple dimensions, including Motivation, Originality, Soundness, Substance, Replicability, Clarity, and Comparison. To evaluate generated reviews, they employ automatic metrics such as ROUGE and BERTScore~\citep{zhang2019bertscore}, as well as human judgments. \citet{lu2024ai} collect 500 ICLR 2022 papers from OpenReview to establish a benchmark for the peer review task, subsequently employing self-reflection, few-shot examples, and response ensembling with LLMs to assess the quality of the generated reviews. \citet{weng2024cycleresearcher} introduces the REVIEW-5k dataset, which contains 782 test samples collected from ICLR 2024. Each sample includes the paper title, abstract, LaTeX or Markdown source, and the corresponding review comments. The dataset also provides structured review information, including summaries, strengths and weaknesses, clarification questions, and review scores. For evaluation, they employ Proxy Mean Squared Error (Proxy MSE) and Proxy Mean Absolute Error (Proxy MAE), which leverage multiple independent reviews of the same submission as unbiased estimators of its true rating~\citep{su2025icml}. Similarly, \citet{gao2024reviewer2} constructed REVIEWER2, a dataset comprising 27,805 papers and reviews collected from CONLL-16, ACL-17, COLING-20, ARR-22, ICLR-17–23, and NeurIPS-16–22. \citet{zhu-etal-2025-deepreview} introduces DeepReview-Bench, a dataset of 1.2K ICLR 2024–2025 submissions collected from OpenReview. The dataset includes textual reviewer assessments, interactive rebuttal-stage discussions, and standardized scoring information. For quantitative evaluation, they employ MAE, MSE, accuracy, F1, and Spearman correlation, while qualitative evaluation is conducted under the LLM-as-a-judge paradigm~\citep{li2024llms, gu2024survey, li2023generative} across five dimensions: constructive value, analytical depth, plausibility, technical accuracy, and overall quality.

\subsection{Software Engineering}

In addition to the above scenarios, DR agents can also be applied to software engineering, representing a shift from assisting with isolated code snippets to autonomously executing complex software development tasks~\citep{liu2024large}.
A pioneering work is SWE-Bench~\citep{jimenez2023swe}, a benchmark designed to evaluate whether AI agents can resolve real-world GitHub issues.
Although SWE-Bench does not yet cover full end-to-end software development, it marks an important step toward bridging idealized benchmarks with practical scenarios.
Meanwhile, DR agents have been deployed in a wide range of complex software engineering domains, including scientific discovery~\citep{wang2022scienceworld, wang2024can, jansen2024discoveryworld, fan2025megascience, siegel2024core, RomeraParedes2023MathematicalDF, Sun2025COBenchBL}, machine learning experimentation~\citep{huang2023mlagentbench, chan2024mle, wijk2024re, Li2025TowardsCA}, data science~\citep{jing2024dsbench, cao2024spider2, zhang2024benchmarking}, earth observation~\citep{kao2025towards}, and software library completion~\citep{zhao2024commit0, Ouyang2025KernelBenchCL}.

%% file: table/resource_QA.tex
\begin{table}[H]
\scriptsize
\setlength{\tabcolsep}{3pt}
\rowcolors{2}{gray!10}{white}
\centering
\caption{Comprehensive overview of existing and emerging benchmarks for Deep Research Systems that focus on question answering scenarios.}
\label{tab:benchmark-qa}
\begin{tabularx}{\textwidth}{l l X l X}
\toprule
\textbf{Benchmark (with link)} & \textbf{Date} & \textbf{Aspect} & \textbf{Data size (train/dev/test)} & \textbf{Evaluation metrics} \\
\midrule
\href{https://ai.google.com/research/NaturalQuestions/}{NQ} & 2019 & QA & 307373/7830/7842 & Exact Match / F1 / Accuracy \\
\href{https://github.com/openai/simple-evals}{SimpleQA} & 2024 & QA & 4,326 & Exact Match / F1 / Accuracy \\
\href{https://github.com/hotpotqa/hotpot}{HotpotQA} & 2019 & QA & 90124 / 5617 / 5813 & Exact Match / F1 / Accuracy \\
\href{https://github.com/Alab-NII/2wikimultihop}{2WikiMultihopQA} & 2020 & QA & 167454/12576/12576 & Exact Match / F1 / Accuracy \\
\href{https://github.com/ofirpress/}{Bamboogle} & 2023 & QA & 8600 & Exact Match / F1 / Accuracy \\
\href{https://github.com/yixuantt/MultiHop-RAG}{MultiHop-RAG} & 2024 & QA & 2556 & Exact Match / F1 / Accuracy \\
\href{https://github.com/stonybrooknlp/musique}{MuSiQue} & 2022 & QA & 25K & Exact Match / F1 / Accuracy \\
\href{https://github.com/idavidrein/gpqa/}{GPQA} & 2023 & QA & 448 & Accuracy \\
\href{https://huggingface.co/datasets/gaia-benchmark/GAIA}{GAIA} & 2023 & QA & 450 & Exact Match \\
\href{https://github.com/openai/simple-evals}{BrowseComp} & 2025 & QA & 1266 & Exact Match \\
\href{https://texttron.github.io/BrowseComp-Plus/}{BrowseComp-Plus} & 2025 & QA & 830 & Accuracy / Recall / Search Call / Calibration Error \\
\href{https://lastexam.ai/}{HLE} & 2025 & QA & 2500 & Exact Match / Accuracy \\
\bottomrule
\end{tabularx}
\end{table}

%% file: table/resource.tex
\begin{table}[H]
{\scriptsize
\setlength{\tabcolsep}{3pt}
\rowcolors{2}{gray!10}{white}
\centering
\caption{Comprehensive overview of existing and emerging benchmarks for Deep Research Systems that focus on more boarder scenarios.}
\label{tab:benchmarks-more}
\begin{tabularx}{\textwidth}{l l X l X}
\toprule
\textbf{Benchmark (with link)} & \textbf{Date} & \textbf{Aspect} & \textbf{Data size (train/dev/test)} & \textbf{Evaluation metrics} \\
\midrule
\href{https://huggingface.co/datasets/google/frames-benchmark}{FRAMES} & 2024 & QA & 824 & Exact Match / F1 / Accuracy \\
\href{https://infodeepseek.github.io/}{InfoDeepSeek} & 2025 & QA & 245 & Accuracy / Utilization / Compactness \\
\href{https://assistantbench.github.io/}{AssistantBench} & 2025 & QA & 214 & F1 / Similarity \\
\href{https://osu-nlp-group.github.io/Mind2Web/}{Mind2Web} & 2025 & QA & 2350 & Accuracy / F1 / Step Success Rate \\
\href{https://osu-nlp-group.github.io/Mind2Web-2/}{Mind2Web 2} & 2025 & QA & 130 & Agent-as-a-Judge \\
\href{https://drb.futuresearch.ai/}{Deep Research Bench} & 2025 & QA & 89 & Precision / Recall / F1 \\
\href{https://www.deepresearchgym.ai/}{DeepResearchGym} & 2025 & QA & 96,000 & Report Relevance / Retrieval Faithfulness / Report Quality \\
\href{https://webarena.dev/}{WebArena} & 2024 & Complex Task & 812 & Correctness \\
\href{https://github.com/Alibaba-NLP/DeepResearch}{WebWalkerQA} & 2025 & QA & 680 & Accuracy / Action Count \\
\href{https://widesearch-seed.github.io/}{WideSearch} & 2025 & QA & 200 & LLM Judge \\
\href{https://github.com/shulin16/MMInA}{MMInA} & 2025 & Complex Task & 1050 & Success Rate \\
\href{https://github.com/AutoSurveys/AutoSurvey}{AutoSurvey} & 2024 & Survey Generation & 530,000 & Citation Quality / Content Quality \\
\href{https://github.com/ByteDance-BandAI/ReportBench}{ReportBench} & 2025 & Survey Generation & 600 & Content Quality / Cited Statement / Non-Cited Statements \\
\href{https://github.com/tongbao96/SurveyGen}{SurveyGen} & 2025 & Survey Generation & 4200 & Topical Relevance / Academic Impact / Content Diversity \\
\href{https://github.com/cxcscmu/Deep-Research-Comparator}{Deep Research Comparator} & 2025 & Report Generation & 176 & BradleyTerry Score \\
\href{https://github.com/Ayanami0730/deep_research_bench}{DeepResearch Bench} & 2025 & Report Generation & 100 & LLM Judge \\
\href{https://github.com/GAIR-NLP/ResearcherBench}{ResearcherBench} & 2025 & Report Generation & 65 & Rubric Assessment / Factual Assessment \\
\href{https://github.com/microsoft/LiveDRBench}{LiveDRBench} & 2025 & Report Generation & 100 & Precision / Recall / F1 \\
\href{https://proxy-qa.com/}{PROXYQA} & 2025 & Report Generation & 100 & LLM Judge \\
\href{https://github.com/AkariAsai/ScholarQABench}{SCHOLARQABENCH} & 2025 & Report Generation & 2967 & Accuracy / Citations / Rubrics \\
\href{https://github.com/Paper2Poster/Paper2Poster}{Paper2Poster} & 2025 & Poster Generation & 100 & Visual Quality / Textual Coherence / VLM Judge \\
\href{https://y-research-sbu.github.io/PosterGen/}{PosterGen} & 2025 & Poster Generation & 10 & Poster Content / Poster Design \\
\href{https://github.com/multimodal-art-projection/P2P}{P2PInstruct} & 2025 & Poster Generation & 121 & LLM Judge \\
\href{https://doc2ppt.github.io/}{Doc2PPT} & 2022 & Slides Generation & 6000 & ROUGE / Figure Subsequence / Text-Figure Relevance \\
\href{https://github.com/para-lost/AutoPresent}{SLIDESBENCH} & 2025 & Slides Generation & 7000/0/585 & Text / Image / Layout / Color / LLM Judge \\
\href{https://github.com/icip-cas/PPTAgent}{Zenodo10K} & 2025 & Slides Generation & 10,448 & Content / Design / Coherence \\
\href{https://github.com/KyuDan1/Talk-to-Your-Slides}{TSBench} & 2025 & Slides Generation & 379 & Editing Success / Efficiency \\
\href{https://ai-idea-bench.github.io/}{AI Idea Bench} & 2025 & Idea Generation & 0/0/3495 & LLM Judge \\
\href{https://github.com/HKUDS/AI-Researcher}{Scientist-Bench} & 2025 & Idea Generation, Experimental Execution & 0/0/52 & LLM Judge, Human Judge \\
\href{https://arxiv.org/pdf/2504.01848}{PaperBench} & 2025 & Experimental Execution & 0/0/20 & LLM Judge \\
\href{https://github.com/neulab/ReviewAdvisor}{ASAP-Review} & 2021 & Peer Review & 0/0/8877 & Human / ROUGE / BERTScore \\
\href{https://github.com/zhu-minjun/Researcher}{DeepReview} & 2025 & Peer Review & 13378/0/1286 & LLM Judge \\
\href{https://www.swebench.com/}{SWE-Bench} & 2023 & Software Engineering & 0/0/500 & Environment \\
\href{https://github.com/allenai/ScienceWorld}{ScienceWorld} & 2022 & Scientific Discovery & 3600/1800/1800 & Environment \\
\href{https://github.com/cognitiveailab/GPT-simulator}{GPT-Simulator} & 2024 & Scientific Discovery & 0/0/76369 & LLM Judge \\
\href{https://github.com/allenai/discoveryworld}{DiscoveryWorld} & 2024 & Scientific Discovery & 0/0/120 & LLM Judge \\
\href{https://github.com/siegelz/core-bench}{CORE-Bench} & 2024 & Scientific Discovery & 0/0/270 & Environment \\
\href{https://github.com/openai/mle-bench?tab=readme-ov-file}{MLE} & 2024 & Machine Learning Engineering & 0/0/75 & Environment \\
\href{https://github.com/METR/RE-Bench/tree/main}{RE-Bench} & 2024 & Machine Learning Engineering & 0/0/7 & Environment \\
\href{https://github.com/LiqiangJing/DSBench}{DSBench} & 2024 & Data Science & 0/0/540 & Environment \\
\href{https://spider2-v.github.io/}{Spider2-V} & 2024 & Data Science & 0/0/494 & Environment \\
\href{https://github.com/MetaCopilot/dseval}{DSEval} & 2024 & Data Science & 0/0/513 & LLM Judge \\
\href{https://iandrover.github.io/UnivEarth/}{UnivEARTH} & 2025 & Earth Observation & 0/0/140 & Exact Match \\
\href{https://github.com/commit-0/commit0}{Commit0} & 2024 & Software Engineering & 0/0/54 & Unit test \\
\bottomrule
\end{tabularx}}
\end{table}

%% file: sections/06-challenges.tex
\section{Challenges and Outlook}\label{sec:challenge}

\subsection{Retrieval Timing}

Although determining when to retrieve has become a standard feature of various DR systems, several fundamental challenges remain. Existing DR systems, such as Search-R1~\citet{jin2025search}, rely too heavily on answer correctness to guide the entire search pipeline and lack fine-grained guidance on when to retrieve, leading to both over-retrieval and under-retrieval~\citet{wu-etal-2025-search}. Moreover, even with continued retrieval, the model may still produce an incorrect answer, and when no relevant evidence can be retrieved, generating an answer regardless risks misleading users, particularly in safety-critical domains such as healthcare and finance.

Future research could explore fine-grained reward designs that assess, at each step, whether the model lacks the knowledge needed to answer the question~\cite{wu-etal-2025-search} and whether relevant documents can be retrieved~\cite{wang2025stepsearch}. Such signals would help determine when retrieval is necessary.
Beyond deciding when to retrieve, the system should also evaluate whether the model’s post-retrieval answer is correct and, after completing the entire process, estimate the uncertainty of the final output to avoid misleading users.

\subsection{Memory Evolution}

DR systems aim to mimic the research process of human experts by integrating autonomous planning, multi-source information acquisition, dynamic memory management, and deep knowledge synthesis. However, existing memory modules face significant challenges in fulfilling this vision. To develop more capable and DR systems, it is essential to re-examine the role of memory and identify future directions in personalization, structurization, adaptivity, and goal-driven optimization~\cite{gao2025survey,jiang2024long,fang2025comprehensive}.

\subsubsection{Proactive Personalization Memory Evolution}

\header{Recap of previous work.} Personalized memory in current systems often serves as a \textit{passive knowledge buffer}, primarily designed to record user interaction histories and preferences for enhancing retrieval-based responses \cite{zhang2025personaagent, niu2025part}. 
While effective for maintaining conversational consistency, this potentially limits the agent to a reactive stance \cite{niu2025part}. 
The memory's primary function is to serve as a repository of past events, such as the fine-grained, timestamped interactions stored in episodic memory or the consolidated user traits in semantic memory \cite{wang2025mirix, zhang2025personaagent}. 
Even advanced management techniques, such as the reflective mechanisms proposed by RMM~\cite{tan2025prospect}, are chiefly focused on optimizing the organization and retrieval of this historical data to improve the relevance of future responses, rather than enabling forward-looking planning.

A necessary paradigm shift is emerging, moving from memory as a historical archive to memory as a dynamic, predictive user model \cite{niu2025part}. To transition from mere assistants to true collaborators, future agents must leverage memory to engage in proactive reasoning. The foundation for such a model is a comprehensive, multi-dimensional user profile, as conceptualized in benchmarks like PersonaLens, which integrates demographics, detailed cross-domain preferences, and summaries of past interactions to form a holistic view of the user~\cite{zhao2025personalens}. Early steps in this direction can be seen in goal-oriented systems like MemGuide, which employs proactive reasoning by using the user's task intent and analyzing missing information \textit{slots} to strategically filter memories~\cite{du2025bridging}. The ultimate vision is for future memory modules to empower agents as proactive partners by capturing not only explicit preferences but also implicit signals, such as communication styles and latent intents. The PaRT framework exemplifies this future, using its dynamic user profile to actively guide conversations by generating personalized new topics and retrieving real-time external information~\cite{niu2025part}. A blueprint for the underlying architecture can be found in systems like MIRIX, whose multi-component design could support diverse proactive functions; for instance, its Procedural Memory could store workflows to anticipate a user's next steps in a complex task~\cite{wang2025mirix}. By integrating these capabilities, the system can anticipate user needs, proactively acquire and present relevant information, and adapt its interaction style in real time, thus shifting from reactive responses to proactive planning for more effective, intuitive, personalized support~\cite{niu2025part}.

\subsubsection{Cognitive-Inspired Structured Memory Evolution}

The predominant memory architecture in current systems (\eg vector stores of text chunks) follows a \textit{flat} storage paradigm, which lacks the capacity to capture deep logical or relational structures between knowledge elements. This architectural deficiency fundamentally hinders complex multi-hop reasoning, as the system cannot traverse explicit relationships between concepts. Recent work has begun to address this by moving towards structured representations like knowledge graphs, where entities are explicitly linked by semantic relationships, thereby providing a scaffold for more sophisticated inference \cite{chhikara2025mem0, xu2025memory, rasmussen2025zep}. Moreover, memory is often treated as a static snapshot, making it incapable of addressing the temporal dynamics of knowledge. This is a critical failure point in real-world scenarios where information evolves. Pioneering work has introduced bi-temporal models into knowledge graphs \cite{rasmussen2025zep}, allowing memory to track not only when a fact was recorded but also the period during which it was valid in the real world, using non-destructive updates that preserve historical context \cite{chhikara2025mem0, rasmussen2025zep}.

A key future direction is to integrate these structured memory representations with dynamic, autonomous update mechanisms, drawing inspiration from cognitive science. Agents should be capable of autonomously transforming unstructured inputs into structured representations (\eg knowledge graphs \cite{chhikara2025mem0, rasmussen2025zep}, operator trees \cite{christmann2025recursive}, or multi-faceted memory fragments \cite{zhang2025multiple}) in real time during interaction.
Importantly, this is not a one-time conversion, but a continuous \textit{stream-processing} procedure. As new information arrives, the memory structure must dynamically expand, prune, and reorganize itself \cite{chhikara2025mem0, xu2025mem, liu2023think}. This vision is partially realized in systems that employ agentic, cognitive-inspired operations such as \texttt{INSERT}, \texttt{FORGET}, and \texttt{MERGE} to refine memory content \cite{liu2023think}, or processes like \textit{memory evolution}, in which new memories trigger updates and recontextualization of existing, linked memories \cite{xu2025mem}. The ultimate goal is to create a unified cognitive framework that addresses the dual challenges of representational depth and timeliness of knowledge. This framework would likely emulate the distinction between human episodic and semantic memory, a principle already explored in several advanced architectures \cite{nan2025nemori, rasmussen2025zep, zhang2025multiple, yang2025coarse}, allowing an agent to both ground its knowledge in specific experiences and evolve a generalized, abstract understanding of the world.

\subsubsection{Goal-Driven Reinforced Memory Evolution}

Existing strategies for memory retention are primarily heuristic-based, relying on static signals such as recency or semantic relevance \cite{yan2025memory, zhang2025learn}. However, these heuristics fail to guarantee that preserved memories are truly useful for achieving the final task goal, as they often ignore the interconnected memory cycle effect of storage, retrieval, and utilization \cite{zhang2025learn}. A more powerful paradigm is to formulate memory management as a decision-making problem within a RL framework \cite{yu2025memagent, zhou2025mem1, long2025seeing, zhou2025agentfly, yan2025memory}. In this approach, the agent learns an optimal policy for memory operations, such as updating a fixed-length internal state \cite{yu2025memagent, zhou2025mem1} or executing structured commands like \texttt{ADD}, \texttt{UPDATE}, and \texttt{DELETE} on a memory store \cite{yan2025memory}. The learning process is guided solely by the reward from the final task outcome, forcing memory management to emerge as a goal-aligned, adaptive capability \cite{yu2025memagent, zhou2025mem1, yan2025memory}.

A key direction lies in extending this RL paradigm to jointly optimize the entire memory cycle, where agents learn not just to store information but to dynamically retrieve and utilize it through sophisticated strategies, such as multi-round reasoning~\cite{long2025seeing} and experience reuse~\cite{zhou2025agentfly, zhang2025learn}. This goal is becoming increasingly practical due to two key advances. First, emerging frameworks enable policy learning for memory management at low cost and in real time, without requiring expensive LLM fine-tuning~\cite{zhou2025agentfly}. Second, the data efficiency of RL training makes this approach viable even in data-scarce domains~\cite{yan2025memory}. However, despite these promising developments, a fundamental obstacle remains: the long-term credit assignment problem, which involves developing reliable algorithms to attribute a final outcome to a long sequence of intermediate memory decisions~\cite{yu2025memagent}.

\subsection{Instability in Training Algorithms}
In DR systems, multiple rounds of interaction with the environment are required. 
Although RL algorithms such as PPO~\cite{schulman2017proximal} and GRPO~\cite{shao2024deepseekmath} exhibit stable behavior in single-turn scenarios, they often become unstable when extended to multi-turn settings.
This instability typically appears as a gradual or abrupt drop in reward, the generation of invalid responses, and symptoms such as entropy collapse and gradient explosion~\cite{jin2025search,xue2025simpletir,wang2025ragen,verl2025issue}. These issues remain persistent challenges for training agentic RL systems.
Below, we examine two newly emerging solutions and outline future directions for further study.

\subsubsection{Existing Solutions}

\header{Filtering void turns.}
The first representative solution is proposed by \citet{xue2025simpletir}, who identify \textit{void turns} as a major cause of collapse in multi-turn RL.
Void turns refer to responses that do not advance the task, such as fragmented text, repetitive content, or premature termination; and once produced, they propagate through later turns, creating a harmful feedback loop.
These errors largely stem from the distribution shift between pre-training and multi-turn inference, where the model must process external tool outputs or intermediate signals that were not present during pre-training, increasing the chance of malformed generations.
To address this, SimpleTIR~\cite{xue2025simpletir} filters out trajectories containing void turns, effectively removing corrupted supervision and stabilizing multi-turn RL training.

\header{Mitigating the Echo Trap.}
\citet{wang2025ragen} identify the \textit{Echo Trap} as a central cause of collapse in multi-turn RL.
The Echo Trap refers to rapid policy homogenization, where the model abandons exploration and repeatedly produces conservative outputs that yield short-term rewards. Once this happens, reward variance and policy entropy drop sharply, forming a self-reinforcing degenerative loop.
The root cause is a misalignment between reward-driven optimization and reasoning quality.
In multi-turn settings, sparse binary rewards cannot distinguish coincidental success from genuine high-quality reasoning, encouraging reward hacking behaviors such as hallucinated reasoning or skipping essential steps.
To address this, the proposed StarPO-S~\cite{wang2025ragen} uses uncertainty-based trajectory filtering to retain trajectories exhibiting meaningful exploration. This breaks the Echo Trap cycle and stabilizes multi-turn RL training.

\subsubsection{Future Directions}
Beyond the solutions discussed above, we highlight two additional directions for achieving more stable agentic RL training.

\header{Cold-start methods that preserve exploration.}
SFT is a practical cold-start strategy for multi-turn RL, yet it introduces a significant drawback: it rapidly reduces output entropy, constraining the model’s ability to explore and develop new reasoning strategies~\cite{zhu2025proximal}.
A promising research direction is to design cold-start methods that improve initial task performance while maintaining exploratory behavior. Such techniques should aim to avoid early entropy collapse and preserve the model’s capacity for innovation in multi-turn reasoning.

\header{Denser and smoother reward design.}
Although StarPO-S~\cite{wang2025ragen} effectively mitigates training collapse in PPO-based multi-turn RL, its benefit is more limited for GRPO. The critic module inherent to PPO algorithm~
\cite{schulman2017proximal} naturally smooths reward signals, while GRPO relies on group-wise normalization, which makes it more sensitive to reward variance and extreme values.
Developing denser, smoother, and more informative reward functions for multi-turn scenarios, especially for GRPO-style algorithms, remains an important direction for future research.

\subsection{Evaluation of Deep Research System}

Evaluation of DR generally falls into two complementary aspects: 
(i) the evaluation of agentic information-seeking capabilities and 
(ii) the evaluation of long-form generation~\citep{li2025reportbenchevaluatingdeepresearch,yang2018hotpotqa}. 
Considerable progress has been made in the former, with benchmarks such as HotpotQA~\cite{yang2018hotpotqa}, GAIA~\cite{mialon2023gaia}.
The more recent Deep Research Bench~\cite{du2025deepresearch,wang2025liveresearchbench} further provides increasingly complex and interactive settings for evaluating agents' abilities to retrieve, navigate, and synthesize information across dynamic web environments. 
However, reliably evaluating model-generated long-form outputs, especially research-style reports in response to open-ended and high-level queries, remains an open and pressing challenge~\cite{wei2024long}. 
Most existing approaches rely on LLM-as-a-Judge to directly evaluate general dimensions such as content factuality, structural coherence, and readability~\cite{xu2025researcherbench}. 
While effective for scalable comparisons, these evaluation strategies ignore crucial dimensions and are subject to several limitations. 

\subsubsection{Logical Evaluation}

Since DR typically requires long-form context generation, maintaining logical coherence throughout the text is essential~\citep{zheng2025long}.
Existing studies suggest that while LLMs demonstrate strong capabilities for recognizing logical patterns, such as in summarization tasks or the detection of inconsistencies in short passages, their ability to create rigorous logical chains during DR remains uncertain~\cite{li2025summary}.
In particular, when required to synthesize insights from multiple retrieved supporting documents, models often fail to consistently transform fragmented evidence into a logically connected narrative.
The generated reasoning may contain gaps, abrupt leaps, or even circular justifications, compromising the argument's fidelity~\cite{que2024hellobench}.
This limitation underscores a key challenge for DR: the task is not merely to produce fluent and factually plausible text, but to articulate insights that are logically well-founded and epistemically defensible~\cite{liu2024longgenbench}.

Accordingly, robust logical evaluation emerges as a central challenge. 
However, most existing research on logical assessment remains narrowly scoped.
Current benchmarks typically address limited logical tasks, such as solving symbolic logic puzzles, identifying entailment in short sentences, or handling deductive reasoning in synthetic settings~\citep{wei2025satbench,parmar2024towards}.
While these tasks provide valuable insights into basic reasoning abilities, they fall short of capturing the complexities of long-form logical consistency.
Specifically, they do not address whether models can sustain coherent argumentative structures across extended contexts, reconcile conflicting sources, or systematically avoid introducing unsupported claims.
One potential approach is to design evaluation frameworks that assess coherence across multiple granularities (\eg sentence-level, paragraph-level, and document-level), capturing both local and global logical dependencies~\cite{guan2021long}.

\subsubsection{Boundary between Novelty and Hallucination}
In DR, progressing beyond faithful summarization toward the generation of genuinely novel hypotheses or perspectives is a central goal~\cite{franceschelli2025creativity}. 
However, in practice, outputs that appear original may embed unverifiable claims, fabricated connections between sources, or spurious inferences lacking epistemic grounding~\cite{lin2025evaluating}. 
This challenge is exacerbated in open-ended settings, where no single ground truth exists and retrieval broadens the hypothesis space, increasing the likelihood that superficially plausible but unsupported statements evade detection, especially by surface-level or style-sensitive evaluation methods~\cite{jiang2024survey}.
Current practices often depend on density-based novelty scores or LLM-as-judge assessments of originality~\citep{wang2025enabling,zhang2025noveltybench}, yet these alone do not ensure verifiability or differentiate between creative recombination and unfounded speculation.

A potential solution is to differentiate between two types of novelty. 
Generative novelty refers to new combinations or perspectives, while deductive novelty refers to conclusions logically derived from known facts. 
To achieve this, novelty scoring can be combined with validity-checking mechanisms~\cite{azerbayev2023proofnet, yu2025formalmath, ren2025deepseek}. 
For example, researchers can pre-register testable claims along with verification plans, ensure that each claim is linked to clear sources, and systematically ablate sources to determine which are necessary or sufficient~\citep{zheng2021minif2f, tsoukalas2024putnambench}. 
Additionally, inserting control examples or testing the system with false information can reveal how often it generates incorrect but seemingly original results~\cite{ming2024faitheval}. 
Another useful method is to restrict the system to documents published before a certain cutoff date and then examine whether its outputs are later validated by subsequently published sources—providing insight into the independence and robustness of novel ideas~\cite{liu2025exante,karger2024forecastbench}.

\subsubsection{Bias and Efficiency of LLM-as-Judge}

LLM-as-Judge has become a mainstream approach for evaluating long-form model outputs.
However, this practice introduces two major challenges.
\textbf{The first challenge is bias.}
LLM judges may prefer longer responses, be affected by answer ordering, reward particular writing styles, or favor systems that resemble themselves~\citep{gu2024survey, li2024llms}.
Such biases may reduce the robustness and fairness of existing evaluation protocols.
\textbf{The second challenge is efficiency.}
Large-scale pairwise evaluation is resource-intensive, especially when relying on paid APIs and applying costly comparison methods to long outputs~\cite{zhu2023judgelm}.
These limitations motivate two directions for improvement: mitigating bias and improving efficiency.

\header{Mitigating bias.}
Bias can be reduced by incorporating human evaluators for critical or ambiguous cases, providing a grounded reference for calibration~\cite{li2024llms}.
Another direction is to fine-tune judge models using datasets that highlight diverse reasoning styles and explicit debiasing signals~\cite{zhu2023judgelm}.
Such training may lessen systematic preferences for particular formats or linguistic patterns.

\header{Improving efficiency.}
Efficiency can be improved by adopting open-source, general-purpose judge models, which reduce evaluation cost while offering greater transparency and reproducibility~\cite{sahoo2025quantitative}.
Further improvements may come from smarter candidate selection algorithms that focus on the most informative comparisons~\cite{zhen2025enhancing}.
By lowering the number of required pairwise evaluations without sacrificing quality, such methods enable LLM-based evaluation in more resource-constrained settings.

%% file: sections/07-discussion.tex
\section{Open Discussion: Deep Research to General Intelligence}\label{sec:discussion}

As DR systems advance, they must navigate key challenges that bridge specialized task-solving with broader cognitive capabilities. 
This subsection examines three pivotal areas (\ie creativity, fairness, safety, and reliability) that may shape the development from current DR paradigms to AGI-level autonomy, ensuring these systems not only augment human inquiry but also foster equitable, innovative, and trustworthy ecosystems.

\subsection{Creativity}
Despite the considerable attention and rapid development in both academia and industry, Previous studies highlight fundamental limitations in LLM creativity based on next-token prediction \citep{lu2025rethinking}. While they excel at recombination~\cite{xu2025memory,yan2025memory}, emotion~\cite{yang2025uncertain, yang2024emollm}, imitation~\cite{li2024llms, wang2025coser}, and logical reasoning~\cite{wan2024logicasker, sprague2024cot}, the question remains whether AI can evolve from these capabilities to achieve genuine innovation and novel concept generation. This transition may require mechanisms beyond statistical learning, drawing on psychological theories of human creativity, such as \textit{insight} or \textit{eureka moments,} which involve sudden restructuring of mental representations and are not easily explained by probabilistic models~\cite{wu2025}. Some argue that hallucinations in AI could be interpreted as a form of creativity~\cite{lu2025rethinking}, potentially bridging this gap, but this perspective needs careful examination to distinguish between productive divergence and erroneous output.

\subsection{Fairness}
As noted in prior work~\cite{ferrara2024fairness}, DR powered by autonomous agents may inadvertently inherit and amplify existing biases in academia.
For example, they could favor mainstream fields, methodologies, or prominent researchers, thereby overlooking emerging interdisciplinary work or contributions from non-mainstream regions.r
To mitigate this, such systems should incorporate built-in fairness frameworks that ensure comprehensive and impartial evaluation of all data, prevent the reinforcement of academic hierarchies, and provide equitable support to researchers from diverse backgrounds. A critical consideration is the impact of each agent's decision step on the overall fairness of outcomes: how much does bias in early steps shape subsequent decision spaces during interactions with the environment? Recent work~\cite{kheya2024pursuit} indicates that this cascading effect could limit exploration and perpetuate inequities if not addressed through debiasing techniques at every stage.

\subsection{Safety and Reliability}

Although some studies suggest that AI hallucinations can spark diversity~\cite{lu2025rethinking,2025cognibench}, they also pose risks of disseminating serious academic errors. To enhance safety and reliability, \Drs should ensure conclusions are supported by clear, traceable evidence chains; offer highly transparent reasoning processes to avoid ``black-box'' decisions; and implement robust validation mechanisms to curb the spread of hallucinated science~\cite{chen2024spiral, si2024can}. These measures are essential for maintaining trust in AI-assisted research and preventing misinformation in scholarly pursuits.

%% file: sections/08-conclusion.tex
\section{Conclusion and Future Outlook}\label{sec:conclusion}

Deep research (DR) stands at the frontier of transforming large language models from passive responders into autonomous investigators capable of iterative reasoning, evidence synthesis, and verifiable knowledge creation. This survey consolidates recent advances in architectures, optimization methods, and evaluation frameworks, providing a unified roadmap for understanding and building future DR systems. By investigating relevant works, this survey facilitates future research and accelerates the advancement of DR systems toward more general, reliable, and interpretable intelligence.
Given the rapid evolution of this field, we will continuously update this survey to encompass emerging paradigms such as multimodal reasoning, self-evolving memory, and agentic reinforcement learning. 
This effort aims to provide a comprehensive and up-to-date understanding of deep research systems.